\documentclass[a4paper,fleqn]{cas-dc}

\usepackage[numbers,sort&compress]{natbib}

\usepackage{amsmath,amssymb,amsthm}
\usepackage{graphicx}
\usepackage{booktabs}
\usepackage{multirow}
\usepackage{array}
\usepackage[table]{xcolor}
\usepackage{pifont}
\usepackage{url}
\usepackage{tikz}
\usetikzlibrary{arrows.meta,positioning,fit,backgrounds,calc,shapes.geometric,decorations.pathreplacing,patterns}
\usepackage{pgfplots}
\pgfplotsset{compat=1.17}
\usepgfplotslibrary{groupplots,fillbetween}
\usepackage{hyperref}
\newtheorem{theorem}{Theorem}
\newtheorem{lemma}{Lemma}
\newtheorem{proposition}{Proposition}
\theoremstyle{definition}
\newtheorem{definition}{Definition}
 
\newcommand{\cmark}{\ding{51}}
\newcommand{\xmark}{\ding{55}}
\newcommand{\F}{\mathrm{F}}
\DeclareMathOperator{\diag}{diag}
\DeclareMathOperator{\spec}{spec}
\DeclareMathOperator{\GELU}{GELU}
\DeclareMathOperator{\LN}{LN}

\definecolor{cGWN}{HTML}{B5651D}
\definecolor{cMega}{HTML}{7F7F7F}
\definecolor{cSTEP}{HTML}{E69F00}
\definecolor{cD2}{HTML}{56B4E9}
\definecolor{cSTAE}{HTML}{CC79A7}
\definecolor{cTG}{HTML}{009E73}
\definecolor{cOurs}{HTML}{0B4F9C}

\def\tsc#1{\csdef{#1}{\textsc{\lowercase{#1}}\xspace}}
\tsc{WGM}
\tsc{QE}

\begin{document}
\let\WriteBookmarks\relax
\def\floatpagepagefraction{1}
\def\textpagefraction{.001}

\shorttitle{ST-GFM}    

\shortauthors{A. Bohra and V. Vijay}

\title [mode = title]{ST-GFM: Spectral-Temporal Graph Forecasting with Measured Perturbation Bounds and Masked Spectral Reconstruction}

\author[1]{Aashish Bohra}[orcid=0009-0004-5871-4310]
\cormark[1]
\ead{bohra.1@iitj.ac.in}

\affiliation[1]{organization={Department of Computer Science and Engineering, Indian Institute of Technology Jodhpur},
            city={Jodhpur},
            state={Rajasthan},
            country={India}}

\author[2]{Vivek Vijay}[orcid=0000-0002-1359-1861]
\ead{vivek@iitj.ac.in}

\affiliation[2]{organization={Department of Mathematics, Indian Institute of Technology Jodhpur},
            city={Jodhpur},
            state={Rajasthan},
            country={India}}

\cortext[1]{Corresponding author}


\begin{abstract}
Graph forecasters for sensor networks are trained in one regime and deployed in others, where sensor coupling and signal frequency content both change. We split this spectral shift into an operator channel and a frequency channel, each with a dimensionless measure. ST-GFM (Spectral-Temporal Graph Forecasting Model) filters every input step with Chebyshev and bidirectional diffusion filters on one learned, non-symmetric operator and alternates them with per-node temporal attention. For the operator channel we prove Frobenius-norm perturbation bounds for both graph branches, show that the Chebyshev bound is attained to first order, and compute its constant from the trained weights and the measured operator norm: $17.9$ on METR-LA and $21.5$ on PEMS-BAY. Under random edge dropout the trained layer moves by $2.0\%$ to $3.1\%$ of this bound. Masked Spectral Reconstruction (MSR), a self-supervised loss active in the first 40 epochs, matches the log power spectrum of the encoder output to that of the input. For small log-spectral residuals on unclamped bins it is proportional to an Itakura--Saito divergence to leading order. Removing it raises the 60-minute MAE by $0.30$ on METR-LA, close to the $0.33$ lost when all three self-supervised losses go. Under one shared evaluation pipeline, the 60-minute MAE of ST-GFM is $3.5\%$ below a retrained STAEformer on METR-LA and $13.0\%$ below it on PEMS-BAY. Its 60-minute MAE is $3.098\pm0.017$ on METR-LA (five seeds) and $1.576\pm0.006$ on PEMS-BAY (three seeds), the lowest PEMS-BAY MAE among twelve published baselines. 

\end{abstract}





\maketitle

\section{Introduction}
\label{sec:intro}
 
A loop detector network on a metropolitan freeway looks different at 8~AM and at 10~PM. In the morning, sensors along a few commuter corridors move together, cross-corridor coupling is weak, and speed series contain sharp transitions as queues form and clear. Late at night those clusters dissolve, coupling becomes diffuse, and the series are smooth. A graph neural forecaster fitted across both regimes uses one graph operator and one set of filters for both~\cite{li2018dcrnn,wu2019gwnet}. Agencies use its output for ramp metering, incident response and traveller information, so a forecaster that holds up when the regime changes is worth more than one that only wins on a fixed split. Energy, climate and financial networks show the same pattern at daily, weekly and seasonal scales.
 
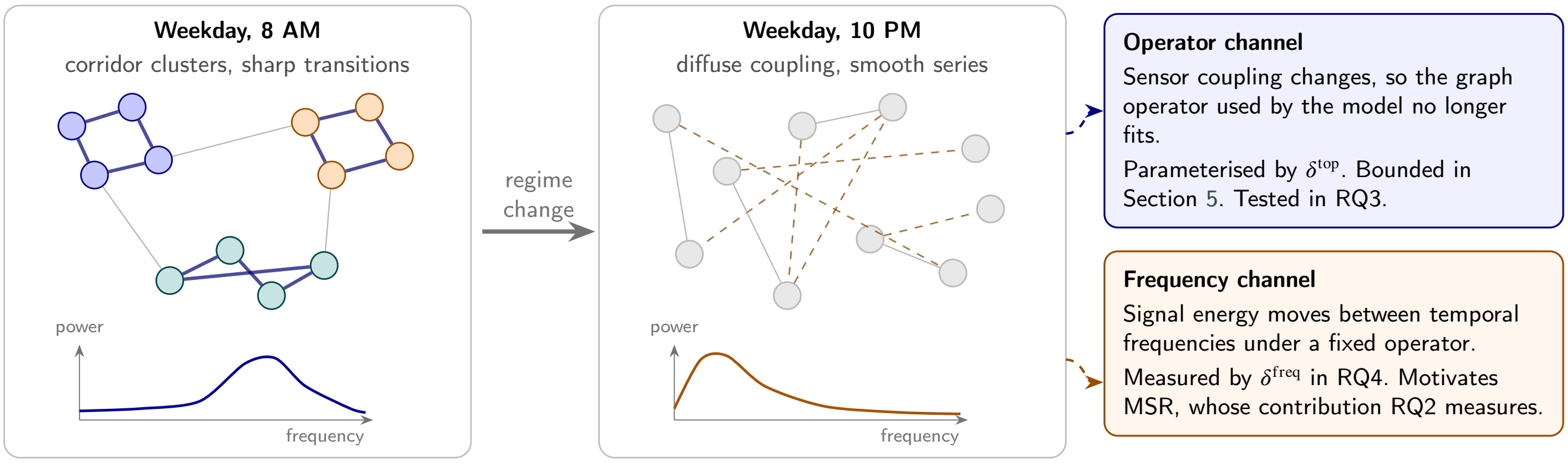
\begin{figure*}[!t]
\centering
\resizebox{\textwidth}{!}{%
\begin{tikzpicture}[
  font=\sffamily\small, >=Stealth,
  sen/.style={circle, draw, line width=0.6pt, minimum size=0.36cm, inner sep=0pt},
  sA/.style={sen, fill=blue!22, draw=blue!55!black},
  sB/.style={sen, fill=orange!25, draw=orange!60!black},
  sC/.style={sen, fill=teal!22, draw=teal!55!black},
  sG/.style={sen, fill=gray!18, draw=gray!55},
  strong/.style={line width=1.3pt, draw=blue!45!black, opacity=0.7},
  weak/.style={line width=0.5pt, draw=gray!55},
  long/.style={line width=0.6pt, draw=orange!55!black, dashed, opacity=0.75},
  panel/.style={rounded corners=6pt, draw=black!25, line width=0.6pt, fill=white},
  chan/.style={rounded corners=5pt, draw, line width=0.7pt, align=left, inner sep=7pt, text width=5.6cm},
]
\node[panel, minimum width=6.2cm, minimum height=6.0cm] (pa) at (3.1,-2.6) {};
\node[font=\sffamily\bfseries] at (3.1,0.05) {Weekday, 8 AM};
\node[text=black!70] at (3.1,-0.4) {corridor clusters, sharp transitions};
\node[sA] (a1) at (0.9,-1.2) {}; \node[sA] (a2) at (1.7,-0.95) {};
\node[sA] (a3) at (1.2,-1.85) {}; \node[sA] (a4) at (2.05,-1.65) {};
\node[sB] (b1) at (4.0,-1.15) {}; \node[sB] (b2) at (4.85,-0.95) {};
\node[sB] (b3) at (4.35,-1.85) {}; \node[sB] (b4) at (5.25,-1.6) {};
\node[sC] (c1) at (2.2,-3.25) {}; \node[sC] (c2) at (3.0,-2.85) {};
\node[sC] (c3) at (3.55,-3.45) {}; \node[sC] (c4) at (4.25,-3.05) {};
\foreach \u/\v in {a1/a2,a1/a3,a2/a4,a3/a4,b1/b2,b1/b3,b2/b4,b3/b4,c1/c2,c2/c3,c3/c4,c1/c4} \draw[strong] (\u)--(\v);
\foreach \u/\v in {a4/b1,a3/c1,b3/c4} \draw[weak] (\u)--(\v);
\begin{scope}[shift={(1.0,-5.1)}]
  \draw[->, black!55] (0,0)--(3.9,0) node[below left, font=\sffamily\scriptsize]{frequency};
  \draw[->, black!55] (0,0)--(0,1.0) node[above, font=\sffamily\scriptsize]{power};
  \draw[line width=1.0pt, blue!60!black] plot[smooth] coordinates {(0,0.12) (0.8,0.15) (1.6,0.25) (2.2,0.75) (2.6,0.82) (3.0,0.45) (3.6,0.15) (3.8,0.1)};
\end{scope}
\draw[->, line width=1.6pt, black!55] (6.35,-2.6) -- (7.85,-2.6) node[midway, above, align=center, font=\sffamily\small]{regime\\change};
\node[panel, minimum width=6.2cm, minimum height=6.0cm] (pb) at (11.0,-2.6) {};
\node[font=\sffamily\bfseries] at (11.0,0.05) {Weekday, 10 PM};
\node[text=black!70] at (11.0,-0.4) {diffuse coupling, smooth series};
\node[sG] (d1) at (8.8,-1.1) {}; \node[sG] (d2) at (9.6,-1.8) {};
\node[sG] (d3) at (10.6,-1.2) {}; \node[sG] (d4) at (11.8,-0.95) {};
\node[sG] (d5) at (12.9,-1.5) {}; \node[sG] (d6) at (9.1,-2.9) {};
\node[sG] (d7) at (10.4,-3.45) {}; \node[sG] (d8) at (11.5,-2.7) {};
\node[sG] (d9) at (12.6,-3.15) {}; \node[sG] (d10) at (13.1,-2.3) {};
\foreach \u/\v in {d1/d9,d2/d5,d3/d7,d4/d6,d8/d10,d7/d4} \draw[long] (\u)--(\v);
\foreach \u/\v in {d1/d6,d8/d9,d2/d7,d3/d4} \draw[weak] (\u)--(\v);
\begin{scope}[shift={(8.9,-5.1)}]
  \draw[->, black!55] (0,0)--(3.9,0) node[below left, font=\sffamily\scriptsize]{frequency};
  \draw[->, black!55] (0,0)--(0,1.0) node[above, font=\sffamily\scriptsize]{power};
  \draw[line width=1.0pt, orange!65!black] plot[smooth] coordinates {(0,0.15) (0.35,0.8) (0.7,0.85) (1.2,0.45) (2.0,0.18) (3.0,0.1) (3.8,0.08)};
\end{scope}
\node[chan, draw=blue!50!black, fill=blue!6, anchor=north west] (ch1) at (14.6,0.3)
  {\textbf{Operator channel}\\[2pt]
   Sensor coupling changes, so the graph operator used by the model no longer fits.\\[2pt]
   Parameterised by $\delta^{\mathrm{top}}$. Bounded in Section~\ref{sec:theory}. Tested in RQ3.};
\node[chan, draw=orange!60!black, fill=orange!7, anchor=north west] (ch2) at (14.6,-2.85)
  {\textbf{Frequency channel}\\[2pt]
   Signal energy moves between temporal frequencies under a fixed operator.\\[2pt]
   Measured by $\delta^{\mathrm{freq}}$ in RQ4. Motivates MSR, whose contribution RQ2 measures.};
\draw[->, line width=0.9pt, blue!50!black, dashed] (14.1,-1.3) to[out=0,in=180] (ch1.west |- 0,-1.0);
\draw[->, line width=0.9pt, orange!60!black, dashed] (14.1,-4.3) to[out=0,in=180] (ch2.west |- 0,-4.6);
\end{tikzpicture}}%
\caption{Spectral shift between two regimes of the same freeway sensor network. The left and centre panels are schematic: sensor coupling above, the power spectrum of a typical speed series below. Spectral shift reaches a deployed model through an operator channel and a frequency channel, which the paper analyses and tests separately.}
\label{fig:shift}
\end{figure*}

We call the joint change in sensor coupling and signal frequency content {spectral shift}. It reaches a deployed model through two channels (Figure~\ref{fig:shift}). In the {operator channel}, the graph operator the model uses no longer describes the coupling it faces. This happens when a distance graph is re-estimated, when failed sensors take their edges with them, or when a graph fitted on one period is reused on another. In the {frequency channel}, the operator is unchanged but the inputs carry energy at temporal frequencies that differ from training. The two channels call for different tools. The first asks how a filter responds when its operator moves, which spectral graph theory can answer. The second asks what the encoder has learned to represent, which the training objective can influence.
 
Spatio-temporal forecasters are judged almost entirely on in-distribution accuracy. DCRNN \cite{li2018dcrnn}, Graph WaveNet \cite{wu2019gwnet}, MTGNN \cite{wu2020mtgnn}, D$^2$STGNN \cite{shao2022d2stgnn}, STAEformer \cite{liu2023staeformer} and others report strong numbers on METR-LA and PEMS-BAY, and none of them states how its predictions respond to a change of the operator. Perturbation bounds for graph filters exist~\cite{gama2020stability,levie2021transferability,kenlay2021interpretable,gao2021stochastic}, and CITRUS~\cite{einizade2024citrus} carries such an analysis into spatio-temporal forecasting. These bounds are written for generic filter banks or for other architectures, and none of these works computes its constants from the trained weights of an attention-based traffic forecaster. Existing self-supervised and frequency-aware objectives~\cite{zhang2022tfc,woo2022cost,shao2022step,gao2024stdmae,kara2024freqmae,wang2025fredf} supervise embeddings, reconstructed values, spectrogram values or forecast coefficients. Under a squared error on values, the weak, fast bins of a speed window carry little weight next to the slow trend (Section~\ref{subsec:msr_is}). We have not found a pretext task that compares the log power spectrum of a graph encoder's output with that of its input, which expresses every spectral discrepancy as a log-power ratio.
 
ST-GFM (Spectral-Temporal Graph Forecasting Model) is built around the two channels. Its encoder alternates Chebyshev-plus-diffusion graph filtering on a learned, non-symmetric operator with per-node temporal attention, and it keeps all twelve input steps through every graph layer, so each filter acts on the traffic state at one step. For the operator channel we bound how far each graph branch moves under an operator perturbation, compute the constants from the trained weights and the measured operator, and compare the Chebyshev bound with the measured response of the layer it describes. The frequency channel motivates the training objective. Masked Spectral Reconstruction (MSR) matches the log power spectral density (PSD) of the encoder output to that of the input during the first 40 epochs. Under one shared evaluation pipeline, the 60-minute MAE of ST-GFM is $3.5\%$ below a retrained STAEformer on METR-LA and $13.0\%$ below it on PEMS-BAY. The evaluation follows four research questions (RQs).
 
\begin{itemize}
  \item \textbf{RQ1.} Does ST-GFM improve multi-horizon accuracy over strong published baselines on speed and flow data, and is the improvement stable across seeds and across evaluation pipelines?
  \item \textbf{RQ2.} Which components account for the accuracy?
  \item \textbf{RQ3.} How sensitive is the forecast to perturbations of the graph operator, and how does the measured response compare with the bound?
  \item \textbf{RQ4.} Where does frequency shift occur in traffic, and how does ST-GFM's accuracy on METR-LA weekends compare with its full-test accuracy?
\end{itemize}
 
\noindent The paper makes four contributions.
 
\begin{enumerate}
  \item {Perturbation bounds whose constants are measured on the trained model.} We bound both graph branches and the encoder stack in the Frobenius norm without assuming a symmetric operator, so the bounds cover the learned operator ST-GFM uses. The Chebyshev bound is attained to first order on a constructed example. Its constant, computed from the trained weights and the measured norm of the trained operator, is $17.9$ on METR-LA and $21.5$ on PEMS-BAY. Under edge dropout the observed change of the first DCSE layer is $2.0\%$ to $3.1\%$ of the bound and grows close to linearly in the size of the edit.
  \item {A two-channel account of spectral shift with a protocol to match.} Definition~\ref{def:shift} gives the operator channel and the frequency channel each a dimensionless measure. The protocol perturbs the operator with the inputs held fixed (RQ3) and measures frequency shift on real data with the operator held fixed (RQ4). On both test sets the frequency term concentrates in weekday peak hours and is lowest on weekends.
  \item {MSR, a pretext loss on the log spectrum of the encoder output.} MSR aligns the log-PSD of the encoder output with that of its input. For small log-spectral residuals on unclamped bins, neglecting numerical stabilisation, the objective is proportional to the Itakura--Saito divergence to leading order. Among the objectives reviewed in Section~\ref{sec:related}, it is the only one that compares log spectra. Removing MSR raises the 60-minute MAE by $0.30$ on METR-LA and $0.22$ on PEMS-BAY, against $0.33$ and $0.24$ when all three self-supervised losses are removed.
  \item {A spectral encoder on a learned, non-symmetric operator.} DCSE runs an isotropic Chebyshev filter and a two-direction diffusion filter on one learned anchor operator at every input step and alternates with per-node temporal attention. Replacing DCSE with a standard GCN costs $0.32$ MAE at 60 minutes on METR-LA and $0.21$ on PEMS-BAY. Collapsing the time axis before the graph layers costs $0.12$ and $0.08$.
\end{enumerate}
 
Section~\ref{sec:related} reviews related work. Section~\ref{sec:problem} defines the task and spectral shift, Section~\ref{sec:model} the model, and Section~\ref{sec:theory} the analysis. Sections~\ref{sec:setup} and~\ref{sec:results} report the experiments, including an exploratory transfer to industry portfolio returns (Section~\ref{subsec:transfer}). Section~\ref{sec:discussion} interprets them and Section~\ref{sec:limitations} states their scope.
 
\section{Related Work}\label{sec:related}
The related work is categorized into four categories which are spatio-temporal forecasting on sensor graphs, distribution shift in temporal and graph data, spectral graph filters and their stability, and self-supervised and frequency-aware pre-training. We examine how existing methods capture spatial and temporal dependencies, adapt to changing data distributions, characterize the sensitivity of graph filters to structural perturbations, and learn representations from temporal and spectral information.

\subsection{Spatio-Temporal Forecasting on Sensor Graphs}
Most models pair a graph module for spatial coupling with a sequence module for time. DCRNN~\cite{li2018dcrnn} places diffusion convolution inside a recurrent encoder-decoder. STGCN~\cite{yu2018stgcn} and ASTGCN~\cite{guo2019astgcn} apply Chebyshev graph convolution to each time slice of a fixed distance graph, with temporal convolution in STGCN and spatio-temporal attention in ASTGCN. Graph WaveNet~\cite{wu2019gwnet} learns an adaptive adjacency $\mathrm{softmax}(\mathrm{ReLU}(E_1E_2^\top))$ from two node-embedding tables, the form ST-GFM also uses. AGCRN~\cite{bai2020agcrn} builds its graph from one embedding table and adds node-specific parameters, MTGNN~\cite{wu2020mtgnn} learns the graph jointly with dilated temporal convolutions, and AdpSTGCN~\cite{zhang2024adpstgcn} combines several feature graphs with the road graph through adaptive graph convolution. GMAN~\cite{zheng2020gman}, D$^2$STGNN~\cite{shao2022d2stgnn}, DSTAGNN~\cite{lan2022dstagnn} and MegaCRN~\cite{jiang2023megacrn} model time-varying or decoupled spatial structure. Attention-based models now hold most of the leading numbers. STAEformer~\cite{liu2023staeformer} shows that adaptive spatio-temporal embeddings make a plain Transformer competitive, PDFormer~\cite{jiang2023pdformer} and Trafformer~\cite{jin2023trafformer} model propagation delay and joint space-time attention, T-Graphormer~\cite{bai2025tgraphormer} applies Graphormer-style structural encodings to space-time tokens, and TITAN~\cite{wang2024titan} routes between sequence-centric and variable-centric experts. STID~\cite{shao2022stid} is a useful corrective: identity embeddings and an MLP recover much of the accuracy on these benchmarks. ST-GFM uses identity embeddings for that reason, and Section~\ref{subsec:rq2} reads its ablation with this in mind. Graph WaveNet and MTGNN shorten the window layer by layer with dilated temporal convolutions, so later graph convolutions see fewer and coarser steps. STGCN shortens it between graph layers with unpadded convolutions. DCRNN applies its diffusion filter to a recurrent state that summarises all earlier steps. GMAN and STAEformer keep every step but use attention in place of a spectral filter. Keeping the time axis is therefore not new by itself. ST-GFM combines it with a spectral graph path on a learned, non-symmetric operator and states how far that path moves when the operator moves.
 
\subsection{Distribution Shift in Temporal and Graph Data}
RevIN \cite{kim2022revin} and Non-stationary Transformers \cite{liu2022nonstationary} handle drift in per-series statistics through normalisation. DIDA \cite{zhang2022dida} learns invariant patterns in dynamic graphs under spatio-temporal distribution shift, and the survey of Kazemi et al. \cite{kazemi2020representation} covers models that track evolving structure. These methods target the input distribution or the graph sequence. They do not state how a fixed spectral filter responds when its operator changes, which is the question behind our operator channel.
 
\subsection{Spectral Graph Filters and Their Stability}
Spectral graph convolution was introduced through the graph Fourier transform~\cite{bruna2014spectral}. Chebyshev polynomial filters~\cite{hammond2011wavelets,defferrard2016chebnet} avoid eigendecomposition and are localised to $K-1$ hops, and GCN~\cite{kipf2017gcn} is their first-order simplification. BernNet~\cite{he2021bernnet}, ChebNetII~\cite{he2022chebnetii} and the analysis of Wang and Zhang~\cite{wang2022powerful} study what polynomial filters can represent and how their coefficients should be constrained. ChebNetII shows that unconstrained Chebyshev coefficients behave poorly at high order, which bears on the coefficient norms we report. On stability, Gama et al.~\cite{gama2020stability} bound filter response under relative perturbations, Levie et al.~\cite{levie2021transferability} prove transferability of spectral filters across graphs, Kenlay et al.~\cite{kenlay2021interpretable} give bounds in terms of edge edits, Gao et al.~\cite{gao2021stochastic} treat stochastic perturbations, and Keriven et al.~\cite{keriven2020convergence} study large random graphs. Ruiz et al.~\cite{ruiz2021gnn} survey the area. CITRUS~\cite{einizade2024citrus} brings a perturbation analysis to spatio-temporal forecasting with continuous product-graph heat kernels. Our analysis uses standard tools. Its addition is a statement for the encoder we deploy, with a Chebyshev branch, a degree-normalised diffusion branch and interleaved attention, a tightness result, constants computed from trained weights, and a comparison of the bound with the measured response of the trained layer.
 
\subsection{Self-Supervised and Frequency-Aware Pre-Training}
GraphCL~\cite{you2020graphcl} and GRACE~\cite{zhu2020grace} learn graph representations from augmented views. For time series, TNC~\cite{tonekaboni2021tnc}, TS2Vec~\cite{yue2022ts2vec} and CoST~\cite{woo2022cost} use temporal contrast, and SimMTM~\cite{dong2023simmtm} reconstructs masked series in the time domain. STEP~\cite{shao2022step} and STD-MAE~\cite{gao2024stdmae} pre-train masked autoencoders on long traffic histories and pass the result to a downstream graph network. TF-C~\cite{zhang2022tfc} enforces consistency between time-based and frequency-based embeddings from two contrastive encoders. FreqMAE~\cite{kara2024freqmae} masks patches of a short-time Fourier spectrogram of IoT sensor input and reconstructs them with a frequency-weighted squared error on the raw values. FEDformer~\cite{zhou2022fedformer} and FreTS~\cite{yi2023frets} forecast in the frequency domain, wavelet-based financial forecasters~\cite{bohra2026wavefuse,bohra2026dstnet} build features from wavelet and filter-bank representations, and FreDF~\cite{wang2025fredf} adds an $\ell_1$ loss between the Fourier coefficients of the forecast and those of the label. None of these compares log spectra. MSR compares the log-PSD of the graph encoder's own output with the log-PSD of its input, which turns the error in every bin into a relative one. Log-spectral matching connects to the Itakura--Saito divergence~\cite{fevotte2009nmf}, made precise in Section~\ref{subsec:msr_is}. Table~\ref{tab:positioning} summarises the comparison.
 
\begin{table}[!t]
\centering
\caption{Positioning of ST-GFM. ``Cheb.\ Filter'' Means a Chebyshev Polynomial Graph Filter. CITRUS Uses Continuous Heat-Kernel Filters on a Product Graph. ``Bound'' Means a Stated Bound on How the Output Moves When the Graph Operator Is Perturbed.}
\label{tab:positioning}
\footnotesize
\setlength{\tabcolsep}{3.2pt}
\begin{tabular}{lccccc}
\toprule
Method & Cheb. & Learned & SSL & Freq. & Bound \\
       & filter & graph & & pretext & \\
\midrule
DCRNN~\cite{li2018dcrnn}                & \xmark & \xmark & \xmark & \xmark & \xmark \\
STGCN~\cite{yu2018stgcn}                & \cmark & \xmark & \xmark & \xmark & \xmark \\
ASTGCN~\cite{guo2019astgcn}             & \cmark & \xmark & \xmark & \xmark & \xmark \\
Graph WaveNet~\cite{wu2019gwnet}        & \xmark & \cmark & \xmark & \xmark & \xmark \\
MTGNN~\cite{wu2020mtgnn}                & \xmark & \cmark & \xmark & \xmark & \xmark \\
D$^2$STGNN~\cite{shao2022d2stgnn}       & \xmark & \cmark & \xmark & \xmark & \xmark \\
STEP~\cite{shao2022step}                & \xmark & \cmark & \cmark & \xmark & \xmark \\
STD-MAE~\cite{gao2024stdmae}            & \xmark & \cmark & \cmark & \xmark & \xmark \\
STAEformer~\cite{liu2023staeformer}     & \xmark & \xmark & \xmark & \xmark & \xmark \\
CITRUS~\cite{einizade2024citrus}        & \xmark & \xmark & \xmark & \xmark & \cmark \\
\textbf{ST-GFM (ours)}                  & \cmark & \cmark & \cmark & \cmark & \cmark \\
\bottomrule
\end{tabular}
\end{table}
 
\section{Problem Formulation and Spectral Shift}
\label{sec:problem}
 
\subsection{Graph Setting and Task}
A sensor network with $N$ nodes is described by a symmetric nonnegative adjacency $A\in\mathbb{R}_{\ge 0}^{N\times N}$ with positive degrees $D=\diag(A\mathbf{1})$. Road-distance kernels are asymmetric in general, and we symmetrise them as $A\leftarrow(A+A^\top)/2$. The symmetric normalised Laplacian
\begin{equation}
  L = I - D^{-1/2}AD^{-1/2}
  \label{eq:laplacian}
\end{equation}
is real symmetric with $\spec(L)\subset[0,2]$. Chebyshev filters act on a rescaled operator $\Lt=(2/\lambda_{\max})L-I$ with spectrum in $[-1,1]$. The learned operator of Section~\ref{subsec:graph} need not be symmetric, and the perturbation analysis of Section~\ref{sec:theory} covers that case.
 
At each forecast origin we observe $X\in\mathbb{R}^{N\times W}$, the last $W$ readings of every sensor, and predict $Y\in\mathbb{R}^{N\times H}$, the next $H$ readings. We write $\tilde{X}$ for the window with each sensor standardised by its training mean and standard deviation. A model is a map $f_\theta(X, L)$. In all experiments $W=H=12$ at five-minute resolution, so horizons $H\in\{3,6,12\}$ are 15, 30 and 60 minutes ahead.
 
\subsection{Spectral Shift}
\begin{definition}[Spectral shift]
\label{def:shift}
Let $L_{\mathrm{ref}}$ be a reference operator. For an input window $X_t$ with standardised form $\tilde{X}_t$, let $S(X_t)=|\mathcal{F}(\tilde{X}_t)|^2\in\mathbb{R}^{N\times\lfloor W/2\rfloor}$ be its per-node temporal PSD from the real FFT, restricted to the non-DC bins. Let $\bar{S}$ be the mean of $S$ over training windows. The spectral shift of a pair $(L_t, X_t)$ has an operator component and a frequency component,
\begin{equation}
  \delta_t^{\mathrm{top}} = \frac{\|L_t-L_{\mathrm{ref}}\|_2}{\|L_{\mathrm{ref}}\|_2},
  \qquad
  \delta_t^{\mathrm{freq}} = \frac{\|S(X_t)-\bar{S}\|_\F}{\|\bar{S}\|_\F}.
  \label{eq:shift}
\end{equation}
\end{definition}
 
We drop the DC bin because for a 12-step window it equals 144 times the squared window mean, which would turn $\delta^{\mathrm{freq}}$ into a measure of speed level rather than of frequency content. Sensors with a missing reading in the window are left out of both norms. A window with almost no non-DC energy, such as free-flowing traffic at night, has $\delta^{\mathrm{freq}}\approx1$. The measure detects departures in non-DC spectral power, including changes in fluctuation amplitude and in the distribution of energy across frequencies. It does not isolate spectral shape, because multiplying a standardised window by $c$ multiplies its PSD by $c^2$.
 
Both components are relative and dimensionless. The normalisation expresses both changes relative to their reference magnitudes. The symmetric normalised Laplacian of~\eqref{eq:laplacian} has spectral norm at most $2$, but that bound does not extend to the learned, non-symmetric operator. The operator component is measured on the operator rather than its spectrum, so it registers changes that leave the eigenvalues in place. For symmetric operators it also controls the spectrum. By Weyl's inequality every eigenvalue moves by at most $\|L_t-L_{\mathrm{ref}}\|_2$. By the Davis--Kahan theorem~\cite{davis1970rotation} in the form of Yu et al.~\cite{yu2015useful}, an $r$-dimensional eigenspace with eigengap $g$ rotates by at most $\|\sin\Theta\|_\F\le 2\min(\sqrt{r}\,\|L_t-L_{\mathrm{ref}}\|_2,\ \|L_t-L_{\mathrm{ref}}\|_\F)/g$. Neither result applies to the non-symmetric learned operator, and the bounds of Section~\ref{sec:theory} do not rely on them.
 
The two components differ in what can be observed. $\delta^{\mathrm{freq}}$ is computable from data, and we measure it on every test window. $\delta^{\mathrm{top}}$ needs the true sensor coupling of a window, which is not observed, so we use it to parameterise controlled perturbations and not to describe the data.
 
\subsection{How Each Channel Reaches the Model}
ST-GFM uses a learned anchor operator $L_\theta$ (Section~\ref{subsec:graph}), fixed once training ends. The operator channel acts when the encoder receives $L_\theta+\Delta L$ for any of the reasons listed in Section~\ref{sec:intro}. The frequency channel acts through $X$ alone. RQ3 perturbs the operator with the test inputs fixed, so only the operator component varies, and Section~\ref{sec:theory} bounds the response. RQ4 holds the operator at $L_\theta$, measures $\delta^{\mathrm{freq}}$ on the test data, and compares forecasters on the low-shift weekend windows with the full test set. RQ2 measures what MSR, the objective motivated by the frequency channel, contributes to accuracy.
\section{The ST-GFM Model}
\label{sec:model}
\begin{figure*}[!t]
\centering
\resizebox{\textwidth}{!}{%
\begin{tikzpicture}[
  font=\sffamily\footnotesize, >={Stealth[length=2.3mm,width=1.7mm]},
  panel/.style={rounded corners=7pt, draw=black!20, fill=black!2, line width=0.6pt},
  ptitle/.style={font=\sffamily\bfseries\footnotesize, text=black!80, anchor=north west, inner sep=0pt},
  blk/.style={rounded corners=4pt, draw, line width=0.7pt, align=center},
  lane/.style={font=\sffamily\bfseries\scriptsize, text=black!45, rotate=90, anchor=south},
  flow/.style={->, line width=0.95pt, black!72},
  gflow/.style={->, line width=0.9pt, violet!70!black, dash pattern=on 3pt off 2pt},
  sflow/.style={->, line width=0.9pt, purple!65!black, dash pattern=on 3pt off 2pt},
  tflow/.style={->, line width=0.8pt, red!60!black, densely dotted},
  tag/.style={font=\sffamily\scriptsize, text=black!55},
  nd/.style={circle, inner sep=0pt, minimum size=1.7mm, draw=#1!60!black, fill=#1!28, line width=0.5pt},
  nd/.default=blue,
  mgraph/.pic={
    \node[nd] (-a) at (-0.55,0.22) {}; \node[nd] (-b) at (-0.05,0.38) {};
    \node[nd] (-c) at (0.5,0.18) {};   \node[nd] (-d) at (-0.3,-0.28) {};
    \node[nd] (-e) at (0.3,-0.3) {};
    \foreach \u/\v in {a/b,b/c,a/d,d/e,b/e,c/e} \draw[blue!50!black, line width=0.55pt] (-\u)--(-\v);
  },
  attgrid/.pic={
    \foreach \i/\j/\o in {0/0/90,1/0/35,2/0/15,3/0/10, 0/1/30,1/1/85,2/1/40,3/1/15, 0/2/12,1/2/35,2/2/80,3/2/45, 0/3/10,1/3/15,2/3/40,3/3/90}
      \fill[orange!\o!white, draw=orange!50!black, line width=0.3pt] (\i*0.22-0.44, -\j*0.22+0.22) rectangle ++(0.22,0.22);
  },
]
 
\node[panel, minimum width=3.3cm, minimum height=3.7cm, anchor=north west] (pin) at (0,2.0) {};
\node[ptitle] at (0.18,1.85) {Input window};
\foreach \k in {2,1} \draw[fill=white, draw=blue!35] ({0.55+\k*0.14},{-0.45+\k*0.14}) rectangle ({2.6+\k*0.14},{1.1+\k*0.14});
\draw[fill=blue!4, draw=blue!50!black, line width=0.6pt] (0.55,-0.45) rectangle (2.6,1.1);
\foreach \y in {-0.06,0.33,0.72} \draw[blue!20] (0.55,\y)--(2.6,\y);
\draw[blue!60!black, line width=0.55pt] plot[smooth] coordinates {(0.62,0.92)(0.85,1.0)(1.05,0.85)(1.3,1.02)(1.55,0.8)(1.8,0.97)(2.05,0.86)(2.3,0.99)(2.52,0.9)};
\draw[blue!60!black, line width=0.55pt] plot[smooth] coordinates {(0.62,0.5)(0.85,0.45)(1.05,0.6)(1.3,0.42)(1.55,0.38)(1.8,0.62)(2.05,0.55)(2.3,0.45)(2.52,0.58)};
\draw[blue!60!black, line width=0.55pt] plot[smooth] coordinates {(0.62,0.15)(0.85,0.2)(1.05,0.12)(1.3,0.18)(1.55,0.26)(1.8,0.12)(2.05,0.05)(2.3,0.16)(2.52,0.2)};
\draw[blue!60!black, line width=0.55pt] plot[smooth] coordinates {(0.62,-0.3)(0.85,-0.18)(1.05,-0.28)(1.3,-0.35)(1.55,-0.2)(1.8,-0.24)(2.05,-0.32)(2.3,-0.2)(2.52,-0.27)};
\draw[<->, black!45, line width=0.5pt] (0.38,-0.45)--(0.38,1.1) node[midway, left, tag, inner sep=1pt]{$N$};
\draw[<->, black!45, line width=0.5pt] (0.55,-0.62)--(2.6,-0.62) node[midway, below, tag, inner sep=1.5pt]{12 steps};
\node[blk, draw=black!35, fill=gray!10, font=\sffamily\scriptsize, inner sep=2pt] at (1.0,-1.33) {time of day};
\node[blk, draw=black!35, fill=gray!10, font=\sffamily\scriptsize, inner sep=2pt] at (2.45,-1.33) {day of week};
 
\node[panel, minimum width=5.6cm, minimum height=3.7cm, anchor=north west] (pdec) at (3.75,2.0) {};
\node[ptitle] at (3.93,1.85) {Multi-scale decomposition};
\node[tag, anchor=north east] at (9.2,1.85) {time axis kept};
\foreach \r/\y/\nm/\fr/\pl in {1/0.92/s1/1050/1, 2/0.07/s2/480/2, 3/-0.78/s3/170/4}{
  \node[blk, draw=teal!55!black, fill=teal!10, minimum width=1.75cm, minimum height=0.62cm, font=\sffamily\scriptsize] (\nm) at (5.35,\y) {\textbf{stack \r}\\pool \pl};
  \draw[fill=white, draw=teal!40, rounded corners=2pt] (6.55,\y-0.29) rectangle (9.05,\y+0.29);
  \draw[teal!60!black, line width=0.65pt] plot[domain=0:2.3, samples=110, smooth] ({6.65+\x}, {\y+0.19*sin(\x*\fr)});
  \draw[->, teal!50!black, line width=0.6pt] (\nm.east) -- (6.55,\y);
}
\draw[->, black!50, line width=0.7pt] ([yshift=-1.6mm]s1.west) to[out=200,in=160] node[left, tag, inner sep=1pt]{residual} ([yshift=1.6mm]s2.west);
\draw[->, black!50, line width=0.7pt] ([yshift=-1.6mm]s2.west) to[out=200,in=160] node[left, tag, inner sep=1pt]{residual} ([yshift=1.6mm]s3.west);
\node[tag, align=center] at (7.8,-1.42) {conv1d + FiLM per stack};
 
\node[blk, draw=black!45, fill=gray!8, minimum width=2.0cm, minimum height=3.7cm, anchor=north west] (proj) at (9.75,2.0) {};
\node[font=\sffamily\bfseries\footnotesize] at (10.75,1.6) {Projection};
\node[tag] at (10.75,1.22) {192 + 48 $\to$ 320};
\node[blk, draw=orange!60!black, fill=orange!18, minimum width=1.6cm, minimum height=0.42cm, font=\sffamily\scriptsize] at (10.75,0.6) {node emb.};
\node[blk, draw=green!45!black, fill=green!14, minimum width=1.6cm, minimum height=0.42cm, font=\sffamily\scriptsize] at (10.75,0.0) {step emb.};
\node[blk, draw=blue!45!black, fill=blue!10, minimum width=1.6cm, minimum height=0.42cm, font=\sffamily\scriptsize] at (10.75,-0.6) {time emb.};
\node[tag] at (10.75,-1.2) {added to tokens};
 
\node[panel, minimum width=8.75cm, minimum height=3.7cm, anchor=north west, fill=violet!3, draw=violet!25] (pg) at (12.25,2.0) {};
\node[ptitle] at (12.43,1.85) {Learned anchor graph};
\begin{scope}[shift={(13.65,0.2)}]
  \node[nd=gray] (r1) at (-0.75,0.5) {}; \node[nd=gray] (r2) at (-0.1,0.75) {};
  \node[nd=gray] (r3) at (0.65,0.5) {};  \node[nd=gray] (r4) at (-0.55,-0.25) {};
  \node[nd=gray] (r5) at (0.2,-0.05) {}; \node[nd=gray] (r6) at (0.8,-0.45) {};
  \node[nd=gray] (r7) at (-0.15,-0.7) {};
  \foreach \u/\v in {r1/r2,r2/r3,r1/r4,r2/r5,r3/r6,r4/r7,r5/r6,r5/r7} \draw[gray!60!black, line width=0.55pt] (\u)--(\v);
  \node[tag] at (0,-1.2) {road distances};
\end{scope}
\node[font=\sffamily\bfseries\large, text=black!60] at (15.0,0.2) {$+$};
\node[tag] at (15.0,0.62) {gate $\alpha$};
\begin{scope}[shift={(15.55,-0.45)}]
  \foreach \i/\j/\o in {0/0/70,1/0/20,2/0/45,3/0/10,4/0/30, 0/1/25,1/1/80,2/1/15,3/1/55,4/1/12, 0/2/40,1/2/12,2/2/75,3/2/20,4/2/50, 0/3/10,1/3/50,2/3/25,3/3/85,4/3/18, 0/4/35,1/4/15,2/4/45,3/4/12,4/4/65}
    \fill[violet!\o!white, draw=violet!40!black, line width=0.25pt] (\i*0.26, \j*0.26) rectangle ++(0.26,0.26);
  \node[tag] at (0.65,-0.33) {learned $\mathrm{softmax}(E_1E_2^{\top})$};
\end{scope}
\node[blk, draw=violet!60!black, fill=violet!10, text width=2.95cm, minimum height=1.9cm] (ag) at (19.45,0.2)
  {\textbf{Anchor operator $\tilde{L}_\theta$}\\[2pt]\scriptsize renormalised Laplacian\\rescaled by power iteration\\not symmetric\\fixed after training};
\draw[->, violet!60!black, line width=0.8pt] (16.95,0.2) -- (ag.west);
 
\node[panel, minimum width=21.0cm, minimum height=3.55cm, anchor=north west, fill=blue!1.5] (penc) at (0,-2.5) {};
\node[lane] at (-0.08,-4.27) {ENCODER};
\node[blk, draw=blue!55!black, fill=blue!7, minimum width=3.8cm, minimum height=2.85cm, line width=0.9pt] (e1) at (2.55,-4.27) {};
\node[blk, draw=orange!65!black, fill=orange!7, minimum width=3.8cm, minimum height=2.85cm] (t1) at (7.55,-4.27) {};
\node[blk, draw=blue!55!black, fill=blue!7, minimum width=3.8cm, minimum height=2.85cm, line width=0.9pt] (e2) at (12.55,-4.27) {};
\node[blk, draw=orange!65!black, fill=orange!7, minimum width=3.8cm, minimum height=2.85cm] (t2) at (17.55,-4.27) {};
\node[font=\sffamily\bfseries\footnotesize] at (2.55,-3.12) {DCSE layer 1};
\node[font=\sffamily\bfseries\footnotesize] at (12.55,-3.12) {DCSE layer 2};
\node[font=\sffamily\bfseries\footnotesize] at (7.55,-3.12) {Temporal attention 1};
\node[font=\sffamily\bfseries\footnotesize] at (17.55,-3.12) {Temporal attention 2};
\pic (g1) at (2.55,-4.05) {mgraph};
\pic (g2) at (12.55,-4.05) {mgraph};
\pic at (7.55,-4.12) {attgrid};
\pic at (17.55,-4.12) {attgrid};
\foreach \x in {2.55,12.55}{
  \node[font=\sffamily\scriptsize, align=center] at (\x,-4.95) {Chebyshev + diffusion};
  \node[tag] at (\x,-5.35) {sensors mixed at each step};
}
\foreach \x in {7.55,17.55}{
  \node[font=\sffamily\scriptsize, align=center] at (\x,-4.95) {4 heads, pre-norm};
  \node[tag] at (\x,-5.35) {steps mixed for each sensor};
}
\draw[flow] (e1.east) to[out=25,in=155] node[above, tag]{reshape} (t1.west);
\draw[flow] (t1.east) to[out=25,in=155] node[above, tag]{reshape} (e2.west);
\draw[flow] (e2.east) to[out=25,in=155] node[above, tag]{reshape} (t2.west);
 
\draw[->, black!60, line width=0.7pt] (pin.east |- 0,0.92) -- (s1.west);
\foreach \y in {0.92,0.07,-0.78} \draw[->, teal!45!black, line width=0.6pt] (9.05,\y) to[out=0,in=180] (proj.west |- 0,0.07);
\draw[flow] (proj.south) .. controls (10.75,-2.2) and (2.55,-1.9) .. (e1.north);
\draw[gflow] (ag.south) .. controls (19.4,-1.75) and (14.6,-1.65) .. ([xshift=7mm]e2.north);
\draw[gflow] (ag.south) .. controls (19.0,-2.3) and (4.4,-1.35) .. ([xshift=7mm]e1.north);
\node[tag, text=violet!70!black, fill=white, inner sep=1.5pt] at (16.9,-2.08) {same $\tilde{L}_\theta$};
 
\node[lane] at (-0.08,-7.65) {GUARANTEES AND OUTPUT};
\node[blk, draw=red!50!black, fill=red!4, text width=9.6cm, align=left, anchor=north west, inner sep=6pt] (thy) at (0.2,-6.45)
  {\textbf{Operator-perturbation bounds} (Section~\ref{sec:theory})\\[2pt]
   \scriptsize
   $\bullet$ Chebyshev branch: change in output $\le\ c_{\mathrm{cheb}}\times$ change in operator, tight to first order\\[1pt]
   $\bullet$ $c_{\mathrm{cheb}}$ computed from the trained weights and the measured operator norm\\[1pt]
   $\bullet$ diffusion and stack bounds: linear on a bounded operator set with positive row sums};
\draw[tflow] ([xshift=-14mm]thy.north) to[out=90,in=-90] ([xshift=-6mm]e1.south);
\draw[tflow] ([xshift=30mm]thy.north) to[out=70,in=-110] ([xshift=-6mm]e2.south);
 
\node[blk, draw=black!45, fill=gray!8, minimum width=2.3cm, minimum height=1.1cm] (agg) at (11.6,-7.5) {\textbf{Aggregate}\\\scriptsize 12 steps $\to$ 1};
\node[blk, draw=black!45, fill=gray!8, minimum width=2.2cm, minimum height=1.1cm] (dec) at (14.35,-7.5) {\textbf{MLP decoder}\\\scriptsize + skip, degree emb.};
\node[circle, draw=black!65, line width=0.8pt, minimum size=0.55cm, inner sep=0pt, font=\sffamily\bfseries] (plus) at (16.25,-7.5) {$+$};
\node[blk, draw=black!35, fill=gray!5, minimum width=2.6cm, minimum height=0.6cm, font=\sffamily\scriptsize] (pers) at (14.35,-8.8) {persistence: last reading, learnable};
\draw[flow] (t2.south) .. controls (17.55,-6.6) and (11.6,-6.3) .. (agg.north);
\draw[flow] (agg) -- (dec);
\draw[flow] (dec) -- (plus);
\draw[flow] (pers.east) to[out=0,in=-90] (plus.south);
\node[panel, minimum width=4.05cm, minimum height=2.6cm, anchor=north west, fill=green!3, draw=green!35!black] (pf) at (16.95,-6.3) {};
\node[ptitle] at (17.1,-6.42) {Forecast, 12 steps};
\begin{scope}[shift={(17.2,-8.5)}]
  \draw[->, black!45] (0,0)--(3.5,0); \draw[->, black!45] (0,0)--(0,1.45);
  \fill[cOurs!15] (1.8,0.95) -- (2.3,0.72) -- (2.8,0.9) -- (3.3,0.7) -- (3.3,0.28) -- (2.8,0.5) -- (2.3,0.4) -- (1.8,0.75) -- cycle;
  \draw[black!75, line width=0.75pt] plot[smooth] coordinates {(0.1,0.9)(0.45,1.05)(0.8,0.75)(1.15,0.95)(1.5,0.7)(1.8,0.85)};
  \draw[cOurs, line width=0.9pt, dash pattern=on 3pt off 1.5pt] plot[smooth] coordinates {(1.8,0.85)(2.3,0.56)(2.8,0.7)(3.3,0.49)};
  \draw[black!40, densely dotted] (1.8,0)--(1.8,1.35) node[above, tag, inner sep=1pt]{now};
\end{scope}
\draw[flow] (plus.east) -- (pf.west |- plus);
 
\node[panel, minimum width=21.0cm, minimum height=4.15cm, anchor=north west, fill=purple!2, draw=purple!25] (pssl) at (0,-9.45) {};
\node[lane] at (-0.08,-11.52) {PRE-TRAINING};
\node[blk, draw=purple!50!black, fill=white, minimum width=6.6cm, minimum height=2.15cm, anchor=north west] (h1) at (0.25,-9.7) {};
\begin{scope}[shift={(1.35,-10.75)}]
  \node[nd] (m1) at (-0.65,0.32) {}; \node[nd=gray] (m2) at (0,0.5) {};
  \node[nd] (m3) at (0.6,0.25) {};   \node[nd] (m4) at (-0.4,-0.35) {};
  \node[nd=gray] (m5) at (0.35,-0.4) {};
  \foreach \u/\v in {m1/m2,m2/m3,m1/m4,m4/m5,m2/m5,m3/m5} \draw[blue!40!black!60, line width=0.5pt] (\u)--(\v);
  \node[font=\sffamily\bfseries\tiny, text=purple!70!black] at (0,0.5) {?};
  \node[font=\sffamily\bfseries\tiny, text=purple!70!black] at (0.35,-0.4) {?};
\end{scope}
\node[anchor=north west, text width=4.4cm, align=left] at (2.35,-9.85) {\textbf{Masked node reconstruction}\\[1pt]\scriptsize 30\% of sensors hidden at the input. Their decomposition features are recovered from the rest.};
\node[blk, draw=purple!60!black, fill=white, line width=0.95pt, minimum width=6.95cm, minimum height=2.15cm, anchor=north west] (h2) at (7.05,-9.7) {};
\begin{scope}[shift={(7.35,-11.55)}]
  \draw[->, black!45] (0,0)--(1.75,0) node[below left, tag, inner sep=1pt]{freq.};
  \draw[->, black!45] (0,0)--(0,1.4) node[above right, tag, inner sep=0pt]{log power};
  \draw[black!65, line width=0.75pt] plot[smooth] coordinates {(0.08,1.15)(0.35,0.95)(0.65,0.8)(0.95,0.55)(1.25,0.45)(1.6,0.3)};
  \draw[purple!70!black, line width=0.85pt, dash pattern=on 2.5pt off 1.5pt] plot[smooth] coordinates {(0.08,1.0)(0.35,1.02)(0.65,0.66)(0.95,0.62)(1.25,0.35)(1.6,0.38)};
  \foreach \x/\a/\b in {0.35/0.95/1.02, 0.95/0.55/0.62, 1.25/0.45/0.35} \draw[purple!60!black, line width=0.45pt] (\x,\a)--(\x,\b);
\end{scope}
\node[anchor=north west, text width=4.55cm, align=left] at (9.35,-9.85) {\textbf{Masked spectral reconstruction}\\[1pt]\scriptsize log-PSD of the encoder output (dashed) pulled onto the log-PSD of the input (solid). Relative error in every bin.};
\node[blk, draw=purple!50!black, fill=white, minimum width=6.6cm, minimum height=2.15cm, anchor=north west] (h3) at (14.2,-9.7) {};
\begin{scope}[shift={(14.5,-10.8)}]
  \draw[cOurs, line width=0.65pt] plot[smooth] coordinates {(0,0.45)(0.3,0.6)(0.6,0.4)(0.9,0.62)(1.2,0.45)};
  \draw[cOurs!60, line width=0.65pt] plot[smooth] coordinates {(0,-0.3)(0.3,-0.12)(0.6,-0.35)(0.9,-0.1)(1.2,-0.3)};
  \draw[<->, purple!60!black, line width=0.6pt] (1.35,0.45) to[out=-40,in=40] (1.35,-0.25);
  \node[tag] at (0.6,0.85) {view 1};
  \node[tag] at (0.6,-0.6) {view 2};
\end{scope}
\node[anchor=north west, text width=4.3cm, align=left] at (16.4,-9.85) {\textbf{Temporal contrast}\\[1pt]\scriptsize InfoNCE between two views made by scaling, shifting and noise.};
\fill[purple!32] (1.5,-12.72) rectangle (3.61,-12.32);
\shade[left color=purple!32, right color=gray!10] (3.61,-12.72) rectangle (5.72,-12.32);
\fill[gray!10] (5.72,-12.72) rectangle (20.5,-12.32);
\draw[black!40, line width=0.5pt] (1.5,-12.72) rectangle (20.5,-12.32);
\node[font=\sffamily\scriptsize] at (2.56,-12.52) {SSL only};
\node[font=\sffamily\scriptsize] at (4.67,-12.52) {SSL $\to$ 0};
\node[font=\sffamily\scriptsize] at (13.1,-12.52) {supervised forecasting with masked MAE};
\foreach \x/\l in {1.5/0, 3.61/20, 5.72/40, 20.5/180} {\draw[black!45] (\x,-12.72)--(\x,-12.82); \node[tag, below, inner sep=1pt] at (\x,-12.82) {\l};}
\node[tag, anchor=west] at (20.6,-12.52) {epoch};
\draw[sflow] (penc.east |- 0,-4.9) to[out=0,in=0, looseness=0.7] node[pos=0.5, right, tag, text=purple!70!black, align=left]{encoder\\output} ([yshift=-0.6cm]pssl.east |- 0,-9.45);
\end{tikzpicture}}%
\caption{Overview of ST-GFM. Row 1: time-preserving decomposition, embeddings and the learned anchor operator. Row 2: DCSE layers, which share one operator, alternate with per-node temporal attention. Row 3: the bounds of Section~\ref{sec:theory} and the forecasting head with its persistence term. Row 4: self-supervised branches, active for the first 40 epochs.}
\label{fig:overview}
\end{figure*}
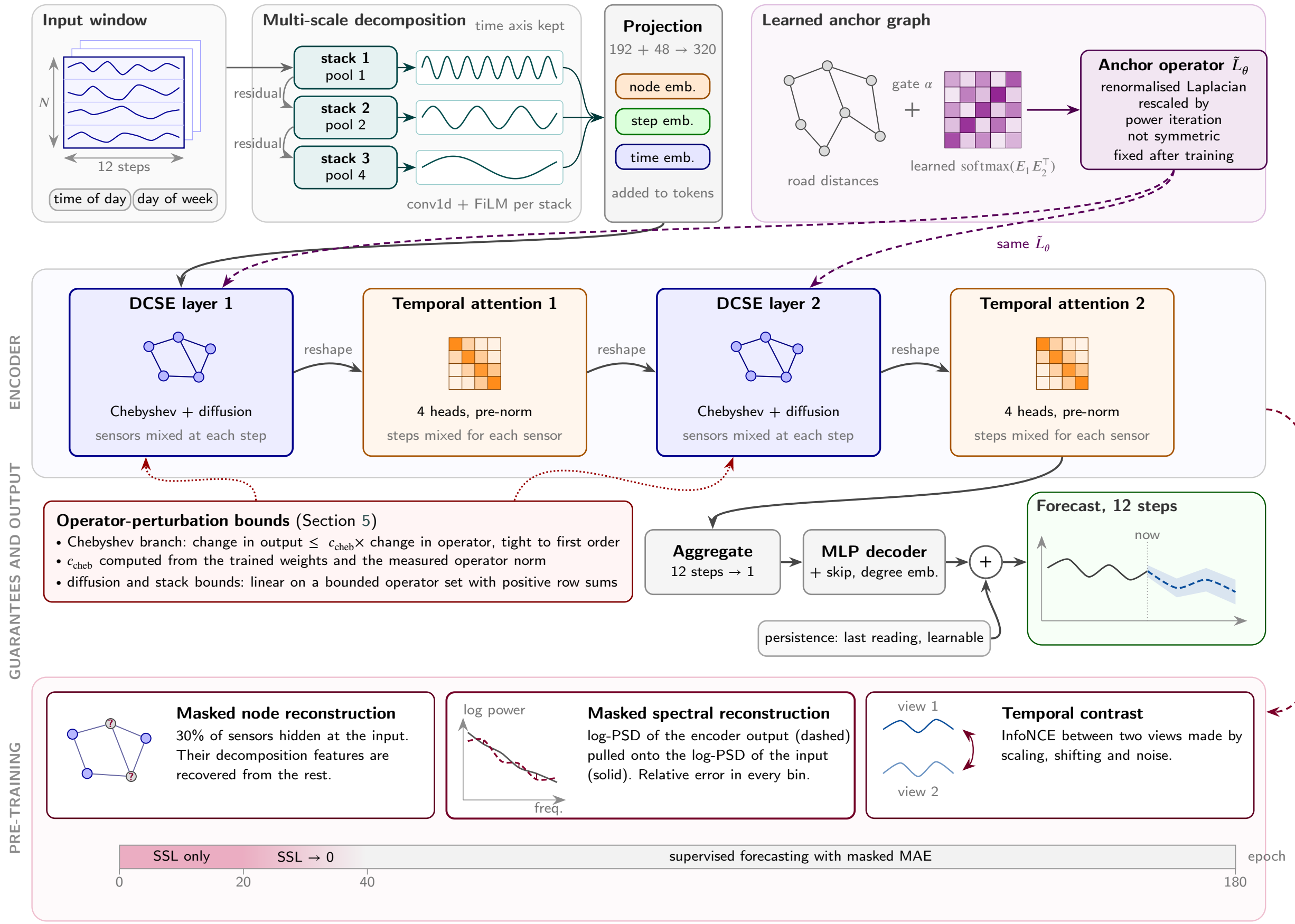

The full model of ST-GFM is shown in Figure~\ref{fig:overview}. As shown in this figure, inputs pass through a multi-scale decomposition that keeps all $W$ steps, are projected and given node, step and time embeddings, and enter an encoder of four layers that alternate DCSE graph filtering with per-node temporal attention. An aggregation layer and an MLP decoder produce the forecast, to which a learnable persistence term is added. During the first 40 epochs three self-supervised branches also train the encoder. The model has $5.22$ million parameters. Only the two node embeddings (the additive one and the one in the decoder) and the adaptive-graph vectors depend on $N$, and they change the count by less than one percent between the two speed networks.
 
\subsection{Input Embedding and Multi-Scale Decomposition}
\label{subsec:decomp}
 
The reading is fused with two scalar time features, time of day on a 288-bin grid and day of week on a 7-bin grid, both scaled to $[0,1]$, through a learned map $\mathrm{Linear}(3,1)$. The result is decomposed by $K_s=3$ stacks in the style of N-HiTS~\cite{challu2023nhits}. Stack $k$ applies max pooling with rate $r_k\in\{1,2,4\}$, a two-layer MLP with GELU~\cite{hendrycks2016gelu}, and a linear backcast to $W$ steps. The input to stack $k+1$ is the residual after the backcast of stack $k$ is subtracted, so successive stacks capture progressively slower content.
 
Standard N-HiTS collapses the time axis. Each stack applies a one-dimensional convolution with kernel size $3$ to its input and modulates the per-step features by the pooled summary through FiLM~\cite{perez2018film}. Stack $k$ returns $D_{\mathrm{dec}}^{(k)}\in\mathbb{R}^{N\times W\times F}$, and
\begin{equation}
  D_{\mathrm{dec}} = \big[D_{\mathrm{dec}}^{(1)}\,\|\,D_{\mathrm{dec}}^{(2)}\,\|\,D_{\mathrm{dec}}^{(3)}\big]\in\mathbb{R}^{N\times W\times FK_s},
  \label{eq:decomp}
\end{equation}
with $F=64$, so each step carries $192$ features. Keeping the time axis lets the graph layers that follow act on the traffic state at each step rather than on a shortened or summarised window (Section~\ref{sec:related}). If the steps were collapsed first, every graph filter would act on a temporal summary, and a filter applied to a summary cannot tell the onset of a queue at minute 50 from the state at minute 5 unless the summary has kept that difference. Collapsing the axis before the graph layers costs $0.12$ MAE at 60 minutes on METR-LA (Section~\ref{subsec:rq2}).
 
The decomposition is concatenated with a $48$-dimensional learned time embedding ($24$ dimensions each for time of day and day of week) and projected to width $d=320$ with $\mathrm{Linear}(FK_s+48,d)$ and GELU. Two learnable additive embeddings follow: a step embedding $P^{\mathrm{tmp}}\in\mathbb{R}^{W\times d}$ shared by all nodes and a node embedding $P^{\mathrm{nod}}\in\mathbb{R}^{N\times d}$, both initialised from $\mathcal{N}(0,0.02^2)$. These identity embeddings follow STAEformer~\cite{liu2023staeformer} and STID~\cite{shao2022stid} and reach every later layer through the residual connections.
 
\subsection{Learned Anchor Graph}
\label{subsec:graph}
 
The static adjacency $A_{\mathrm{static}}$ comes from road distances through a thresholded Gaussian kernel (Appendix~\ref{app:impl}). It is combined with a learned component~\cite{wu2019gwnet,bai2020agcrn},
\begin{equation}
  \begin{aligned}
  A_\theta &= (1-\alpha)\,A_{\mathrm{static}} + \alpha\,\mathrm{softmax}\big(\mathrm{ReLU}(E_1E_2^\top)\big),\\
  \alpha &= \sigma(\alpha_0),
  \end{aligned}
  \label{eq:adaptive_adj}
\end{equation}
where $E_1,E_2\in\mathbb{R}^{N\times 10}$ are learnable node embeddings and $\alpha_0$ is a learnable scalar. After training $\alpha$ is $0.45$ on both speed networks (Table~\ref{tab:diffconst}), so the static and learned parts carry similar weight. From $A_\theta$ we form the renormalised Laplacian with self-loops, $L_\theta=I-\hat{D}^{-1/2}(A_\theta+I)\hat{D}^{-1/2}$ with $\hat{D}=\diag((A_\theta+I)\mathbf{1})$, and rescale it as $\Lt_\theta=(2/\hat{\lambda})L_\theta-I$, where $\hat{\lambda}$ is the largest eigenvalue magnitude estimated by 20 power iterations. The softmax term is normalised by rows and $A_\theta$ is not symmetrised, so $\Lt_\theta$ is in general not symmetric. This rescaling is used as an operator-scaling convention. For a non-symmetric matrix, even an exact estimate of the largest eigenvalue magnitude does not guarantee that the eigenvalues of $\Lt_\theta$ lie in the unit disk or that $\|\Lt_\theta\|_2\le1$. The perturbation analysis therefore uses the measured spectral norm directly. On the trained models it is $1.30$ on METR-LA and $1.35$ on PEMS-BAY, and Section~\ref{subsec:cheb_bound} keeps this norm explicit. The operator is recomputed at each forward pass from the current parameters and is a fixed matrix after training. We write $L_\theta$ to separate it from any externally supplied operator in the perturbation experiments.
 
\subsection{Decoupled Chebyshev Spectral Encoder}
\label{subsec:dcse}
 
\begin{figure*}[!t]
\centering
\resizebox{\textwidth}{!}{%
\begin{tikzpicture}[
  font=\sffamily\footnotesize, >={Stealth[length=2.3mm,width=1.7mm]},
  blk/.style={rounded corners=4pt, draw, line width=0.7pt, align=center},
  op/.style={blk, draw=teal!55!black, fill=teal!10, minimum height=1.0cm, minimum width=1.5cm},
  flow/.style={->, line width=0.95pt, black!72},
  gflow/.style={->, line width=0.9pt, violet!70!black, dash pattern=on 3pt off 2pt},
  tag/.style={font=\sffamily\scriptsize, text=black!55},
  thm/.style={rounded corners=2pt, draw=red!55!black, fill=red!6, font=\sffamily\scriptsize\bfseries, text=red!60!black, inner sep=2pt},
  nd/.style={circle, inner sep=0pt, minimum size=2.3mm, draw=#1!60!black, fill=#1!25, line width=0.5pt},
  nd/.default=gray,
]
\foreach \k/\c in {0/blue!14, 1/blue!20, 2/blue!26, 3/blue!32}
  \fill[\c, draw=blue!50!black, line width=0.4pt, rounded corners=1pt] (0,{-0.55+\k*0.3}) rectangle (1.3,{-0.33+\k*0.3});
\node[font=\sffamily\bfseries\footnotesize] at (0.65,0.9) {Input};
\node[tag, align=center] at (0.65,-1.0) {$N\times d$ at\\one step};
\coordinate (sp) at (2.0,0);
\draw[line width=0.95pt, black!72] (1.3,0) -- (sp);
\fill[black!70] (sp) circle (1.3pt);
 
\node[blk, draw=violet!60!black, fill=violet!9, text width=4.3cm, minimum height=1.0cm] (lap) at (6.4,3.35)
  {\textbf{Anchor operator} $\tilde{L}_\theta$\\\scriptsize rescaled Laplacian of the learned graph};
 
\node[blk, draw=blue!55!black, fill=blue!6, minimum width=6.4cm, minimum height=2.05cm, line width=0.85pt] (cb) at (6.4,1.3) {};
\node[anchor=north west, text width=3.1cm, align=left] at (3.35,2.2) {\textbf{Chebyshev branch}\\[1pt]\scriptsize polynomial filter of orders 0, 1, 2\\one $d\times d$ weight per order\\isotropic, $K-1=2$ hops};
\begin{scope}[shift={(8.05,1.1)}]
  \draw[->, black!40, line width=0.4pt] (-0.95,0)--(0.98,0);
  \draw[->, black!40, line width=0.4pt] (0,-0.72)--(0,0.8);
  \draw[cTG, line width=0.8pt] (-0.85,0.6)--(0.85,0.6);
  \draw[cSTEP, line width=0.8pt] plot[domain=-1:1] ({0.85*\x},{0.6*\x});
  \draw[cOurs, line width=0.8pt] plot[domain=-1:1, samples=40] ({0.85*\x},{0.6*(2*\x*\x-1)});
  \node[font=\sffamily\tiny, text=cTG!80!black, anchor=south] at (-0.45,0.6) {$T_0$};
  \node[font=\sffamily\tiny, text=cSTEP!80!black, anchor=north west] at (0.62,0.22) {$T_1$};
  \node[font=\sffamily\tiny, text=cOurs, anchor=north west] at (0.05,-0.55) {$T_2$};
\end{scope}
\node[thm, anchor=north east] at ([xshift=-3pt,yshift=-3pt]cb.north east) {Theorem~\ref{thm:cheb}};
 
\node[blk, draw=orange!65!black, fill=orange!6, minimum width=6.4cm, minimum height=2.05cm] (df) at (6.4,-1.3) {};
\node[anchor=north west, text width=3.1cm, align=left] at (3.35,-0.4) {\textbf{Diffusion branch}\\[1pt]\scriptsize 2 propagation steps, forward and backward\\one weight per step and direction};
\begin{scope}[shift={(7.2,-1.4)}]
  \node[nd] (q1) at (-0.75,0.3) {}; \node[nd] (q2) at (0.0,0.55) {};
  \node[nd] (q3) at (0.8,0.2) {};   \node[nd] (q4) at (0.05,-0.45) {};
  \draw[black!35, line width=0.5pt] (q1)--(q2)--(q3)--(q4)--(q1) (q2)--(q4);
  \draw[->, cOurs, line width=0.7pt] (q1) to[bend left=35] (q2);
  \draw[->, cOurs, line width=0.7pt] (q2) to[bend left=35] (q3);
  \draw[->, cSTEP!90!black, line width=0.7pt] (q3) to[bend left=35] (q4);
  \draw[->, cSTEP!90!black, line width=0.7pt] (q4) to[bend left=35] (q1);
  \node[font=\sffamily\tiny, text=cOurs, anchor=west] at (1.0,0.4) {forward $P_f$};
  \node[font=\sffamily\tiny, text=cSTEP!80!black, anchor=west] at (1.0,-0.3) {backward $P_b$};
\end{scope}
\node[thm, anchor=north east] at ([xshift=-3pt,yshift=-3pt]df.north east) {Lemma~\ref{lem:diff}};
 
\draw[flow] (sp) to[out=90,in=180] (cb.west);
\draw[flow] (sp) to[out=-90,in=180] (df.west);
\draw[gflow] (lap.south) -- (cb.north);
\draw[gflow] (lap.east) .. controls (10.9,3.35) and (10.9,-0.95) .. ([yshift=3.5mm]df.east);
 
\node[circle, draw=teal!55!black, fill=teal!10, line width=0.8pt, minimum size=0.75cm, inner sep=0pt, font=\sffamily\bfseries] (cat) at (11.15,0) {$\Vert$};
\node[tag] at (11.15,-0.62) {concat};
\draw[flow] (cb.east) to[out=0,in=110] (cat.north west);
\draw[flow] ([yshift=-3.5mm]df.east) to[out=0,in=-110] (cat.south west);
\node[op] (lin) at (12.95,0) {Linear\\\scriptsize $2d\to d$};
\node[op] (gel) at (14.85,0) {GELU};
\node[circle, draw=black!65, line width=0.9pt, minimum size=0.62cm, inner sep=0pt, font=\sffamily\bfseries] (add) at (16.45,0) {$+$};
\node[op] (ln) at (18.05,0) {Layer\\norm};
\foreach \k/\c in {0/teal!14, 1/teal!20, 2/teal!26, 3/teal!32}
  \fill[\c, draw=teal!50!black, line width=0.4pt, rounded corners=1pt] (19.35,{-0.55+\k*0.3}) rectangle (20.65,{-0.33+\k*0.3});
\node[font=\sffamily\bfseries\footnotesize] at (20.0,0.9) {Output};
\node[tag, align=center] at (20.0,-1.0) {$N\times d$ at\\one step};
\draw[flow] (cat) -- (lin);
\draw[flow] (lin) -- (gel);
\draw[flow] (gel) -- (add);
\draw[flow] (add) -- (ln);
\draw[flow] (ln) -- (19.35,0);
\draw[->, line width=0.85pt, black!50, rounded corners=14pt] (sp) -- (2.0,-2.85) -- node[pos=0.5, below, tag]{identity path} (16.45,-2.85) -- (add.south);
\end{tikzpicture}}%
\caption{One DCSE layer, applied to the $N\times d$ state at one time step. Both graph branches read the anchor operator $\tilde{L}_\theta$. The Chebyshev branch applies a polynomial filter of orders 0 to 2, and the diffusion branch takes two propagation steps in each direction. Their outputs are concatenated, mixed by a linear map and added to the input before layer normalisation. The red tags mark the result in Section~\ref{sec:theory} that bounds each branch.}
\label{fig:dcse}
\end{figure*}
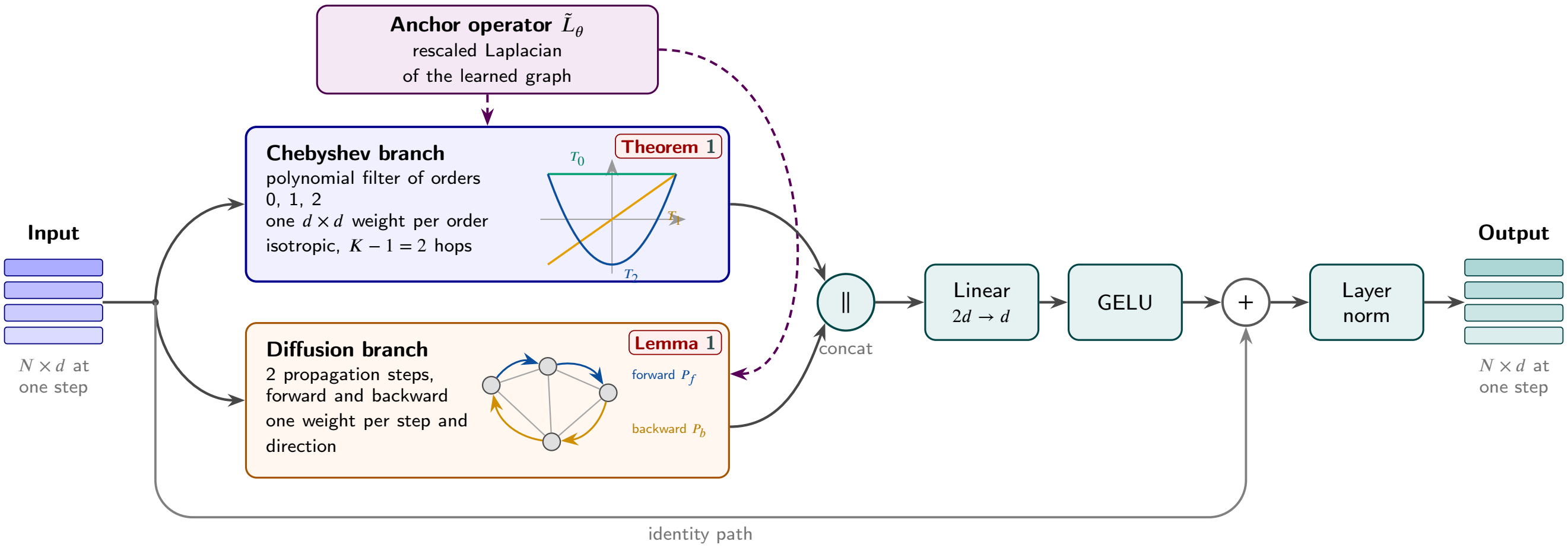

The encoder treats time as a batch axis. The tensor $Z\in\mathbb{R}^{B\times N\times W\times d}$ is reshaped to $(B\cdot W)\times N\times d$, so the graph filter at step $w$ acts on the state at step $w$. A DCSE layer (shown in Figure~\ref{fig:dcse}) has two branches. The Chebyshev branch computes
\begin{equation}
  Z_{\mathrm{cheb}} = \sum_{k=0}^{K-1} T_k(\Lt_\theta)\,Z\,\Theta_k,
  \label{eq:cheb}
\end{equation}
with $K=3$, Chebyshev polynomials $T_0(x)=1$, $T_1(x)=x$, $T_k(x)=2xT_{k-1}(x)-T_{k-2}(x)$ evaluated on $\Lt_\theta$, and trainable $\Theta_k\in\mathbb{R}^{d\times d}$~\cite{defferrard2016chebnet}. The diffusion branch runs $K_{\mathrm{diff}}=2$ propagation steps in two directions,
\begin{equation}
\begin{aligned}
  Z_{\mathrm{diff}} &= \sum_{s=1}^{K_{\mathrm{diff}}}\big(P_f S_s W_{f,s} + P_b S_s W_{b,s}\big),\\
  S_1 &= Z,\qquad S_{s+1}=(P_f+P_b)\,S_s,\\
  P_f &=\hat{A}\,\diag(\hat{A}\mathbf{1})^{-1},\quad P_b=P_f^\top,\quad \hat{A}=I-\tfrac12\Lt_\theta,
\end{aligned}
\label{eq:diff}
\end{equation}
with separate weights per step and direction. The state carried to the next step is the sum of the forward and backward propagations. $P_f$ divides each column of $\hat{A}$ by the matching row sum of $\hat{A}$, and $P_b$ is its transpose. The static distance kernel is symmetric, so neither direction encodes the direction of travel. The two differ through the learned, non-symmetric part of $A_\theta$ and through degree normalisation. The Chebyshev branch applies one isotropic polynomial filter of reach two hops, while the diffusion branch can weight the two directions and the two hops separately. The branches are merged in a residual block,
\begin{equation}
  Z' = \LN\Big(\GELU\big(W_{\mathrm{merge}}[Z_{\mathrm{cheb}}\,\|\,Z_{\mathrm{diff}}]\big) + Z\Big),
  \label{eq:dcse}
\end{equation}
where $W_{\mathrm{merge}}$ maps $2d$ to $d$ and $\LN$ is layer normalisation~\cite{ba2016layernorm}. The learned part of $A_\theta$ is dense, so one multiplication by $\Lt_\theta$ or $P_f$ costs $O(N^2d)$ and not the $O(|\mathcal{E}|d)$ of a sparse graph. At $N\le325$ this is small on one GPU. Larger networks would need top-$k$ sparsification of the learned term.
 
\subsection{Per-Node Temporal Attention, Aggregation and Decoder}
\label{subsec:temporal}
 
Between DCSE layers the tensor is reshaped to $(B\cdot N)\times W\times d$, so each sensor attends only over its own $W=12$ steps. The block uses pre-layer-norm multi-head self-attention~\cite{vaswani2017attention} with four heads and dropout $0.1$, a residual connection, and a two-layer feed-forward network of width $2d$ with GELU. No positional encoding is added inside the block, because $P^{\mathrm{tmp}}$ already reaches it through the residual stream. The encoder is
\begin{equation}
  \mathrm{DCSE}_1 \rightarrow \mathrm{TempAttn}_1 \rightarrow \mathrm{DCSE}_2 \rightarrow \mathrm{TempAttn}_2 .
  \label{eq:stack}
\end{equation}
 
The encoder output is reshaped to $B\times N\times W\times d$. At each node the $W\cdot d$ features are flattened and mapped to $d$ by one linear layer, so every step contributes to every horizon. A gated skip from the encoder input $Z_0$ is added, $Z_{\mathrm{out}}=Z_{\mathrm{agg}}+\sigma(g)\,W_{\mathrm{skip}}\,\mathrm{mean}_W(Z_0)$, with the scalar $g$ initialised at $0$. Following T-Graphormer~\cite{bai2025tgraphormer}, a degree embedding indexed by the number of nonzero entries in each row of $A_{\mathrm{static}}$, clamped at $60$, is added and initialised at zero. A $32$-dimensional node embedding is concatenated and a two-layer MLP decoder produces $H$ outputs per node. The forecast adds a persistence term $W_{\mathrm{base}}X$ with $W_{\mathrm{base}}\in\mathbb{R}^{W\times H}$ initialised to copy the last observed reading to every horizon, so the decoder learns the departure from persistence.
 
\subsection{Self-Supervised Objectives and Masked Spectral Reconstruction}
\label{subsec:ssl}
 
Three self-supervised branches act on the encoder during the first $40$ epochs.
 
\subsubsection{Masked Node Reconstruction}
A fraction $q_{\mathrm{m}}=0.3$ of nodes is zeroed at the encoder input. A two-layer MLP decoder reconstructs, for each masked node $i$, the decomposition features $\bar{D}_{\mathrm{dec},i}\in\mathbb{R}^{FK_s}$ averaged over three randomly drawn steps:
\begin{equation}
  \mathcal{L}_{\mathrm{spatial}} = \frac{1}{|\mathcal{M}|}\sum_{i\in\mathcal{M}}\big\|\mathrm{Dec}(z_i)-\bar{D}_{\mathrm{dec},i}\big\|_2^2,
  \label{eq:spatial}
\end{equation}
where $\mathcal{M}$ is the masked set and $z_i\in\mathbb{R}^d$ is the encoder summary of node $i$.
 
\subsubsection{Masked Spectral Reconstruction}
MSR uses the same masked input. Let $\bar{Z}\in\mathbb{R}^{N\times W}$ be the encoder output averaged over the feature axis, and let $\ell_Z=\log(|\mathcal{F}(\bar{Z})|^2+\epsilon)$ and $\ell_X=\log(|\mathcal{F}(\tilde{X})|^2+\epsilon)$ be the log-PSDs of the encoder output and of the standardised input window $\tilde{X}$ before decomposition, both along time with the real FFT and $\epsilon=10^{-8}$. The loss is
\begin{equation}
  \mathcal{L}_{\mathrm{freq}} = \frac{1}{NW'}\left\|\frac{\ell_Z-\ell_X}{s_X}\right\|_\F^2,
  \qquad
  s_X = \mathrm{sg}\big(\langle|\ell_X|\rangle\big),
  \label{eq:msr}
\end{equation}
where $W'=\lfloor W/2\rfloor+1=7$ is the number of frequency bins, $\langle\cdot\rangle$ is the mean over nodes and bins and $\mathrm{sg}$ stops the gradient. The loss averages over all $N$ nodes, so it covers the $30\%$ of sensors hidden at the input as well as the visible ones. The scale $s_X$ keeps gradients stable when the input spectrum spans several orders of magnitude. For half-precision safety, $\ell_Z$ is clamped to $[-8,8]$ before division and both normalised log-PSDs are clamped to $[-10,10]$. The target is the input $\tilde{X}$ and not the decomposition $D_{\mathrm{dec}}$, so MSR asks the encoder to keep the spectral profile of the observed series while the masked branch asks it to fill in hidden nodes.
 
Three properties bound what MSR can enforce. The logarithm weights relative power error in every bin (Section~\ref{subsec:msr_is}). The PSD discards phase, so MSR constrains how energy is spread over frequency but not when events happen. And the loss acts on one readout of the encoder output, its feature mean. This readout is taken after the final temporal-attention block, whose residual updates can produce feature means that vary across nodes and time steps. Using a parameter-free mean readout applies spectral supervision directly to the encoder representation without introducing an additional learned projection head. Unlike $\delta^{\mathrm{freq}}$, MSR keeps the DC bin, so it also asks the readout to track the window mean.
 
\subsubsection{Temporal Contrast}
Two views are made by random amplitude scaling within $\pm10\%$, a temporal shift of up to two steps and Gaussian noise with standard deviation $0.02$. A symmetric InfoNCE loss with cosine similarity and temperature $\tau=0.07$ is computed on up to $4096$ sampled node embeddings~\cite{yue2022ts2vec,tonekaboni2021tnc}.
 
\subsubsection{Training Schedule}
The self-supervised objective is
\begin{equation}
  \mathcal{L}_{\mathrm{SSL}} = \mathcal{L}_{\mathrm{spatial}} + \lambda_1\mathcal{L}_{\mathrm{freq}} + \lambda_2\mathcal{L}_{\mathrm{temp}},
  \quad \lambda_1=0.5,\ \lambda_2=0.4 .
  \label{eq:ssl}
\end{equation}
The first $E_w=20$ epochs use $\mathcal{L}_{\mathrm{SSL}}$ alone. From epoch $e>E_w$ the objective is $w(e)\,\mathcal{L}_{\mathrm{SSL}}+\mathcal{L}_{\mathrm{sup}}$ with
\begin{equation}
  w(e) = \max\!\Big(0,\ 1-\frac{e-E_w-1}{19}\Big),
  \label{eq:schedule}
\end{equation}
which reaches zero at epoch $40$. Training is purely supervised afterwards. $\mathcal{L}_{\mathrm{sup}}$ is the MAE over valid readings (Section~\ref{sec:setup}). Supplementary Algorithm~S1 lists the procedure.
 
\section{Stability Analysis}
\label{sec:theory}
 
This section bounds how far the encoder output moves when the operator moves from $L$ to $L'=L+\Delta L$ with the input fixed. Section~\ref{subsec:cheb_bound} treats the Chebyshev branch, Section~\ref{subsec:diff_bound} the diffusion branch, and Section~\ref{subsec:stack_bound} the stack. All bounds use the Frobenius norm, which matches the perturbation budgets of Section~\ref{subsec:rq3}. Proofs not in the text are in Appendix~\ref{app:proofs}.
 
\subsection{The Chebyshev Branch}
\label{subsec:cheb_bound}
 
The bound needs only the recurrence of the Chebyshev polynomials and the inequality $\|AB\|_\F\le\|A\|_2\|B\|_\F$, so it holds for any square operator, symmetric or not. This matters because the learned operator $\Lt_\theta$ is not symmetric. The argument is elementary. What it adds is a constant that can be evaluated on the trained weights and checked against the measured response of the layer, which Section~\ref{subsec:rq3} does.
 
\begin{theorem}[Chebyshev branch, $K=3$]
\label{thm:cheb}
Let $\Lt\in\mathbb{R}^{N\times N}$ be any matrix with $\rho=\|\Lt\|_2$, let $\Delta L$ be a perturbation with $\eta=\|\Delta L\|_2$, and let $g(\Lt)=\sum_{k=0}^{2}T_k(\Lt)\,Z\,\Theta_k$. Then
\begin{equation}
  \begin{aligned}
  \|g(\Lt+\Delta L)-g(\Lt)\|_\F &\le \|Z\|_2\,\|\Delta L\|_\F\\
  &\quad\times\big(\|\Theta_1\|_2+2(2\rho+\eta)\|\Theta_2\|_2\big).
  \end{aligned}
  \label{eq:cheb_bound}
\end{equation}
\end{theorem}
\begin{proof}
The $k=0$ term cancels. $T_1(\Lt+\Delta L)-T_1(\Lt)=\Delta L$, and $T_2(\Lt+\Delta L)-T_2(\Lt)=2(\Lt\,\Delta L+\Delta L\,\Lt+\Delta L^2)$, whose Frobenius norm is at most $2(2\rho+\eta)\|\Delta L\|_\F$. Since $\|(T_k(\Lt+\Delta L)-T_k(\Lt))Z\Theta_k\|_\F\le\|T_k(\Lt+\Delta L)-T_k(\Lt)\|_\F\,\|Z\|_2\,\|\Theta_k\|_2$, summing over $k=1,2$ gives~\eqref{eq:cheb_bound}.
\end{proof}
 
 
\begin{proposition}[Tightness]
\label{prop:tight}
Let $L$ be a symmetric normalised Laplacian with eigenvalue $2$, as when the graph has a bipartite component, with unit eigenvector $u$, and let $\Lt=L-I$, so that $\rho=1$. Take $Z=uv^\top$ for a unit vector $v$, $\Theta_k=M I_d$ for $k=0,1,2$, and $\Delta L=-\varepsilon\,uu^\top$ with $0<\varepsilon<2$. Then
\[
  \|g(\Lt+\Delta L)-g(\Lt)\|_\F = M(5\varepsilon-2\varepsilon^2),
\]
while the right-hand side of~\eqref{eq:cheb_bound} equals $M(5\varepsilon+2\varepsilon^2)$. The bound is attained to first order in $\varepsilon$.
\end{proposition}
\begin{proof}
$u$ is an eigenvector of $\Lt$ with eigenvalue $1$ and of $\Lt+\Delta L$ with eigenvalue $1-\varepsilon$. Hence $g(\Lt+\Delta L)-g(\Lt)=M\big[(T_1(1-\varepsilon)-1)+(T_2(1-\varepsilon)-1)\big]uv^\top=M(2\varepsilon^2-5\varepsilon)\,uv^\top$, and $\|uv^\top\|_\F=1$. On the right-hand side $\|Z\|_2=1$, $\rho=1$ and $\eta=\|\Delta L\|_\F=\varepsilon$.
\end{proof}
 
For general $K$, symmetric operators and perturbations that keep the spectrum in $[-1,1]$, Markov's inequality $|T_k'|\le k^2$ on $[-1,1]$~\cite{mason2002chebyshev} gives a constant that grows as $\sum_{k<K}k^2\sim K^3/3$, which argues for a small polynomial order. In relative form, Theorem~\ref{thm:cheb} reads
\begin{equation}
  \begin{aligned}
  \frac{\|g(\Lt+\Delta L)-g(\Lt)\|_\F}{\|Z\|_2} &\le c_{\mathrm{cheb}}(\rho,\eta)\,\|\Delta L\|_\F,\\
  c_{\mathrm{cheb}}(\rho,\eta) &= \|\Theta_1\|_2 + 2(2\rho+\eta)\|\Theta_2\|_2 .
  \end{aligned}
  \label{eq:ccheb}
\end{equation}
 
Table~\ref{tab:coeffs} evaluates the coefficient norms in the first DCSE layer of every trained model and reports the unit-norm constant $c_{\mathrm{cheb}}(1,0)=\|\Theta_1\|_2+4\|\Theta_2\|_2$. Across runs it is $14.49\pm0.55$ on METR-LA and $16.80\pm1.20$ on PEMS-BAY, $22\%$ and $24\%$ below the value it would take if $\Theta_1$ and $\Theta_2$ had the norm of $\Theta_0$. In every run $\Theta_0$ has the largest norm. The order of the other two is not fixed: in PEMS-BAY seed 101, $\|\Theta_2\|_2=3.59$ exceeds $\|\Theta_1\|_2=3.43$. Three coefficients cannot support a decay law, so we report the norms as measured.
 
The trained operator is not of unit norm. Each unit of $\rho$ above one adds $4\|\Theta_2\|_2$ to the constant, about $11$ on METR-LA and $13$ on PEMS-BAY. Measured on the trained operators, $\|\Lt_\theta\|_2$ is $1.30$ on METR-LA and $1.35$ on PEMS-BAY (Table~\ref{tab:diffconst}), which gives $c_{\mathrm{cheb}}(\rho,0)=17.93$ and $21.48$, about $24\%$ and $28\%$ above the unit-norm values. These are the constants of the deployed layer, and Section~\ref{subsec:rq3} compares the bound they give with the measured layer response. The constant grows linearly in $\eta$ with slope $2\|\Theta_2\|_2$, about $5.7$ on METR-LA and $6.7$ on PEMS-BAY. Supplementary Figure~S1 plots it against $\eta$ at both operator norms.
 
\begin{table}[!t]
\centering
\caption{Spectral Norms of the Trained Chebyshev Coefficients in the First DCSE Layer of Every Run, and the Unit-Norm Constant $c_{\mathrm{cheb}}(1,0)=\|\Theta_1\|_2+4\|\Theta_2\|_2$ of~\eqref{eq:ccheb}, Computed From Unrounded Norms.}
\label{tab:coeffs}
\footnotesize
\setlength{\tabcolsep}{4pt}
\begin{tabular}{lcccc}
\toprule
& $\|\Theta_0\|_2$ & $\|\Theta_1\|_2$ & $\|\Theta_2\|_2$ & $c_{\mathrm{cheb}}(1,0)$ \\
\midrule
\multicolumn{5}{l}{\textbf{METR-LA}}\\
Seed 13 & 4.133 & 3.005 & 2.958 & 14.84 \\
Seed 42 & 4.004 & 2.866 & 2.730 & 13.79 \\
Seed 101 & 3.535 & 3.153 & 2.900 & 14.75 \\
Seed 202 & 3.344 & 3.170 & 2.971 & 15.05 \\
Seed 303 & 3.608 & 2.977 & 2.766 & 14.04 \\
Mean & 3.725 & 3.034 & 2.865 & 14.49 \\
Std.\ deviation & 0.331 & 0.127 & 0.111 & 0.55 \\
\midrule
\multicolumn{5}{l}{\textbf{PEMS-BAY}}\\
Seed 42 & 4.026 & 3.267 & 3.052 & 15.47 \\
Seed 101 & 4.869 & 3.429 & 3.592 & 17.80 \\
Seed 303 & 4.405 & 3.589 & 3.385 & 17.13 \\
Mean & 4.433 & 3.428 & 3.343 & 16.80 \\
Std.\ deviation & 0.422 & 0.161 & 0.272 & 1.20 \\
\bottomrule
\end{tabular}
\end{table}
 
\subsection{The Diffusion Branch}
\label{subsec:diff_bound}
 
The diffusion branch depends on the operator through $\hat{A}$ and through the degree normalisation, which couples all entries of a row.
 
\begin{lemma}[Diffusion branch]
\label{lem:diff}
Let $\Lt$ and $\Lt'=\Lt+\Delta L$ be two operators such that $\hat{A}=I-\tfrac12\Lt$ and $\hat{A}'=I-\tfrac12\Lt'$ have row sums of at least $m>0$ and spectral norms of at most $a$. Let $s_\Delta$ be the largest number of nonzero entries in a row of $\Delta L$. For the branch in~\eqref{eq:diff},
\begin{equation}
  \|g_{\mathrm{diff}}(\Lt')-g_{\mathrm{diff}}(\Lt)\|_\F \le \|Z\|_2\,c_{\mathrm{diff}}\,\frac{1}{2m}\Big(1+\frac{a\sqrt{s_\Delta}}{m}\Big)\|\Delta L\|_\F,
  \label{eq:diff_bound}
\end{equation}
where $c_{\mathrm{diff}}=\sum_{s'=1}^{K_{\mathrm{diff}}} s'\,(2a/m)^{s'-1}\big(\|W_{f,s'}\|_2+\|W_{b,s'}\|_2\big)$.
\end{lemma}
 
The factor $\sqrt{s_\Delta}$ comes from the degree normalisation. It counts the nonzero entries in a row of the operator perturbation, not of the adjacency edit that produced it. Removing a few edges changes the degrees at their endpoints, and the normalisation and the rescaling by $\hat{\lambda}$ then spread the change over whole rows of $\Lt_\theta$. The learned part of $A_\theta$ is dense, so for most edits to the graph of ST-GFM $s_\Delta$ is close to $N$ and the bound is looser by about $\sqrt{N}$ than for a perturbation with few entries per row.
 
For $K_{\mathrm{diff}}=2$ the constant is $c_{\mathrm{diff}}=w_1+(4a/m)\,w_2$ with $w_s=\|W_{f,s}\|_2+\|W_{b,s}\|_2$. For $\Lt=L-I$, the rescaling of a symmetric normalised Laplacian with $\lambda_{\max}=2$, the matrix $\hat{A}=\tfrac32I-\tfrac12L$ has spectral norm at most $\tfrac32$ and row sums $1+\tfrac12\sum_jA_{ij}/\sqrt{d_id_j}\ge1$, with $d_i$ the degrees of $A$, so $4a/m\le6$. The trained operator is not of this form, and we measure $a$ and $m$ on it directly. Table~\ref{tab:diffconst} lists the result. The row sums of $\hat{A}$ are at least $1.38$ on METR-LA and $1.33$ on PEMS-BAY, and its spectral norm is $1.54$ and $1.55$, slightly above the symmetric value of $\tfrac32$. This gives $4a/m=4.46$ and $4.66$, below the symmetric worst case, and $c_{\mathrm{diff}}=27.4$ on METR-LA and $32.0$ on PEMS-BAY. These values hold at $\Lt_\theta$. Lemma~\ref{lem:diff} needs $m$ and $a$ at both operators, so a finite edit uses $\min(m,m')$ and $\max(a,a')$, and we do not instantiate the diffusion bound under edge dropout. The second propagation step accounts for about four fifths of the constant on both networks. For a perturbation that fills whole rows, $s_\Delta=N$, the full normalisation prefactor $\frac{1}{2m}(1+\frac{a\sqrt{s_\Delta}}{m})$ in~\eqref{eq:diff_bound} is about $6.2$ on METR-LA and $8.3$ on PEMS-BAY, so the diffusion bound per unit $\|\Delta L\|_\F$ is about $170$ and $265$. That is roughly ten times the Chebyshev constant, and the $\sqrt{s_\Delta}$ factor accounts for most of the gap.
 
\begin{table}[!t]
\centering
\caption{Operator and Branch Constants in the First DCSE Layer, Mean Over Runs. $c_{\mathrm{cheb}}(\rho,0)=\|\Theta_1\|_2+4\rho\|\Theta_2\|_2$ and $c_{\mathrm{diff}}=w_1+(4a/m)\,w_2$ With $w_s=\|W_{f,s}\|_2+\|W_{b,s}\|_2$, $a=\|\hat{A}\|_2$ and $m=\min_i(\hat{A}\mathbf{1})_i$, as in Lemma~\ref{lem:diff} for $K_{\mathrm{diff}}=2$.}
\label{tab:diffconst}
\footnotesize
\setlength{\tabcolsep}{4pt}
\begin{tabular}{lcc}
\toprule
& METR-LA & PEMS-BAY \\
\midrule
$\|\Lt_\theta\|_2$                       & 1.30  & 1.35  \\
$c_{\mathrm{cheb}}(\rho,0)$             & 17.93 & 21.48 \\
Learned-graph gate $\alpha$              & 0.45  & 0.45  \\
$a=\|\hat{A}\|_2$                        & 1.54  & 1.55  \\
$m=\min_i(\hat{A}\mathbf{1})_i$          & 1.38  & 1.33  \\
$4a/m$                                   & 4.46  & 4.66  \\
$w_1=\|W_{f,1}\|_2+\|W_{b,1}\|_2$        & 6.0   & 6.8   \\
$w_2=\|W_{f,2}\|_2+\|W_{b,2}\|_2$        & 4.8   & 5.4   \\
$c_{\mathrm{diff}}$                      & 27.4  & 32.0  \\
\bottomrule
\end{tabular}
\end{table}
 
\subsection{The Encoder Stack}
\label{subsec:stack_bound}
 
A DCSE layer $F_\ell(Z,L)$ depends on its input and the operator. A temporal block $T_\ell(Z)$ depends only on its input.
 
\begin{proposition}[Encoder stack]
\label{prop:stack}
Let $\mathcal{U}$ be a compact set of rescaled operators such that every row sum of $I-\tfrac12L$ is at least $m_\star>0$ for every $L\in\mathcal{U}$. Assume the layer inputs reachable under these operators belong to a bounded set. Suppose that uniformly on these sets each DCSE layer satisfies $\|F_\ell(Z',L')-F_\ell(Z,L)\|_\F\le \kappa_Z\|Z'-Z\|_\F+\kappa_L\|L'-L\|_\F$ and each temporal block satisfies $\|T_\ell(Z')-T_\ell(Z)\|_\F\le \kappa_T\|Z'-Z\|_\F$. Then, for a fixed input and any $L,L'\in\mathcal{U}$, the encoder~\eqref{eq:stack} satisfies
\begin{equation}
  \|\mathrm{Enc}(L')-\mathrm{Enc}(L)\|_\F \le \kappa_T\,\kappa_L\,(1+\kappa_Z\,\kappa_T)\,\|\Delta L\|_\F .
  \label{eq:stack_bound}
\end{equation}
\end{proposition}
 
These assumptions give finite uniform constants. Compactness of $\mathcal{U}$ bounds the operator norms and perturbation sizes, while $m_\star$ keeps the diffusion normalisation away from zero. For $\kappa_L$, Theorem~\ref{thm:cheb} and Lemma~\ref{lem:diff} bound the two branches, the merge contributes $\|W_{\mathrm{merge}}\|_2$, GELU is $1.13$-Lipschitz~\cite{hendrycks2016gelu}, and layer normalisation with gain $\gamma$ and stabiliser $\epsilon_{\mathrm{LN}}$ is $\|\gamma\|_\infty/\sqrt{\epsilon_{\mathrm{LN}}}$-Lipschitz. The second DCSE layer receives inputs built from layer-normalised rows plus bounded attention and feed-forward updates, so $\|Z\|_2$ in its branch bounds is bounded too. For $\kappa_T$, dot-product attention is not globally Lipschitz~\cite{kim2021lipschitz}, but in a pre-layer-norm block the attention only sees rows of norm at most $\sqrt{d}\|\gamma\|_\infty+\|\beta\|_2$, where it is Lipschitz. The encoder output is therefore Lipschitz in the operator on $\mathcal{U}$ for the bounded reachable input set, with a bound linear in $\|\Delta L\|_\F$.
 
We do not instantiate $\kappa_T\kappa_L(1+\kappa_Z\kappa_T)$. With $\epsilon_{\mathrm{LN}}=10^{-5}$ the layer-norm bound contains the factor $316.2\,\|\gamma\|_\infty$, and products of worst-case layer constants overstate the true Lipschitz constant of deep networks by many orders of magnitude~\cite{virmaux2018lipschitz,fazlyab2019efficient}. The quantitative content sits in the branch constants, above all $c_{\mathrm{cheb}}$. Proposition~\ref{prop:stack} gives a bound linear in $\|\Delta L\|_\F$ on the stated admissible operator set. It does not give a uniform guarantee over arbitrary operator edits or edits that make a diffusion row sum approach zero. Section~\ref{subsec:rq3} compares the bound with a measured response at the first DCSE layer, where both live in the same units.
 
\subsection{MSR and the Itakura--Saito Divergence}
\label{subsec:msr_is}
 
The Itakura-Saito divergence between two PSDs,\\ $D_{\mathrm{IS}}(P_X\|P_Z)=\sum_{n,\omega}[P_X/P_Z-\log(P_X/P_Z)-1]$, is the Bregman divergence of $-\log$ and the maximum-likelihood criterion for matching Gaussian-process spectra~\cite{fevotte2009nmf}. It is invariant to a common rescaling of both spectra.
 
\begin{proposition}[MSR and the Itakura--Saito divergence]
\label{prop:is}
Let $u(n,\omega)=\log P_Z(n,\omega)-\log P_X(n,\omega)$ on bins where no clamp in~\eqref{eq:msr} is active, and ignore $\epsilon$. Then
\[
  \mathcal{L}_{\mathrm{freq}} = \frac{2D_{\mathrm{IS}}(P_X\|P_Z)}{NW's_X^2} + \frac{\sum u^3}{3NW's_X^2} + O\Big(\frac{\sum u^4}{NW's_X^2}\Big).
\]
\end{proposition}
 
To leading order in the log-spectral residual, minimising MSR minimises the Itakura--Saito divergence between the encoder and input spectra. This is a local property that any loss on log-power ratios shares, and we state it to make the weighting of MSR explicit. Two consequences matter for forecasting. First, the residual $u$ is a log-ratio, so an error of $10\%$ in a bin holding little power costs as much as an error of $10\%$ in the dominant bin. Under a squared-error loss on values, which by Parseval's theorem is a squared-error loss on Fourier coefficients, the same relative error in the weak bin costs far less. In a one-hour speed window the weak bins are mostly the fast ones, which carry short transients such as a queue forming. Second, ignoring $\epsilon$ and the clamps, the residual $u$ does not change when both spectra of a sensor are multiplied by the same factor, so up to the stop-gradient scale $s_X$ the loss weights relative spectral ratios and not raw power levels. The scale $s_X$ is the mean absolute log-PSD of the input and does change with the input level. It multiplies every bin of a window by the same factor, so it sets the overall size of the loss and leaves the relative weight of the bins unchanged. Far from spectral agreement the cubic and higher terms matter, and on clamped bins the loss is no longer tied to $D_{\mathrm{IS}}$.
\section{Experimental Setup}
\label{sec:setup}
 
\subsubsection*{\textbf{Datasets and Preprocessing}}
Table~\ref{tab:datasets} lists the four benchmark datasets. METR-LA and PEMS-BAY~\cite{li2018dcrnn} record speed, and PEMS04 and PEMS08~\cite{guo2019astgcn} record flow. All are aggregated to five-minute intervals and cut into windows with stride one. The speed datasets are split chronologically $70/10/20$ and the flow datasets $60/20/20$, the splits used by the baselines we compare against~\cite{li2018dcrnn,liu2023staeformer,jiang2023pdformer}. Inputs and targets are $12$ steps long, and we report horizons $3$, $6$ and $12$ on the speed datasets and the average over all 12 horizons on the flow datasets. Readings are standardised per node with training-set statistics. Missing readings, which the datasets record as zero, are excluded from the loss and from every reported metric by a mask that keeps readings above $1$ in the original units. For the exploratory transfer study of Section~\ref{subsec:transfer} we use monthly value-weighted returns of the Fama--French 49 industry portfolios~\cite{fama1997industry} from the Kenneth French data library, July 1926 to March 2026. Six industries whose histories start in the 1960s are excluded, which leaves 43 portfolios with complete histories. After the 20-month volatility warm-up (Appendix~\ref{app:impl}) the first usable month is March 1928, so $982$ months before 2010 are available for fine-tuning. Months from 2010 onward are held out.
 
\begin{table}[!t]
\centering
\caption{Datasets}
\label{tab:datasets}
\footnotesize
\setlength{\tabcolsep}{3.4pt}
\resizebox{\columnwidth}{!}{%
\begin{tabular}{lccccc}
\toprule
Dataset & Signal & Sensors & Steps & Period & Split (\%) \\
\midrule
METR-LA  & speed & 207 & 34{,}272 & 03/2012--06/2012 & 70/10/20 \\
PEMS-BAY & speed & 325 & 52{,}116 & 01/2017--06/2017 & 70/10/20 \\
PEMS04   & flow  & 307 & 16{,}992 & 01/2018--02/2018 & 60/20/20 \\
PEMS08   & flow  & 170 & 17{,}856 & 07/2016--08/2016 & 60/20/20 \\
\bottomrule
\end{tabular}}
\end{table}
 
\subsubsection*{\textbf{Baselines}}
On the speed datasets we compare with twelve models that report results at horizons 3, 6 and 12 under the chronological split above: DCRNN~\cite{li2018dcrnn}, Graph WaveNet~\cite{wu2019gwnet}, MTGNN~\cite{wu2020mtgnn}, STEP~\cite{shao2022step}, D$^2$STGNN~\cite{shao2022d2stgnn}, MegaCRN~\cite{jiang2023megacrn}, Trafformer~\cite{jin2023trafformer}, STAEformer~\cite{liu2023staeformer}, STD-MAE~\cite{gao2024stdmae}, AdpSTGCN~\cite{zhang2024adpstgcn}, TITAN~\cite{wang2024titan} and T-Graphormer~\cite{bai2025tgraphormer}. They cover recurrent, convolutional, adaptive-graph, pre-training and attention-based designs. All use a 12-step input window, except that STEP and STD-MAE also encode long-term history through pre-trained encoders. On the flow datasets we compare against PDFormer~\cite{jiang2023pdformer} and STAEformer, which use the same 12-step input and report the average over all 12 horizons. STD-MAE~\cite{gao2024stdmae}, which pre-trains on 864 (PEMS04) and 2016 (PEMS08) steps of history, is listed as a long-history reference. The baseline numbers in Tables~\ref{tab:metrla}, \ref{tab:pemsbay} and~\ref{tab:flow} come from the original papers. General time-series Transformers such as FEDformer~\cite{zhou2022fedformer} and PatchTST~\cite{nie2023patchtst} have no official numbers under this protocol. To separate the model from our pipeline, we retrain a reduced Graph WaveNet (hidden width $32$, four layers, $50$ epochs) and STAEformer from its official code, and score two naive forecasters, all under our own preprocessing and evaluation code. Each retrained baseline is a single run with its own training recipe.
 
\subsubsection*{\textbf{Protocol and Implementation}}
One model is trained with $H_{\max}=12$ and evaluated at every horizon by slicing its output, as in STAEformer and T-Graphormer. We use AdamW~\cite{loshchilov2019adamw} with peak learning rate $2\times10^{-4}$, weight decay $3\times10^{-4}$, cosine annealing with a $5\%$ linear warm-up, gradient clipping at $2.0$, dropout $0.1$ and mixed precision. Batch size is $64$ on METR-LA, $32$ on PEMS-BAY and $16$ on PEMS04 and PEMS08. Training runs for at most $180$ epochs and stops once validation MAE has not improved for six consecutive checks, that is, 30 epochs. After the self-supervised warm-up, validation MAE is checked every five epochs, and the five best checkpoints are averaged at the end. Every speed-dataset run stopped between epochs $65$ and $75$. One hyperparameter configuration (Supplementary Table~S1) is used apart from batch size. Hyperparameters were tuned on the validation split only, and early stopping, checkpoint selection and checkpoint averaging also use the validation split. The test split was never used to select anything. The model has $5.22$ million parameters. All runs used one workstation with an AMD Ryzen~9 9900 processor, $32$~GB of RAM and one NVIDIA GeForce RTX 5060 Ti with $16$~GB of memory, running PyTorch 2.12.0 with CUDA 13.0 and cuDNN 9.20.0.
 
\subsubsection*{\textbf{Seeds and Uncertainty}}
METR-LA results are means over five seeds ($13$, $42$, $101$, $202$, $303$) and PEMS-BAY results over three ($42$, $101$, $303$). PEMS-BAY used fewer seeds because each run takes eight to nine hours and the seed spread was already an order of magnitude below the 60-minute margin to the next-best published model. Intervals are $95\%$ Student-$t$ intervals over seeds. They describe training variability, not uncertainty over test windows. PEMS04 and PEMS08 results are single runs with seed $42$ and carry no interval. Baseline papers report single numbers, so comparisons with them are numerical and carry no significance claim. Supplementary Table~S2 lists every run.
 
\section{Results}
\label{sec:results}
This section answers the four research questions. RQ1 compares ST-GFM with twelve published baselines on the speed datasets METR-LA and PEMS-BAY, with published baselines on the flow datasets PEMS04 and PEMS08, and with models retrained under our own evaluation pipeline. RQ2 removes one component at a time. RQ3 perturbs the graph operator and compares the measured response of the first DCSE layer with the bound of Section~\ref{sec:theory}. RQ4 locates frequency shift in the test data and compares accuracy on weekend windows with accuracy on the full test set. The section ends with an exploratory transfer to industry portfolio returns and the computational cost of the model.

\subsection{RQ1: Forecasting Accuracy}
\label{subsec:rq1}
 
\subsubsection{Speed Datasets}
\begin{figure}
\centering
\begin{tikzpicture}
\begin{axis}[
  width=\columnwidth, height=6.4cm,
  xmin=3.0, xmax=3.66, ymin=1.52, ymax=2.12,
  xtick={3.0,3.1,3.2,3.3,3.4,3.5,3.6}, ytick={1.6,1.7,1.8,1.9,2.0,2.1},
  xticklabel style={/pgf/number format/fixed,/pgf/number format/fixed zerofill,/pgf/number format/precision=1},
  yticklabel style={/pgf/number format/fixed,/pgf/number format/fixed zerofill,/pgf/number format/precision=1},
  xlabel={METR-LA, 60-minute MAE (mph)}, ylabel={PEMS-BAY, 60-minute MAE (mph)},
  label style={font=\footnotesize}, tick label style={font=\scriptsize},
  grid=major, grid style={black!8},
  axis line style={black!45}, tick style={black!45},
  clip=false,
]
\fill[cOurs!7] (axis cs:3.098,1.576) rectangle (axis cs:3.66,2.12);
\draw[cOurs!55, densely dashed, line width=0.6pt] (axis cs:3.098,2.12) -- (axis cs:3.098,1.576) -- (axis cs:3.66,1.576);
\node[font=\scriptsize\itshape, text=cOurs!80!black, anchor=north east, align=right] at (axis cs:3.655,1.675) {worse than ST-GFM\\on both datasets};
\addplot[only marks, mark=*, mark size=1.9pt, black!60] coordinates {(3.60,2.07) (3.53,1.95) (3.49,1.94) (3.37,1.79) (3.35,1.85) (3.38,1.88) (3.41,1.88) (3.34,1.88) (3.40,1.77) (3.40,1.92) (3.08,1.69) (3.19,1.63)};
\addplot[only marks, mark=pentagon, mark size=2.8pt, line width=1.0pt, cSTAE] coordinates {(3.21,1.81)};
\addplot[only marks, mark=star, mark size=4.2pt, line width=1.3pt, cOurs] coordinates {(3.098,1.576)};
\node[font=\scriptsize, anchor=east] at (axis cs:3.593,2.07) {DCRNN};
\node[font=\scriptsize, anchor=south] at (axis cs:3.545,1.954) {Graph WaveNet};
\node[font=\scriptsize, anchor=north] at (axis cs:3.49,1.932) {MTGNN};
\node[font=\scriptsize, anchor=east] at (axis cs:3.362,1.79) {STEP};
\node[font=\scriptsize, anchor=west] at (axis cs:3.407,1.77) {STD-MAE};
\node[font=\scriptsize, anchor=east] at (axis cs:3.343,1.85) {D$^2$STGNN};
\node[font=\scriptsize, anchor=east] at (axis cs:3.333,1.885) {STAEformer};
\node[font=\scriptsize, anchor=south] at (axis cs:3.40,1.926) {AdpSTGCN};
\draw[black!40, line width=0.4pt] (axis cs:3.384,1.875) -- (axis cs:3.455,1.815);
\node[font=\scriptsize, anchor=west] at (axis cs:3.455,1.815) {MegaCRN};
\draw[black!40, line width=0.4pt] (axis cs:3.414,1.875) -- (axis cs:3.455,1.855);
\node[font=\scriptsize, anchor=west] at (axis cs:3.455,1.855) {Trafformer};
\node[font=\scriptsize, text=cSTAE!85!black, anchor=north, align=center] at (axis cs:3.21,1.799) {STAEformer\\(our pipeline)};
\node[font=\scriptsize, anchor=south] at (axis cs:3.08,1.698) {TITAN};
\node[font=\scriptsize, anchor=west] at (axis cs:3.198,1.63) {T-Graphormer};
\node[font=\scriptsize\bfseries, text=cOurs, anchor=north west] at (axis cs:3.108,1.569) {ST-GFM (ours)};
\end{axis}
\end{tikzpicture}
\caption{60-minute MAE on METR-LA and PEMS-BAY for every compared model that reports both datasets. Lower left is better. The shaded region holds the models with a higher MAE than ST-GFM on both datasets. Gray dots are published values from the original papers. The open pentagon is STAEformer retrained under our pipeline (Table~\ref{tab:pipeline}).}
\label{fig:teaser}
\end{figure}
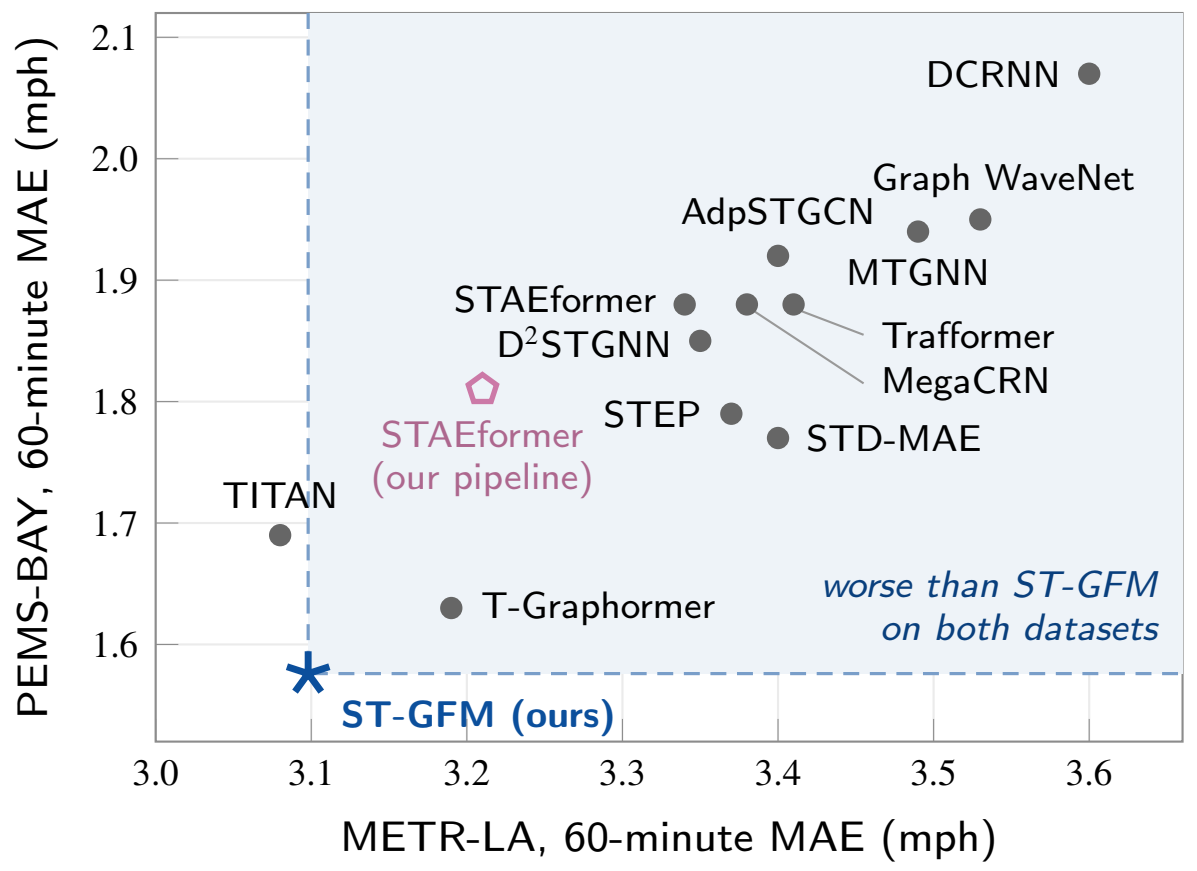

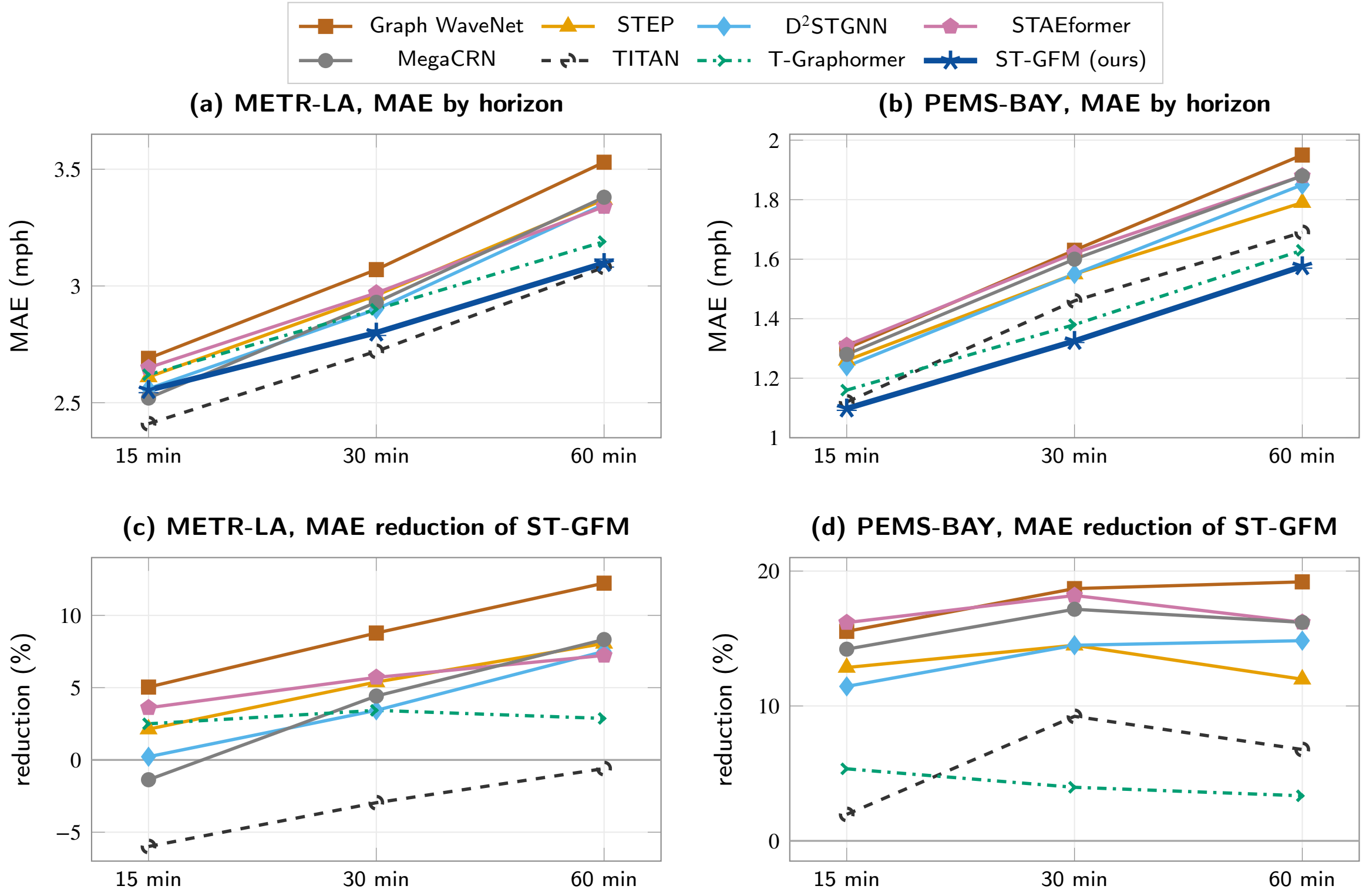
\begin{figure*}
\centering
\begin{tikzpicture}
\begin{groupplot}[
  group style={group size=2 by 2, horizontal sep=1.45cm, vertical sep=1.35cm},
  width=0.46\textwidth, height=5.0cm,
  xtick={1,2,3}, xticklabels={15 min, 30 min, 60 min},
  xmin=0.75, xmax=3.25,
  label style={font=\footnotesize}, tick label style={font=\scriptsize},
  title style={font=\footnotesize\bfseries, yshift=-4pt},
  grid=major, grid style={black!8},
  axis line style={black!45}, tick style={black!45},
  every axis plot/.append style={line width=1.05pt},
]
\nextgroupplot[title={(a) METR-LA, MAE by horizon}, ylabel={MAE (mph)}, ymin=2.35, ymax=3.65,
  legend to name=hzlegend, legend columns=4, legend style={font=\scriptsize, draw=black!20, column sep=3pt}]
\addplot[cGWN, mark=square*, mark size=1.9pt] coordinates {(1,2.69)(2,3.07)(3,3.53)}; \addlegendentry{Graph WaveNet}
\addplot[cSTEP, mark=triangle*, mark size=2.3pt] coordinates {(1,2.61)(2,2.96)(3,3.37)}; \addlegendentry{STEP}
\addplot[cD2, mark=diamond*, mark size=2.4pt] coordinates {(1,2.56)(2,2.90)(3,3.35)}; \addlegendentry{D$^2$STGNN}
\addplot[cSTAE, mark=pentagon*, mark size=2.2pt] coordinates {(1,2.65)(2,2.97)(3,3.34)}; \addlegendentry{STAEformer}
\addplot[cMega, mark=*, mark size=1.9pt] coordinates {(1,2.52)(2,2.93)(3,3.38)}; \addlegendentry{MegaCRN}
\addplot[black!80, dashed, mark=o, mark size=2.0pt] coordinates {(1,2.41)(2,2.72)(3,3.08)}; \addlegendentry{TITAN}
\addplot[cTG, dashdotted, mark=x, mark size=2.4pt] coordinates {(1,2.62)(2,2.90)(3,3.19)}; \addlegendentry{T-Graphormer}
\addplot[cOurs, line width=1.8pt, mark=star, mark size=3.2pt, mark options={line width=1.1pt}, error bars/.cd, y dir=both, y explicit, error bar style={line width=0.9pt}] coordinates {(1,2.554) +- (0,0.011) (2,2.800) +- (0,0.012) (3,3.098) +- (0,0.017)}; \addlegendentry{ST-GFM (ours)}
\nextgroupplot[title={(b) PEMS-BAY, MAE by horizon}, ylabel={MAE (mph)}, ymin=1.0, ymax=2.02]
\addplot[cGWN, mark=square*, mark size=1.9pt] coordinates {(1,1.30)(2,1.63)(3,1.95)};
\addplot[cSTEP, mark=triangle*, mark size=2.3pt] coordinates {(1,1.26)(2,1.55)(3,1.79)};
\addplot[cD2, mark=diamond*, mark size=2.4pt] coordinates {(1,1.24)(2,1.55)(3,1.85)};
\addplot[cSTAE, mark=pentagon*, mark size=2.2pt] coordinates {(1,1.31)(2,1.62)(3,1.88)};
\addplot[cMega, mark=*, mark size=1.9pt] coordinates {(1,1.28)(2,1.60)(3,1.88)};
\addplot[black!80, dashed, mark=o, mark size=2.0pt] coordinates {(1,1.12)(2,1.46)(3,1.69)};
\addplot[cTG, dashdotted, mark=x, mark size=2.4pt] coordinates {(1,1.16)(2,1.38)(3,1.63)};
\addplot[cOurs, line width=1.8pt, mark=star, mark size=3.2pt, mark options={line width=1.1pt}, error bars/.cd, y dir=both, y explicit, error bar style={line width=0.9pt}] coordinates {(1,1.098) +- (0,0.006) (2,1.325) +- (0,0.005) (3,1.576) +- (0,0.006)};
\nextgroupplot[title={(c) METR-LA, MAE reduction of ST-GFM}, ylabel={reduction (\%)}, ymin=-7, ymax=14]
\addplot[black!35, line width=0.6pt, forget plot] coordinates {(0.75,0)(3.25,0)};
\addplot[cGWN, mark=square*, mark size=1.9pt] coordinates {(1,5.04)(2,8.78)(3,12.23)};
\addplot[cSTEP, mark=triangle*, mark size=2.3pt] coordinates {(1,2.13)(2,5.39)(3,8.06)};
\addplot[cD2, mark=diamond*, mark size=2.4pt] coordinates {(1,0.22)(2,3.43)(3,7.51)};
\addplot[cSTAE, mark=pentagon*, mark size=2.2pt] coordinates {(1,3.61)(2,5.71)(3,7.23)};
\addplot[cMega, mark=*, mark size=1.9pt] coordinates {(1,-1.37)(2,4.42)(3,8.33)};
\addplot[black!80, dashed, mark=o, mark size=2.0pt] coordinates {(1,-5.99)(2,-2.96)(3,-0.60)};
\addplot[cTG, dashdotted, mark=x, mark size=2.4pt] coordinates {(1,2.50)(2,3.43)(3,2.87)};
\nextgroupplot[title={(d) PEMS-BAY, MAE reduction of ST-GFM}, ylabel={reduction (\%)}, ymin=-1.5, ymax=21]
\addplot[black!35, line width=0.6pt, forget plot] coordinates {(0.75,0)(3.25,0)};
\addplot[cGWN, mark=square*, mark size=1.9pt] coordinates {(1,15.53)(2,18.70)(3,19.20)};
\addplot[cSTEP, mark=triangle*, mark size=2.3pt] coordinates {(1,12.85)(2,14.50)(3,11.98)};
\addplot[cD2, mark=diamond*, mark size=2.4pt] coordinates {(1,11.45)(2,14.50)(3,14.84)};
\addplot[cSTAE, mark=pentagon*, mark size=2.2pt] coordinates {(1,16.18)(2,18.19)(3,16.20)};
\addplot[cMega, mark=*, mark size=1.9pt] coordinates {(1,14.21)(2,17.17)(3,16.20)};
\addplot[black!80, dashed, mark=o, mark size=2.0pt] coordinates {(1,1.96)(2,9.23)(3,6.77)};
\addplot[cTG, dashdotted, mark=x, mark size=2.4pt] coordinates {(1,5.34)(2,3.97)(3,3.34)};
\end{groupplot}
\node[anchor=south] at ($(group c1r1.north)!0.5!(group c2r1.north)+(0,0.45)$) {\pgfplotslegendfromname{hzlegend}};
\end{tikzpicture}
\caption{Accuracy by horizon against seven baselines, including TITAN and T-Graphormer. (a) and (b) MAE of ST-GFM and the baselines, with $95\%$ seed intervals on ST-GFM (five seeds on METR-LA, three on PEMS-BAY, smaller than the marker on PEMS-BAY). (c) and (d) Relative MAE reduction of ST-GFM over each baseline, computed from unrounded seed means and the values in Tables~\ref{tab:metrla} and~\ref{tab:pemsbay}. Baselines are single published numbers.}
\label{fig:horizons}
\end{figure*}

Tables~\ref{tab:metrla} and~\ref{tab:pemsbay} compare ST-GFM with published numbers. Figure~\ref{fig:teaser} places ST-GFM against every baseline that reports both speed benchmarks, and Figure~\ref{fig:horizons} plots MAE against horizon. On PEMS-BAY, ST-GFM has the lowest MAE and MAPE at every horizon among all thirteen methods. Its 60-minute MAE of $1.576\pm0.006$ is $3.3\%$ below T-Graphormer, the next best, $6.7\%$ below TITAN and $11.0\%$ below STD-MAE.
 
METR-LA behaves differently. TITAN reports lower values than ST-GFM in every cell, with a margin that narrows with horizon from $0.144$ MAE at 15 minutes to $0.018$ at 60, about one seed interval. ST-GFM has the second-lowest MAE and MAPE at 30 and 60 minutes, and its 60-minute MAE is $2.9\%$ below T-Graphormer and $7.2\%$ below STAEformer. At 15 minutes MegaCRN is also lower, $2.52$ against $2.554\pm0.011$. T-Graphormer has the lowest RMSE at every horizon on PEMS-BAY and at 60 minutes on METR-LA. RMSE weights the largest errors, which on freeway speed data mostly come from rare incident-driven drops. ST-GFM is trained on MAE and anchored to a persistence term, and it lowers typical errors more than the largest ones.
 
\begin{table*}[ht]
\centering
\caption{Forecasting Results on METR-LA. ST-GFM Entries Are Five-Seed Means With $95\%$ Seed Intervals. Bold Marks the Best and Underline the Second Best in Each Column. Baseline Numbers Are From the Original Papers.}
\label{tab:metrla}
\footnotesize
\setlength{\tabcolsep}{3.0pt}
\begin{tabular}{lccccccccc}
\toprule
& \multicolumn{3}{c}{$H=3$ (15 min)} & \multicolumn{3}{c}{$H=6$ (30 min)} & \multicolumn{3}{c}{$H=12$ (60 min)} \\
\cmidrule(lr){2-4}\cmidrule(lr){5-7}\cmidrule(lr){8-10}
Method & MAE & RMSE & MAPE (\%) & MAE & RMSE & MAPE (\%) & MAE & RMSE & MAPE (\%) \\
\midrule
DCRNN~\cite{li2018dcrnn}                 & 2.77 & 5.38 & 7.30 & 3.15 & 6.45 & 8.80 & 3.60 & 7.59 & 10.50 \\
Graph WaveNet~\cite{wu2019gwnet}         & 2.69 & 5.15 & 6.90 & 3.07 & 6.22 & 8.37 & 3.53 & 7.37 & 10.01 \\
MTGNN~\cite{wu2020mtgnn}                 & 2.69 & 5.18 & 6.86 & 3.05 & 6.17 & 8.19 & 3.49 & 7.23 & 9.87 \\
STEP~\cite{shao2022step}                 & 2.61 & 4.98 & 6.60 & 2.96 & 5.97 & 7.96 & 3.37 & 6.99 & 9.61 \\
D$^2$STGNN~\cite{shao2022d2stgnn}        & 2.56 & 4.88 & 6.48 & 2.90 & 5.89 & 7.78 & 3.35 & 7.03 & 9.40 \\
MegaCRN~\cite{jiang2023megacrn}          & \underline{2.52} & 4.94 & 6.44 & 2.93 & 6.06 & 7.96 & 3.38 & 7.23 & 9.72 \\
Trafformer~\cite{jin2023trafformer}      & 2.78 & 5.35 & 7.32 & 3.05 & 6.18 & 8.67 & 3.41 & 7.17 & 9.96 \\
STAEformer~\cite{liu2023staeformer}      & 2.65 & 5.11 & 6.85 & 2.97 & 6.00 & 8.13 & 3.34 & 7.02 & 9.70 \\
STD-MAE~\cite{gao2024stdmae}             & 2.62 & 5.02 & 6.70 & 2.99 & 6.07 & 8.04 & 3.40 & 7.07 & 9.59 \\
AdpSTGCN~\cite{zhang2024adpstgcn}        & 2.59 & 5.11 & 6.68 & 2.96 & 6.08 & 8.02 & 3.40 & 7.21 & 9.45 \\
TITAN~\cite{wang2024titan}               & \textbf{2.41} & \textbf{4.43} & \textbf{5.85} & \textbf{2.72} & \textbf{5.33} & \textbf{7.07} & \textbf{3.08} & \underline{6.21} & \textbf{8.43} \\
T-Graphormer~\cite{bai2025tgraphormer}   & 2.62 & 4.93 & 6.61 & 2.90 & 5.53 & 7.56 & 3.19 & \textbf{6.12} & 8.62 \\
\midrule
\textbf{ST-GFM (ours)} & 2.554{\scriptsize$\pm$.011} & \underline{4.768}{\scriptsize$\pm$.023} & \underline{6.39}{\scriptsize$\pm$.07} & \underline{2.800}{\scriptsize$\pm$.012} & \underline{5.480}{\scriptsize$\pm$.039} & \underline{7.30}{\scriptsize$\pm$.11} & \underline{3.098}{\scriptsize$\pm$.017} & 6.300{\scriptsize$\pm$.052} & \underline{8.47}{\scriptsize$\pm$.13} \\
\bottomrule
\end{tabular}
\end{table*}
 
\begin{table*}[!t]
\centering
\caption{Forecasting Results on PEMS-BAY. ST-GFM Entries Are Three-Seed Means With $95\%$ Seed Intervals. Formatting as in Table~\ref{tab:metrla}.}
\label{tab:pemsbay}
\footnotesize
\setlength{\tabcolsep}{3.0pt}
\begin{tabular}{lccccccccc}
\toprule
& \multicolumn{3}{c}{$H=3$ (15 min)} & \multicolumn{3}{c}{$H=6$ (30 min)} & \multicolumn{3}{c}{$H=12$ (60 min)} \\
\cmidrule(lr){2-4}\cmidrule(lr){5-7}\cmidrule(lr){8-10}
Method & MAE & RMSE & MAPE (\%) & MAE & RMSE & MAPE (\%) & MAE & RMSE & MAPE (\%) \\
\midrule
DCRNN~\cite{li2018dcrnn}                 & 1.38 & 2.95 & 2.90 & 1.74 & 3.97 & 3.90 & 2.07 & 4.74 & 4.90 \\
Graph WaveNet~\cite{wu2019gwnet}         & 1.30 & 2.74 & 2.73 & 1.63 & 3.70 & 3.67 & 1.95 & 4.52 & 4.63 \\
MTGNN~\cite{wu2020mtgnn}                 & 1.32 & 2.79 & 2.77 & 1.65 & 3.74 & 3.69 & 1.94 & 4.49 & 4.53 \\
STEP~\cite{shao2022step}                 & 1.26 & 2.73 & 2.59 & 1.55 & 3.58 & 3.43 & 1.79 & 4.20 & 4.18 \\
D$^2$STGNN~\cite{shao2022d2stgnn}        & 1.24 & 2.60 & 2.58 & 1.55 & 3.52 & 3.49 & 1.85 & 4.30 & 4.37 \\
MegaCRN~\cite{jiang2023megacrn}          & 1.28 & 2.72 & 2.67 & 1.60 & 3.68 & 3.57 & 1.88 & 4.42 & 4.41 \\
Trafformer~\cite{jin2023trafformer}      & 1.31 & 2.83 & 2.92 & 1.61 & 3.74 & 3.82 & 1.88 & 4.38 & 4.59 \\
STAEformer~\cite{liu2023staeformer}      & 1.31 & 2.78 & 2.76 & 1.62 & 3.68 & 3.62 & 1.88 & 4.34 & 4.41 \\
STD-MAE~\cite{gao2024stdmae}             & 1.23 & 2.62 & 2.56 & 1.53 & 3.53 & 3.42 & 1.77 & 4.20 & 4.17 \\
AdpSTGCN~\cite{zhang2024adpstgcn}        & 1.29 & 2.75 & 2.69 & 1.64 & 3.69 & 3.67 & 1.92 & 4.49 & 4.62 \\
TITAN~\cite{wang2024titan}               & \underline{1.12} & \underline{2.21} & \underline{2.29} & 1.46 & 3.24 & 3.15 & 1.69 & 3.79 & 3.85 \\
T-Graphormer~\cite{bai2025tgraphormer}   & 1.16 & \textbf{2.10} & 2.38 & \underline{1.38} & \textbf{2.64} & \underline{2.97} & \underline{1.63} & \textbf{3.20} & \underline{3.65} \\
\midrule
\textbf{ST-GFM (ours)} & \textbf{1.098}{\scriptsize$\pm$.006} & 2.284{\scriptsize$\pm$.008} & \textbf{2.24}{\scriptsize$\pm$.03} & \textbf{1.325}{\scriptsize$\pm$.005} & \underline{2.991}{\scriptsize$\pm$.022} & \textbf{2.84}{\scriptsize$\pm$.05} & \textbf{1.576}{\scriptsize$\pm$.006} & \underline{3.698}{\scriptsize$\pm$.053} & \textbf{3.51}{\scriptsize$\pm$.08} \\
\bottomrule
\end{tabular}
\end{table*}
 
On METR-LA the relative MAE reduction grows with horizon against Graph WaveNet, STEP, D$^2$STGNN, STAEformer and MegaCRN in Figure~\ref{fig:horizons}. It rises from $5.0\%$ to $12.2\%$ against Graph WaveNet and from $3.6\%$ to $7.2\%$ against STAEformer. Against TITAN it remains negative, changing from $-6.0\%$ at 15 minutes to $-0.6\%$ at 60 minutes. The reduction against T-Graphormer is $2.5\%$, $3.4\%$ and $2.9\%$ at the three horizons, so it is not monotonic. On PEMS-BAY there is no consistent trend. The margin over Graph WaveNet stays near $19\%$ from 30 minutes on, and the margin over STEP moves between $12\%$ and $14.5\%$. Supplementary Table~S2 lists every run. On METR-LA the seed standard deviation never exceeds $0.014$ MAE, $0.042$ RMSE and $0.11$ percentage points of MAPE. On PEMS-BAY it never exceeds $0.002$ MAE, $0.021$ RMSE and $0.03$ points of MAPE. On PEMS-BAY the 60-minute interval, $\pm0.006$, is about one ninth of the gap to T-Graphormer. On METR-LA the interval, $\pm0.017$, is about the size of the gap to TITAN.
 
\subsubsection{Flow Datasets}
Table~\ref{tab:flow} compares ST-GFM with the two baselines that use a 12-step input and report the average over all 12 horizons on PEMS04 and PEMS08. ST-GFM is lower than both on all six values, by $1.0\%$ to $2.1\%$ against STAEformer and $0.4\%$ to $3.5\%$ against PDFormer. ST-GFM is a single run here with no seed interval, so we cannot say whether the smallest margins, $0.12$ RMSE against PDFormer and $0.19$ MAE against STAEformer on PEMS04, exceed run-to-run variation. The model was designed and tuned on speed data, and the flow result tests the architecture on a signal with a different range and different noise. STD-MAE~\cite{gao2024stdmae}, which pre-trains on several days of history while ST-GFM sees 12 steps, has a lower PEMS04 MAE and RMSE and a lower PEMS08 RMSE. STEP reports a PEMS04 RMSE of $29.71$, also below ST-GFM. ST-GFM is lower than STD-MAE on PEMS08 MAE and on MAPE for both datasets, by $0.03$ points of MAPE on PEMS08.
 
\begin{table}[!t]
\centering
\caption{Average Over All 12 Horizons on the Flow Datasets PEMS04 and PEMS08. Bold Marks the Best and Underline the Second Best Among Models With a 12-Step Input. 
STD-MAE Pre-Trains on 864 to 2016 Steps and Is Listed Below the Rule as a Long-History Reference. 
Baseline Numbers Are From the Original Papers. ST-GFM Is a Single Run (Seed 42).}
\label{tab:flow}
\footnotesize
\setlength{\tabcolsep}{2.9pt}
\resizebox{\columnwidth}{!}{%
\begin{tabular}{lcccccc}
\toprule
& \multicolumn{3}{c}{PEMS04} & \multicolumn{3}{c}{PEMS08} \\
\cmidrule(lr){2-4}\cmidrule(lr){5-7}
Method & MAE & RMSE & MAPE (\%) & MAE & RMSE & MAPE (\%) \\
\midrule
PDFormer~\cite{jiang2023pdformer}    & 18.32 & \underline{29.97} & 12.10 & 13.58 & 23.51 & 9.05 \\
STAEformer~\cite{liu2023staeformer}  & \underline{18.22} & 30.18 & \underline{11.98} & \underline{13.46} & \underline{23.25} & \underline{8.88} \\
\textbf{ST-GFM (ours)}               & \textbf{18.03} & \textbf{29.85} & \textbf{11.79} & \textbf{13.18} & \textbf{22.91} & \textbf{8.73} \\
\midrule
STD-MAE~\cite{gao2024stdmae}         & 17.80 & 29.25 & 11.97 & 13.44 & 22.47 & 8.76 \\
\bottomrule
\end{tabular}}
\end{table}
 
\subsubsection{Same-Pipeline Comparison}
Published numbers come from different code bases, so a margin over them mixes architecture with pipeline. Table~\ref{tab:pipeline} scores reference models with our own preprocessing, windows, mask and metric code. The comparison shares data handling and evaluation. Training follows each model's own recipe. Persistence repeats the last observed reading at every horizon. The historical average predicts, for each sensor, the mean of its non-zero training readings at the same five-minute slot of the week. Our historical average gives $4.19$ on METR-LA, close to the $4.16$ reported with DCRNN~\cite{li2018dcrnn}, which checks the pipeline.
 
\begin{table}[!t]
\centering
\caption{MAE Under Our Pipeline, With the Same Test Windows, Code and Mask for Every Model. Graph WaveNet Is the Reduced Model of Section~\ref{sec:setup}. ST-GFM Is the Mean Over Seeds. Bold Marks the Lowest MAE per Column, Including Ties at the Reported Precision.}
\label{tab:pipeline}
\footnotesize
\setlength{\tabcolsep}{2.2pt}
\resizebox{\columnwidth}{!}{%
\begin{tabular}{lcccccc}
\toprule
& \multicolumn{3}{c}{METR-LA} & \multicolumn{3}{c}{PEMS-BAY} \\
\cmidrule(lr){2-4}\cmidrule(lr){5-7}
Method & 15 min & 30 min & 60 min & 15 min & 30 min & 60 min \\
\midrule
Persistence             & 3.557 & 4.153 & 5.145 & 1.301 & 1.649 & 2.173 \\
Historical average      & 4.193 & 4.193 & 4.193 & 2.631 & 2.631 & 2.630 \\
Graph WaveNet           & --    & --    & 3.258 & --    & --    & --    \\
STAEformer              & \textbf{2.55} & 2.86 & 3.21 & 1.26 & 1.56 & 1.81 \\
\midrule
\textbf{ST-GFM (ours)} & \textbf{2.554} & \textbf{2.800} & \textbf{3.098} & \textbf{1.098} & \textbf{1.325} & \textbf{1.576} \\
\bottomrule
\end{tabular}}
\end{table}
 
At 60 minutes ST-GFM is $39.8\%$ below persistence and $26.1\%$ below the historical average on METR-LA, and $27.5\%$ and $40.1\%$ below them on PEMS-BAY. Both retrained models score better under our pipeline than in their papers. The reduced Graph WaveNet reaches $3.258$ against the $3.53$ published for the full model, and STAEformer reaches $3.21$ on METR-LA and $1.81$ on PEMS-BAY against the published $3.34$ and $1.88$. Part of the margin over published numbers therefore need not come from the architecture. The gap between published and retrained values could come from our preprocessing, which standardises each sensor with its own training statistics and masks readings at or below $1$ (Section~\ref{sec:setup}), from software versions, from run-to-run variation in single runs, or, for Graph WaveNet, from the reduced configuration. The gap, $3.7\%$ to $7.7\%$ on the two retrained models, is as large as the margins of ST-GFM over the strongest published baselines. We therefore treat the same-pipeline numbers as the primary comparison of forecasting performance under shared data handling and evaluation, and the published-number comparisons as context. Under shared evaluation code, ST-GFM is $4.9\%$ below the reduced Graph WaveNet and $3.5\%$ below STAEformer at 60 minutes on METR-LA, against $12.2\%$ and $7.2\%$ below the published values. On METR-LA the margin over STAEformer narrows at shorter horizons. It is $2.1\%$ at 30 minutes, and at 15 minutes the two are level at the precision of the retrained value, $2.55$ against $2.554$. On PEMS-BAY the gap stays large under the shared pipeline, between $12.9\%$ and $15.0\%$ at every horizon and $13.0\%$ at 60 minutes.
 
\subsection{RQ2: Component Contributions}
\label{subsec:rq2}
 
Table~\ref{tab:ablation} removes one component at a time at $H=12$, and Supplementary Figure~S2 plots the increases. Each variant is a single run trained with the early-stopping rule, validation schedule and five-checkpoint averaging of the full model, and every MAE increase in the table is at least five times the seed standard deviation of the full model on its dataset, which we use as a proxy for the run-to-run spread of the variants. Replacing the DCSE layers with a standard GCN and removing MSR give the two largest single-component increases on both datasets: $+0.32$ and $+0.30$ MAE on METR-LA, $+0.21$ and $+0.22$ on PEMS-BAY. Their order swaps between datasets and the gaps between them are $0.02$ and $0.01$, so we treat the spectral encoder and MSR as jointly dominant.
 
Removing all three self-supervised branches at once costs $0.33$ on METR-LA and $0.24$ on PEMS-BAY, only $0.03$ and $0.02$ more than removing MSR alone. The three single-branch increases sum to $0.47$ and $0.30$, well above the joint figure, so the branch effects are not additive, and removing MSR alone accounts for most of the error increase seen when all three branches are dropped. These are ratios between ablations and not a split of the gain among branches. The masked-node and contrastive branches add $0.10$ and $0.07$ when removed on their own from METR-LA and less on PEMS-BAY, and their effects overlap with that of MSR.
 
The decomposition comes third ($+0.19$ and $+0.13$). In this ablation, collapsing the time axis alone while keeping the three stacks produces more than $60\%$ of the error increase caused by removing the entire decomposition: $0.12$ against $0.19$ on METR-LA and $0.08$ against $0.13$ on PEMS-BAY. Keeping all $12$ steps for the graph layers, the design choice of Section~\ref{subsec:decomp}, matters on its own.
 
\begin{table}[!t]
\centering
\caption{Ablation at $H=12$. $\Delta$ Is the MAE Increase Over the Full Model. The Full-Model Reference Is the Multi-Seed Mean of Tables~\ref{tab:metrla} and~\ref{tab:pemsbay}, and Each Variant Is a Single Run.}
\label{tab:ablation}
\footnotesize
\setlength{\tabcolsep}{4pt}
\resizebox{\columnwidth}{!}{%
\begin{tabular}{lcccc}
\toprule
& \multicolumn{2}{c}{METR-LA} & \multicolumn{2}{c}{PEMS-BAY} \\
\cmidrule(lr){2-3}\cmidrule(lr){4-5}
Variant & MAE & $\Delta$ & MAE & $\Delta$ \\
\midrule
Full ST-GFM                         & 3.098 & --      & 1.576 & --      \\
w/o decomposition                   & 3.29  & $+0.19$ & 1.71  & $+0.13$ \\
w/o DCSE (GCN instead)              & 3.42  & $+0.32$ & 1.79  & $+0.21$ \\
w/o MSR                             & 3.40  & $+0.30$ & 1.80  & $+0.22$ \\
w/o masked nodes                    & 3.20  & $+0.10$ & 1.63  & $+0.05$ \\
w/o temporal contrast               & 3.17  & $+0.07$ & 1.61  & $+0.03$ \\
w/o all three SSL branches          & 3.43  & $+0.33$ & 1.82  & $+0.24$ \\
time axis collapsed in decomposition& 3.22  & $+0.12$ & 1.66  & $+0.08$ \\
\bottomrule
\end{tabular}}
\end{table}
 
The variant without DCSE reaches $3.42$ on METR-LA. This is below the $3.53$ published for Graph WaveNet but above the $3.258$ of the reduced Graph WaveNet under our pipeline (Table~\ref{tab:pipeline}), so the comparison does not isolate how much the components shared by all variants contribute. STID~\cite{shao2022stid} has shown how much identity embeddings alone deliver on these datasets.
 
\begin{figure*}
\centering
\begin{tikzpicture}
\begin{groupplot}[
  group style={group size=2 by 2, horizontal sep=1.5cm, vertical sep=1.55cm},
  width=0.465\textwidth, height=5.1cm,
  xmin=0, xmax=77, xtick={0,10,20,30,40,50,60,70},
  label style={font=\footnotesize}, tick label style={font=\scriptsize},
  title style={font=\footnotesize\bfseries, yshift=-4pt},
  grid=major, grid style={black!7},
  axis line style={black!45}, tick style={black!45},
  set layers, clip mode=individual,
]
\nextgroupplot[title={(a) METR-LA: self-supervised losses}, ymode=log, ymin=0.02, ymax=14, ylabel={loss (log scale)}, xlabel={epoch}, legend to name=ssllegend, legend columns=6, legend style={font=\scriptsize, draw=black!20, column sep=5pt, /tikz/every even column/.append style={column sep=9pt}}]
\addplot[fill=cSTAE!13, draw=none, forget plot, on layer=axis background] coordinates {(0,0.02) (20,0.02) (20,14) (0,14)} \closedcycle;
\addplot[fill=cSTAE!5, draw=none, forget plot, on layer=axis background] coordinates {(20,0.02) (40,0.02) (40,14) (20,14)} \closedcycle;
\addplot[black!45, densely dashed, line width=0.6pt, forget plot] coordinates {(40,0.02) (40,14)};
\node[font=\scriptsize\itshape, text=black!55] at (axis cs:10,8.0) {SSL only};
\node[font=\scriptsize\itshape, text=black!55] at (axis cs:30,8.0) {ramp};
\node[font=\scriptsize\itshape, text=black!55, anchor=west] at (axis cs:41.5,8.0) {SSL weight $=0$};
\addplot[draw=none, name path=f0lo, forget plot] coordinates {(1,1.5957) (10,0.0531) (20,0.0400) (21,0.0458) (30,0.0416) (40,0.1562) (50,1.9162) (60,3.4829)};
\addplot[draw=none, name path=f0hi, forget plot] coordinates {(1,1.8264) (10,0.0726) (20,0.0462) (21,0.0525) (30,0.0461) (40,0.3452) (50,4.5175) (60,6.2117)};
\addplot[cOurs, fill opacity=0.16, forget plot] fill between[of=f0lo and f0hi];
\addplot[cOurs, line width=1.35pt, mark=*, mark size=1.8pt, mark options={fill=white, line width=0.9pt}] coordinates {(1,1.7374) (10,0.0600) (20,0.0425) (21,0.0484) (30,0.0436) (40,0.2549) (50,2.8546) (60,4.3847)}; \addlegendentry{MSR}
\addplot[draw=none, name path=t0lo, forget plot] coordinates {(1,3.0555) (10,0.0760) (20,0.0648) (21,0.0843) (30,0.0638) (40,0.1152) (50,2.5384) (60,4.1976)};
\addplot[draw=none, name path=t0hi, forget plot] coordinates {(1,3.1994) (10,0.0821) (20,0.0720) (21,0.0909) (30,0.0682) (40,0.1555) (50,2.9667) (60,4.5840)};
\addplot[cSTAE, fill opacity=0.16, forget plot] fill between[of=t0lo and t0hi];
\addplot[cSTAE, line width=1.35pt, mark=triangle*, mark size=1.8pt, mark options={fill=white, line width=0.9pt}] coordinates {(1,3.1180) (10,0.0798) (20,0.0679) (21,0.0877) (30,0.0655) (40,0.1293) (50,2.7874) (60,4.4301)}; \addlegendentry{temporal contrast}
\addplot[draw=none, name path=s0lo, forget plot] coordinates {(1,0.2741) (10,0.0653) (20,0.0592) (21,0.0558) (30,0.0473) (40,0.0394) (50,0.0494) (60,0.0509)};
\addplot[draw=none, name path=s0hi, forget plot] coordinates {(1,0.3087) (10,0.0829) (20,0.0911) (21,0.0876) (30,0.0770) (40,0.0569) (50,0.0654) (60,0.0743)};
\addplot[cTG, fill opacity=0.16, forget plot] fill between[of=s0lo and s0hi];
\addplot[cTG, line width=1.35pt, mark=square*, mark size=1.8pt, mark options={fill=white, line width=0.9pt}] coordinates {(1,0.2953) (10,0.0716) (20,0.0714) (21,0.0694) (30,0.0607) (40,0.0476) (50,0.0564) (60,0.0628)}; \addlegendentry{masked nodes}
\addlegendimage{black!80, line width=1.35pt, mark=diamond*, mark size=2pt, mark options={fill=white}}\addlegendentry{validation MAE, mean}
\addlegendimage{black!30, line width=0.6pt}\addlegendentry{single run}
\addlegendimage{only marks, mark=*, mark size=1.6pt, black!80}\addlegendentry{best epoch of a run}
\nextgroupplot[title={(b) PEMS-BAY: self-supervised losses}, ymode=log, ymin=0.02, ymax=14, ylabel={loss (log scale)}, xlabel={epoch}]
\addplot[fill=cSTAE!13, draw=none, forget plot, on layer=axis background] coordinates {(0,0.02) (20,0.02) (20,14) (0,14)} \closedcycle;
\addplot[fill=cSTAE!5, draw=none, forget plot, on layer=axis background] coordinates {(20,0.02) (40,0.02) (40,14) (20,14)} \closedcycle;
\addplot[black!45, densely dashed, line width=0.6pt, forget plot] coordinates {(40,0.02) (40,14)};
\node[font=\scriptsize\itshape, text=black!55] at (axis cs:10,8.0) {SSL only};
\node[font=\scriptsize\itshape, text=black!55] at (axis cs:30,8.0) {ramp};
\node[font=\scriptsize\itshape, text=black!55, anchor=west] at (axis cs:41.5,8.0) {SSL weight $=0$};
\addplot[draw=none, name path=f1lo, forget plot] coordinates {(1,1.1925) (10,0.0443) (20,0.0345) (21,0.0399) (30,0.0393) (40,0.2052) (50,1.9383) (60,2.6535)};
\addplot[draw=none, name path=f1hi, forget plot] coordinates {(1,1.2074) (10,0.0457) (20,0.0371) (21,0.0412) (30,0.0414) (40,0.2481) (50,2.1032) (60,3.2208)};
\addplot[cOurs, fill opacity=0.16, forget plot] fill between[of=f1lo and f1hi];
\addplot[cOurs, line width=1.35pt, mark=*, mark size=1.8pt, mark options={fill=white, line width=0.9pt}] coordinates {(1,1.2004) (10,0.0451) (20,0.0357) (21,0.0405) (30,0.0401) (40,0.2314) (50,2.0125) (60,2.8697)};
\addplot[draw=none, name path=t1lo, forget plot] coordinates {(1,2.5605) (10,0.0703) (20,0.0633) (21,0.0764) (30,0.0609) (40,0.1184) (50,2.6353) (60,4.2054)};
\addplot[draw=none, name path=t1hi, forget plot] coordinates {(1,2.7846) (10,0.0725) (20,0.0664) (21,0.0776) (30,0.0653) (40,0.1680) (50,3.1980) (60,4.5291)};
\addplot[cSTAE, fill opacity=0.16, forget plot] fill between[of=t1lo and t1hi];
\addplot[cSTAE, line width=1.35pt, mark=triangle*, mark size=1.8pt, mark options={fill=white, line width=0.9pt}] coordinates {(1,2.6879) (10,0.0716) (20,0.0647) (21,0.0770) (30,0.0633) (40,0.1433) (50,2.8723) (60,4.3852)};
\addplot[draw=none, name path=s1lo, forget plot] coordinates {(1,0.2455) (10,0.0742) (20,0.0750) (21,0.0745) (30,0.0547) (40,0.0434) (50,0.0545) (60,0.0595)};
\addplot[draw=none, name path=s1hi, forget plot] coordinates {(1,0.2791) (10,0.1014) (20,0.1331) (21,0.1181) (30,0.0868) (40,0.0508) (50,0.0609) (60,0.0683)};
\addplot[cTG, fill opacity=0.16, forget plot] fill between[of=s1lo and s1hi];
\addplot[cTG, line width=1.35pt, mark=square*, mark size=1.8pt, mark options={fill=white, line width=0.9pt}] coordinates {(1,0.2634) (10,0.0904) (20,0.1128) (21,0.1032) (30,0.0747) (40,0.0474) (50,0.0571) (60,0.0624)};
\nextgroupplot[title={(c) METR-LA: validation MAE}, ymin=2.785, ymax=3.07, ytick={2.85,2.90,2.95,3.00,3.05}, ylabel={MAE (mph)}, xlabel={epoch}, yticklabel style={/pgf/number format/fixed, /pgf/number format/fixed zerofill, /pgf/number format/precision=2}]
\addplot[fill=cSTAE!13, draw=none, forget plot, on layer=axis background] coordinates {(0,2.785) (20,2.785) (20,3.07) (0,3.07)} \closedcycle;
\addplot[fill=cSTAE!5, draw=none, forget plot, on layer=axis background] coordinates {(20,2.785) (40,2.785) (40,3.07) (20,3.07)} \closedcycle;
\addplot[black!45, densely dashed, line width=0.6pt, forget plot] coordinates {(40,2.785) (40,3.07)};
\node[font=\scriptsize\itshape, text=black!55] at (axis cs:10,3.05) {SSL only};
\node[font=\scriptsize\itshape, text=black!55] at (axis cs:30,3.05) {ramp};
\node[font=\scriptsize\itshape, text=black!55, anchor=west] at (axis cs:41.5,3.05) {SSL weight $=0$};
\addplot[black!30, line width=0.6pt, forget plot] coordinates {(25,3.0405) (30,2.9301) (35,2.8443) (40,2.8864) (45,2.9094) (50,2.9665) (55,2.9538) (60,2.9870) (65,3.0054)};
\addplot[black!30, line width=0.6pt, forget plot] coordinates {(25,2.9779) (30,2.8927) (35,2.8669) (40,2.8551) (45,2.9049) (50,2.9408) (55,2.9773) (60,3.0059) (65,2.9742) (70,3.0162)};
\addplot[black!30, line width=0.6pt, forget plot] coordinates {(25,2.9608) (30,2.8519) (35,2.8324) (40,2.8650) (45,2.9064) (50,2.9439) (55,2.9608) (60,2.9812) (65,3.0107)};
\addplot[black!30, line width=0.6pt, forget plot] coordinates {(25,3.0325) (30,2.8875) (35,2.8413) (40,2.8699) (45,2.8931) (50,2.9217) (55,2.9255) (60,2.9792) (65,2.9941)};
\addplot[black!30, line width=0.6pt, forget plot] coordinates {(25,3.0231) (30,2.8640) (35,2.8397) (40,2.8484) (45,2.8690) (50,2.9197) (55,2.9338) (60,2.9103) (65,2.9667)};
\addplot[black!80, line width=1.35pt, mark=diamond*, mark size=2pt, mark options={fill=white, line width=0.9pt}, forget plot] coordinates {(25,3.0070) (30,2.8852) (35,2.8449) (40,2.8650) (45,2.8966) (50,2.9385) (55,2.9502) (60,2.9727) (65,2.9902)};
\addplot[only marks, mark=*, mark size=1.6pt, black!80, forget plot] coordinates {(35,2.8443) (40,2.8551) (35,2.8324) (35,2.8413) (35,2.8397)};
\draw[black!65, line width=0.7pt] (axis cs:30,2.8220) -- (axis cs:30,2.8107) -- (axis cs:50,2.8107) -- (axis cs:50,2.8220);
\node[font=\scriptsize, text=black!70, anchor=south west, align=left, inner sep=1.5pt, fill=white, fill opacity=0.9, text opacity=1] at (axis cs:51,2.7907) {23 of 25 averaged\\checkpoints in 30 to 50};
\nextgroupplot[title={(d) PEMS-BAY: validation MAE}, ymin=1.545, ymax=1.72, ytick={1.58,1.62,1.66,1.70}, ylabel={MAE (mph)}, xlabel={epoch}, yticklabel style={/pgf/number format/fixed, /pgf/number format/fixed zerofill, /pgf/number format/precision=2}]
\addplot[fill=cSTAE!13, draw=none, forget plot, on layer=axis background] coordinates {(0,1.545) (20,1.545) (20,1.72) (0,1.72)} \closedcycle;
\addplot[fill=cSTAE!5, draw=none, forget plot, on layer=axis background] coordinates {(20,1.545) (40,1.545) (40,1.72) (20,1.72)} \closedcycle;
\addplot[black!45, densely dashed, line width=0.6pt, forget plot] coordinates {(40,1.545) (40,1.72)};
\node[font=\scriptsize\itshape, text=black!55] at (axis cs:10,1.705) {SSL only};
\node[font=\scriptsize\itshape, text=black!55] at (axis cs:30,1.705) {ramp};
\node[font=\scriptsize\itshape, text=black!55, anchor=west] at (axis cs:41.5,1.705) {SSL weight $=0$};
\addplot[black!30, line width=0.6pt, forget plot] coordinates {(25,1.6551) (30,1.6038) (35,1.5709) (40,1.5735) (45,1.6047) (50,1.5899) (55,1.6023) (60,1.6110) (65,1.6226)};
\addplot[black!30, line width=0.6pt, forget plot] coordinates {(25,1.6977) (30,1.6103) (35,1.5906) (40,1.5883) (45,1.5747) (50,1.5822) (55,1.5975) (60,1.6005) (65,1.6098) (70,1.6173) (75,1.6207)};
\addplot[black!30, line width=0.6pt, forget plot] coordinates {(25,1.6812) (30,1.6125) (35,1.5853) (40,1.5779) (45,1.5715) (50,1.5912) (55,1.5923) (60,1.5948) (65,1.6234) (70,1.6138) (75,1.6240)};
\addplot[black!80, line width=1.35pt, mark=diamond*, mark size=2pt, mark options={fill=white, line width=0.9pt}, forget plot] coordinates {(25,1.6780) (30,1.6089) (35,1.5823) (40,1.5799) (45,1.5836) (50,1.5878) (55,1.5974) (60,1.6021) (65,1.6186)};
\addplot[only marks, mark=*, mark size=1.6pt, black!80, forget plot] coordinates {(35,1.5709) (45,1.5747) (45,1.5715)};
\draw[black!65, line width=0.7pt] (axis cs:30,1.5677) -- (axis cs:30,1.5607) -- (axis cs:50,1.5607) -- (axis cs:50,1.5677);
\node[font=\scriptsize, text=black!70, anchor=south west, align=left, inner sep=1.5pt, fill=white, fill opacity=0.9, text opacity=1] at (axis cs:51,1.5485) {12 of 15 averaged\\checkpoints in 30 to 50};
\end{groupplot}
\node[anchor=south] at ($(group c1r1.north)!0.5!(group c2r1.north)+(0,0.45)$) {\pgfplotslegendfromname{ssllegend}};
\end{tikzpicture}
\caption{Self-supervised losses (a), (b) and validation MAE (c), (d) over training, for all eight runs. Thick lines are means, bands span the runs, thin lines are single runs and dots mark each run's best epoch. Shading marks the SSL-only phase and the ramp-down of its weight. After epoch 40 the SSL losses are monitored, not optimised. Brackets count the averaged checkpoints that fall in epochs 30 to 50.}
\label{fig:ssl}
\end{figure*}
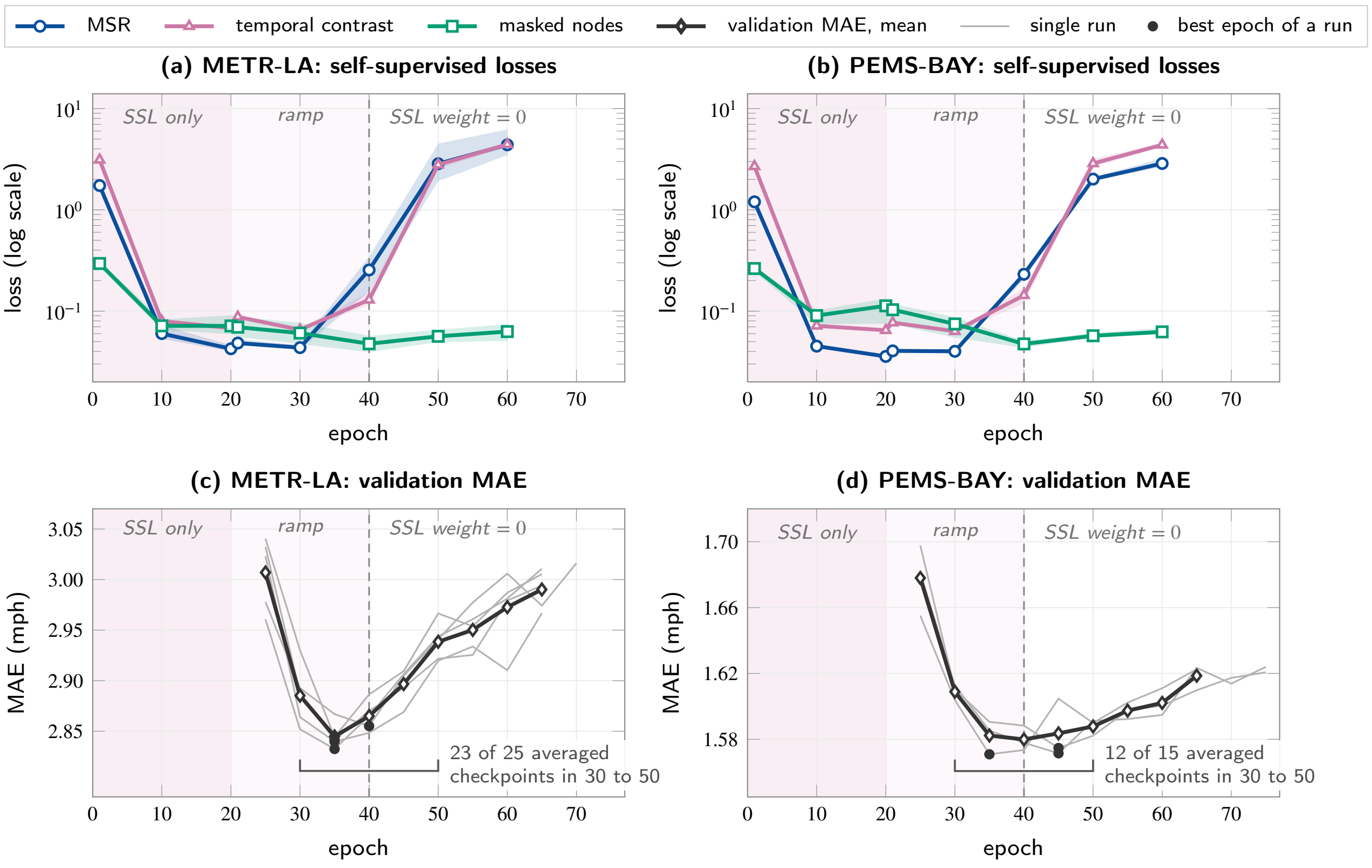

MSR is active only in the first $40$ epochs, and Figure~\ref{fig:ssl} shows why that timing matters. In all eight runs the MSR loss is between $0.035$ and $0.046$ at epoch 20 and still below $0.35$ at epoch 40, when the self-supervised weight reaches zero. By epoch 60 it has risen to between $2.65$ and $6.21$, the contrastive loss has followed the same path, and the masked-node loss lies between $0.05$ and $0.08$. Validation MAE is lowest between epochs 35 and 45 in every run and climbs once the self-supervised terms are switched off, although the supervised training loss keeps falling, from about $0.15$ to $0.13$ on METR-LA and from $0.17$ to $0.15$ on PEMS-BAY between epochs 40 and 60. All 40 averaged checkpoints come from epochs 30 to 60, and 35 of them from epochs 30 to 50, so the deployed model sits close to the end of the self-supervised phase. This fits MSR and the contrastive term acting as regularisers of the encoder while they are active. Once the self-supervised weight reaches zero at epoch 40, the monitored MSR loss rises in every run while the supervised training loss keeps falling, and the averaged checkpoints come from around this transition. We therefore read the MSR ablation gain as an effect of the objective and of the training path it induces. It does not show that the deployed encoder keeps the spectral alignment. The variant without MSR keeps the masked-node and contrastive losses on the same 40-epoch schedule and still loses $0.30$ MAE, so the auxiliary schedule alone does not explain the gain.
 
\subsection{RQ3: Sensitivity to Operator Perturbations}
\label{subsec:rq3}
 
\subsubsection{Random Operator Perturbations}
We add a dense random symmetric $\Delta L$ with $\|\Delta L\|_\F=\varepsilon$, for $\varepsilon\in\{0,0.05,0.1,0.2,0.3,0.5,0.75,1.0\}$, to the trained operator $\Lt_\theta$ and evaluate the unchanged test inputs. The perturbation is symmetric and the trained operator is not, which Theorem~\ref{thm:cheb} allows. For a dense Gaussian perturbation $\eta=\|\Delta L\|_2\approx 2\varepsilon/\sqrt{N}$, about $0.14\varepsilon$ at $N=207$. The sweep therefore reaches $\eta$ of at most about $0.14$ on METR-LA and $0.11$ on PEMS-BAY, which is $\delta^{\mathrm{top}}\approx0.11$ and $0.08$. At the measured operator norm $c_{\mathrm{cheb}}(\rho,\eta)$ stays below $18.8$ on METR-LA (Supplementary Figure~S1).
 
\begin{figure*}[ht]
\centering
\includegraphics[width=\textwidth]{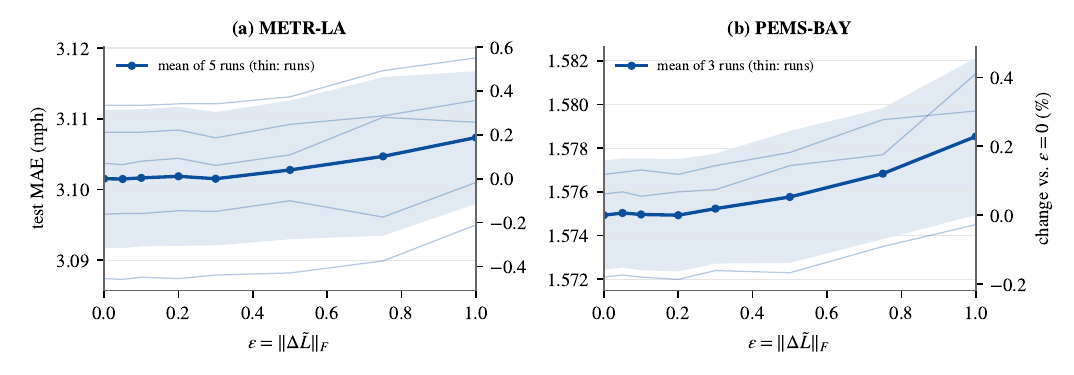}
\caption{Test MAE of ST-GFM under random symmetric perturbations of the operator with Frobenius norm $\varepsilon$. Thick lines: mean over runs, bands: one standard deviation, thin lines: individual runs. One perturbation draw per run and level.}
\label{fig:robustness}
\end{figure*}

Figure~\ref{fig:robustness} shows every run on both datasets. On METR-LA the mean MAE over five runs moves from $3.102$ at $\varepsilon=0$ to $3.107$ at $\varepsilon=1$, a change of $0.19\%$, and no single run changes by more than $0.25\%$. On PEMS-BAY the mean over three runs moves from $1.575$ to $1.579$, a change of $0.23\%$, and no run changes by more than $0.35\%$. Percentages are computed from unrounded values. All levels of $\varepsilon$ are scored in one evaluation pass, and each level is compared with the $\varepsilon=0$ value of that pass. That baseline differs from the clean-test means of Tables~\ref{tab:metrla} and~\ref{tab:pemsbay} by $0.004$ on METR-LA and $0.001$ on PEMS-BAY, which is of the same order as the largest perturbation effect. The sweep therefore supports a flat response and does not resolve its shape. On METR-LA the spread between seeds is also larger than the effect of the largest perturbation.
 
Theorem~\ref{thm:cheb} and Proposition~\ref{prop:stack} bound the encoder representation, while Figure~\ref{fig:robustness} measures forecast error after the aggregation layer, the decoder and the persistence term. The two quantities live in different units, so this sweep does not test the constant of~\eqref{eq:ccheb}. The edge-dropout study below compares the Chebyshev-branch response with its corresponding bound at the realised perturbation sizes.
 
\subsubsection{Matched Adjacency Perturbations}
To compare sensitivity with a baseline under identical conditions, we apply the same random symmetric perturbations with $\|\Delta A\|_\F=\varepsilon$ to the static adjacency of ST-GFM and of the reduced Graph WaveNet, using the seed-101 checkpoint of each and averaging two perturbation draws per level (Table~\ref{tab:matched}). In ST-GFM the static adjacency enters $A_\theta$ with weight $1-\alpha=0.55$~\eqref{eq:adaptive_adj} before degree normalisation and spectral rescaling, so the change that reaches the operator is not a fixed multiple of $\varepsilon$. Both models are almost flat. ST-GFM moves from $3.076$ to $3.087$ ($+0.36\%$) and Graph WaveNet from $3.258$ to $3.264$ ($+0.18\%$). The relative MAE increase is about twice as large for ST-GFM. The models transform the perturbed adjacency differently, so these percentages do not isolate the sensitivity of the operators themselves. The accuracy margin between the two holds across the whole sweep, from $0.182$ at $\varepsilon=0$ to $0.177$ at $\varepsilon=1$, so the gain over Graph WaveNet does not depend on a clean graph.
 
\begin{table}[!t]
\centering
\caption{Matched Adjacency Perturbations on METR-LA at $H=12$ (Seed-101 Checkpoints, Mean of Two Draws). The Graph WaveNet Entries Are From the Reduced Model Retrained Under Our Preprocessing and Evaluation.}
\label{tab:matched}
\footnotesize
\setlength{\tabcolsep}{6pt}
\begin{tabular}{cccc}
\toprule
$\varepsilon$ & Graph WaveNet & ST-GFM & Gap \\
\midrule
0.00 & 3.258 & \textbf{3.076} & 0.182 \\
0.05 & 3.259 & \textbf{3.076} & 0.183 \\
0.10 & 3.259 & \textbf{3.076} & 0.183 \\
0.20 & 3.259 & \textbf{3.077} & 0.182 \\
0.30 & 3.260 & \textbf{3.079} & 0.181 \\
0.50 & 3.261 & \textbf{3.080} & 0.181 \\
0.75 & 3.262 & \textbf{3.084} & 0.178 \\
1.00 & 3.264 & \textbf{3.087} & 0.177 \\
\bottomrule
\end{tabular}
\end{table}
 
\subsubsection{Edge Dropout Scored at the First DCSE Layer}
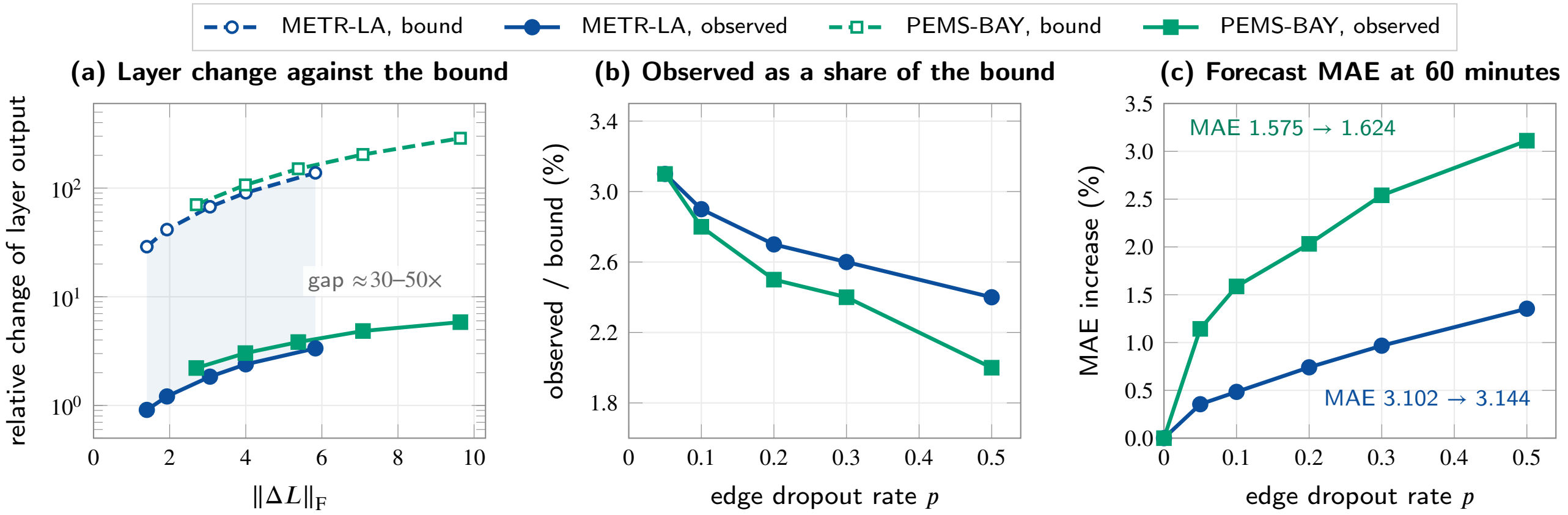
\begin{figure*}
\centering
\begin{tikzpicture}
\begin{groupplot}[
  group style={group size=3 by 1, horizontal sep=1.55cm},
  width=0.335\textwidth, height=5.2cm,
  label style={font=\footnotesize}, tick label style={font=\scriptsize},
  title style={font=\footnotesize\bfseries, yshift=-4pt},
  grid=major, grid style={black!8},
  axis line style={black!45}, tick style={black!45},
  every axis plot/.append style={line width=1.2pt},
]
\nextgroupplot[title={(a) Layer change against the bound}, xlabel={$\|\Delta L\|_\F$}, ylabel={relative change of layer output}, ymode=log, xmin=0, xmax=10.3, ymin=0.5, ymax=600, xtick={0,2,4,6,8,10},
  legend to name=edlegend, legend columns=4, legend style={font=\scriptsize, draw=black!20, column sep=4pt, /tikz/every even column/.append style={column sep=10pt}}]
\addplot[draw=none, name path=bLA, forget plot] coordinates {(1.40,28.9) (1.93,41.5) (3.06,67.3) (4.00,90.5) (5.83,138.1)};
\addplot[draw=none, name path=oLA, forget plot] coordinates {(1.40,0.91) (1.93,1.21) (3.06,1.84) (4.00,2.39) (5.83,3.35)};
\addplot[cOurs, fill opacity=0.07, forget plot] fill between[of=oLA and bLA];
\addplot[cOurs, dash pattern=on 4pt off 2pt, mark=*, mark size=1.7pt, mark options={solid, fill=white, line width=0.8pt}] coordinates {(1.40,28.9) (1.93,41.5) (3.06,67.3) (4.00,90.5) (5.83,138.1)}; \addlegendentry{METR-LA, bound}
\addplot[cOurs, mark=*, mark size=1.9pt, mark options={fill=cOurs}] coordinates {(1.40,0.91) (1.93,1.21) (3.06,1.84) (4.00,2.39) (5.83,3.35)}; \addlegendentry{METR-LA, observed}
\addplot[cTG, dash pattern=on 4pt off 2pt, mark=square*, mark size=1.6pt, mark options={solid, fill=white, line width=0.8pt}] coordinates {(2.70,70.3) (3.99,106.9) (5.38,150.5) (7.08,203.7) (9.64,287.4)}; \addlegendentry{PEMS-BAY, bound}
\addplot[cTG, mark=square*, mark size=1.8pt, mark options={fill=cTG}] coordinates {(2.70,2.21) (3.99,3.03) (5.38,3.83) (7.08,4.84) (9.64,5.84)}; \addlegendentry{PEMS-BAY, observed}
\node[font=\scriptsize, text=black!60, align=center, fill=white, fill opacity=0.85, text opacity=1, inner sep=1.5pt] at (axis cs:7.4,14) {gap $\approx\!30$--$50\times$};
\nextgroupplot[title={(b) Observed as a share of the bound}, xlabel={edge dropout rate $p$}, ylabel={observed / bound (\%)}, xmin=0, xmax=0.54, ymin=1.6, ymax=3.5, xtick={0,0.1,0.2,0.3,0.4,0.5}, ytick={1.8,2.2,2.6,3.0,3.4},
  xticklabel style={/pgf/number format/fixed,/pgf/number format/precision=1},
  yticklabel style={/pgf/number format/fixed,/pgf/number format/fixed zerofill,/pgf/number format/precision=1}]
\addplot[cOurs, mark=*, mark size=1.9pt, mark options={fill=cOurs}] coordinates {(0.05,3.1) (0.10,2.9) (0.20,2.7) (0.30,2.6) (0.50,2.4)};
\addplot[cTG, mark=square*, mark size=1.8pt, mark options={fill=cTG}] coordinates {(0.05,3.1) (0.10,2.8) (0.20,2.5) (0.30,2.4) (0.50,2.0)};
\nextgroupplot[title={(c) Forecast MAE at 60 minutes}, xlabel={edge dropout rate $p$}, ylabel={MAE increase (\%)}, xmin=0, xmax=0.54, ymin=0, ymax=3.5, xtick={0,0.1,0.2,0.3,0.4,0.5}, ytick={0,0.5,1,1.5,2,2.5,3,3.5},
  xticklabel style={/pgf/number format/fixed,/pgf/number format/precision=1},
  yticklabel style={/pgf/number format/fixed,/pgf/number format/fixed zerofill,/pgf/number format/precision=1}]
\addplot[cOurs, mark=*, mark size=1.9pt, mark options={fill=cOurs}] coordinates {(0,0) (0.05,0.355) (0.10,0.484) (0.20,0.741) (0.30,0.967) (0.50,1.354)};
\addplot[cTG, mark=square*, mark size=1.8pt, mark options={fill=cTG}] coordinates {(0,0) (0.05,1.143) (0.10,1.587) (0.20,2.032) (0.30,2.540) (0.50,3.111)};
\node[font=\scriptsize, text=cOurs, anchor=north east] at (axis cs:0.52,0.62) {MAE 3.102 $\to$ 3.144};
\node[font=\scriptsize, text=cTG!80!black, anchor=west] at (axis cs:0.02,3.25) {MAE 1.575 $\to$ 1.624};
\end{groupplot}
\node[anchor=south] at ($(group c2r1.north)+(0,0.45)$) {\pgfplotslegendfromname{edlegend}};
\end{tikzpicture}
\caption{Edge dropout on the static adjacency, mean over runs and three draws (values in Supplementary Table~S3). (a) Observed relative change of the Chebyshev-branch output in the first DCSE layer and the bound of~\eqref{eq:ccheb} at the realised perturbation and the measured operator norm, on a log scale. Shading spans the gap on METR-LA. (b) Observed change as a percentage of the bound. (c) Relative increase of the 60-minute forecast MAE, with the absolute MAE at $p=0$ and $p=0.5$ annotated.}
\label{fig:edgedrop}
\end{figure*}

To test the constant of~\eqref{eq:ccheb} directly, we apply a structured perturbation and read its effect at the layer the bound describes. We remove a fraction $p\in\{0.05,0.1,0.2,0.3,0.5\}$ of the edges of $A_{\mathrm{static}}$ at random, rebuild the anchor operator, and compare the Chebyshev-branch output of the first DCSE layer with and without the edit on the unchanged test inputs. For each $p$ we record the realised $\|\Delta L\|_\F$ and $\eta$, the observed change $\|g(\Lt+\Delta L)-g(\Lt)\|_\F/\|Z\|_2$ averaged over time slices, the right-hand side of~\eqref{eq:ccheb} at the measured $\|\Lt_\theta\|_2$, and the forecast MAE. Figure~\ref{fig:edgedrop} reports the result, averaged over all runs and three dropout draws, and Supplementary Table~S3 lists every value.
 
Edge dropout moves the operator much further than the random sweep. At $p=0.05$ the realised $\|\Delta L\|_\F$ is already $1.40$ on METR-LA and $2.70$ on PEMS-BAY, above the largest random budget, because the degree normalisation and the rescaling by $\hat{\lambda}$ spread each removed edge over whole rows of the operator (Section~\ref{subsec:diff_bound}). At $p=0.5$ it reaches $5.83$ and $9.64$. The edit is also more concentrated in spectral norm: $\eta/\|\Delta L\|_\F$ lies between $0.13$ and $0.33$, against about $0.14$ and $0.11$ for the dense random draws on the two networks.
 
The observed change grows slightly less than linearly in $\|\Delta L\|_\F$. Its ratio to $\|\Delta L\|_\F$ falls from $0.65$ to $0.57$ on METR-LA and from $0.82$ to $0.61$ on PEMS-BAY as $p$ rises from $0.05$ to $0.5$. The bound grows slightly faster than linearly, because $c_{\mathrm{cheb}}$ increases with $\eta$. The observed change is therefore between $2.4\%$ and $3.1\%$ of the bound on METR-LA and between $2.0\%$ and $3.1\%$ on PEMS-BAY, and the ratio falls as the edit grows. A ratio near $3\%$ indicates that random edge edits lie far from the worst case of the bound, which Proposition~\ref{prop:tight} attains only on a constructed operator and input. Across a sevenfold range of $\|\Delta L\|_\F$ the ratio moves by a factor of about $1.6$, far less than the size of the edit.
 
At the forecast, MAE rises by $1.0\%$ at $p=0.3$ and $1.4\%$ at $p=0.5$ on METR-LA, and by $2.5\%$ and $3.1\%$ on PEMS-BAY. These changes are larger than in the random sweep, in line with the larger operator change, and PEMS-BAY is the more sensitive network at every level. With half of the static edges removed, the 60-minute MAE is still $3.144$ on METR-LA and $1.624$ on PEMS-BAY, below every baseline in Table~\ref{tab:pemsbay} and every baseline in Table~\ref{tab:metrla} except TITAN. The edit leaves the learned part of $A_\theta$, which carries weight $\alpha=0.45$, unchanged.
 
\subsection{RQ4: Frequency Shift Across Regimes}
\label{subsec:rq4}
 
\subsubsection{Where Frequency Shift Occurs}
\begin{figure*}
\centering
\includegraphics[width=\textwidth]{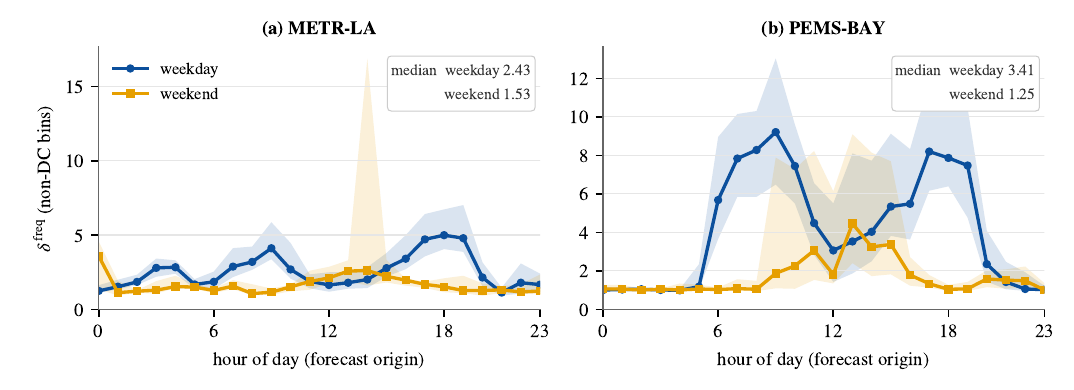}
\caption{Spectral shift $\delta^{\mathrm{freq}}$ of the test windows by hour of the forecast origin. Lines are medians and bands interquartile ranges, which show spread, not confidence intervals. A weekend hour holds about 72 windows on METR-LA and 120 on PEMS-BAY, a weekday hour about 216 and 312.}
\label{fig:dfreq}
\end{figure*}

For every test window of both datasets we computed $\delta^{\mathrm{freq}}$ of Definition~\ref{def:shift}, with $\bar{S}$ estimated from 20{,}000 training windows drawn at random. Figure~\ref{fig:dfreq} groups the windows by the hour of the forecast origin and by weekday or weekend. Spectral shift concentrates in weekday peak hours, when queues form and clear inside the 12-step window. On METR-LA the median $\delta^{\mathrm{freq}}$ is $2.43$ on weekdays and $1.53$ on weekends, and on PEMS-BAY it is $3.41$ and $1.25$. A two-sided Mann--Whitney $U$ test gives $p<10^{-20}$ on both datasets. Consecutive windows overlap in eleven of twelve steps and are not independent, so this $p$-value is nominal, and the evidence is the separation of the medians and interquartile ranges in Figure~\ref{fig:dfreq}. Weekday peaks form the high-shift regime and weekends the low-shift one.
 
\subsubsection{Accuracy Across Regimes}
Table~\ref{tab:naive} scores two naive forecasters on the same windows with the same evaluation code and mask as ST-GFM. Weekend windows are the easy regime as well as the low-shift one. The historical average improves by $16.7\%$ on METR-LA weekends and by $35.5\%$ on PEMS-BAY weekends, and persistence improves by $32$ to $46\%$ on PEMS-BAY. On METR-LA, ST-GFM (seed 42, 60 minutes) moves from $3.112$ on the full test set to $2.581$ on weekends, an improvement of $17.1\%$ and almost exactly the gain of the historical average. Its relative margin over the historical average is therefore almost unchanged, $25.8\%$ on the full test set and $26.1\%$ on weekends, and its margin over persistence grows from $39.5\%$ to $49.1\%$ because persistence gains little from the weekend on this network. ST-GFM tracks the easier regime. The full test set is dominated by weekday windows, which contain the high-shift peak hours, and ST-GFM keeps the same relative lead over the historical average on the full test set as on weekends. We did not score weekday peak windows separately. RQ4 characterises where frequency shift occurs and how the full model fares on weekends relative to the full test set.
 
\begin{table*}[!t]
\centering
\caption{MAE of Two Naive Forecasters on the Full Test Sets and on Weekend Windows (First Forecast Step on Saturday or Sunday), Scored With the Same Windows, Code and Mask as ST-GFM. $\Delta$ Is the Change From Full to Weekend.}
\label{tab:naive}
\footnotesize
\setlength{\tabcolsep}{4pt}
\begin{tabular}{lccccccccc}
\toprule
& \multicolumn{3}{c}{$H=3$} & \multicolumn{3}{c}{$H=6$} & \multicolumn{3}{c}{$H=12$} \\
\cmidrule(lr){2-4}\cmidrule(lr){5-7}\cmidrule(lr){8-10}
Method & Full & Weekend & $\Delta$ (\%) & Full & Weekend & $\Delta$ (\%) & Full & Weekend & $\Delta$ (\%) \\
\midrule
\multicolumn{10}{l}{\textbf{METR-LA}}\\
Persistence (last reading) & 3.557 & 3.337 & $-6.2$ & 4.153 & 3.971 & $-4.4$ & 5.145 & 5.073 & $-1.4$ \\
Historical average (time of week) & 4.193 & 3.496 & $-16.6$ & 4.193 & 3.493 & $-16.7$ & 4.193 & 3.494 & $-16.7$ \\
\midrule
\multicolumn{10}{l}{\textbf{PEMS-BAY}}\\
Persistence (last reading) & 1.301 & 0.887 & $-31.8$ & 1.649 & 1.016 & $-38.4$ & 2.173 & 1.169 & $-46.2$ \\
Historical average (time of week) & 2.631 & 1.698 & $-35.5$ & 2.631 & 1.697 & $-35.5$ & 2.630 & 1.697 & $-35.5$ \\
\bottomrule
\end{tabular}
\end{table*}
 
\subsection{Exploratory Transfer to Industry Portfolio Returns}
\label{subsec:transfer}
 
This exploratory study asks whether an encoder pre-trained only on traffic can support forecasting on a graph with different structure and dynamics. It sits outside the four research questions, and we do not claim that ST-GFM is a strong return model. The graph links the 43 portfolios within nine broad sector groups formed from the Fama--French industry definitions, which gives 218 edges, and enters the model as $A_{\mathrm{static}}$. Weights are transferred by name and shape. Fine-tuning trains the second DCSE layer, the second temporal-attention block, the aggregation layer, the decoder with its node embedding and the persistence term, $3.1$ million parameters in all. The decomposition, the input projection, the time and step embeddings, the gated skip and the first DCSE and attention pair stay frozen at their traffic values. Parameters whose shape depends on the number of nodes are re-initialised, and of these only the decoder node embedding is trained. The additive node embedding, the degree embedding and the adaptive-graph vectors $E_1,E_2$ stay frozen at their new random values, and $\alpha_0$ is set to $0$. The operator the DCSE layers see is therefore half sector graph and half $\mathrm{softmax}(\mathrm{ReLU}(E_1E_2^\top))$ built from random vectors, and the transferred encoder is not a fully pretrained backbone. Transferred weights train at a learning rate of $3\times10^{-5}$ and the decoder node embedding at $3\times10^{-4}$. The loss combines Pearson IC, rank IC and cross-sectional MSE, and early stopping uses validation IC. Each window holds $12$ months of input and $12$ months of targets, and we score the one-month-ahead forecast. Test windows are cut from the post-2010 data alone, so the scored targets are the $172$ months from January 2011 to April 2025.
 
\begin{figure}
\centering
\begin{tikzpicture}
\begin{groupplot}[
  group style={group size=2 by 1, horizontal sep=1.15cm},
  width=0.56\columnwidth, height=4.6cm,
  ybar, /pgf/bar width=7pt,
  xmin=0.4, xmax=6.6,
  xtick={1,2,3,4,5,6},
  xticklabels={LA 1, LA 2, BAY 1, BAY 2, Rnd 1, Rnd 2},
  xticklabel style={font=\tiny, rotate=40, anchor=north east, inner sep=1pt},
  label style={font=\footnotesize}, tick label style={font=\scriptsize},
  title style={font=\footnotesize\bfseries, yshift=-4pt},
  ymajorgrids, grid style={black!8},
  axis line style={black!45}, tick style={black!45},
  enlarge x limits=false,
]
\nextgroupplot[title={(a) Mean IC}, ymin=0, ymax=0.072, ytick={0,0.02,0.04,0.06}, scaled y ticks=false,
  yticklabel style={/pgf/number format/fixed,/pgf/number format/fixed zerofill,/pgf/number format/precision=2}]
\addplot[bar shift=0pt, fill=cOurs!85, draw=cOurs!70!black] coordinates {(1,0.055) (2,0.043)};
\addplot[bar shift=0pt, fill=cTG!75, draw=cTG!70!black] coordinates {(3,0.052) (4,0.059)};
\addplot[bar shift=0pt, fill=cMega!45, draw=cMega!90!black] coordinates {(5,0.042) (6,0.036)};
\draw[black!70, line width=0.9pt, densely dashed] (axis cs:0.55,0.05225) -- (axis cs:4.45,0.05225);
\draw[black!70, line width=0.9pt, densely dashed] (axis cs:4.55,0.039) -- (axis cs:6.45,0.039);
\node[font=\tiny, text=black!75, anchor=south] at (axis cs:2.5,0.0608) {mean 0.052};
\node[font=\tiny, text=black!75, anchor=south] at (axis cs:5.5,0.0435) {0.039};
\nextgroupplot[title={(b) Sharpe ratio, net}, ymin=0, ymax=0.84, ytick={0,0.2,0.4,0.6,0.8},
  yticklabel style={/pgf/number format/fixed,/pgf/number format/fixed zerofill,/pgf/number format/precision=1}]
\addplot[bar shift=0pt, fill=cOurs!85, draw=cOurs!70!black] coordinates {(1,0.57) (2,0.54)};
\addplot[bar shift=0pt, fill=cTG!75, draw=cTG!70!black] coordinates {(3,0.60) (4,0.71)};
\addplot[bar shift=0pt, fill=cMega!45, draw=cMega!90!black] coordinates {(5,0.49) (6,0.42)};
\draw[black!70, line width=0.9pt, densely dashed] (axis cs:0.55,0.605) -- (axis cs:4.45,0.605);
\draw[black!70, line width=0.9pt, densely dashed] (axis cs:4.55,0.455) -- (axis cs:6.45,0.455);
\node[font=\tiny, text=black!75, anchor=south] at (axis cs:2.5,0.725) {mean 0.61};
\node[font=\tiny, text=black!75, anchor=south] at (axis cs:5.5,0.50) {0.46};
\end{groupplot}
\end{tikzpicture}
\caption{Transfer to the 43 Fama--French industry portfolios (Table~\ref{tab:finance}). Blue bars use METR-LA encoders, green bars PEMS-BAY encoders and gray bars the randomly initialised frozen backbone. Dashed lines are group means.}
\label{fig:finance}
\end{figure}
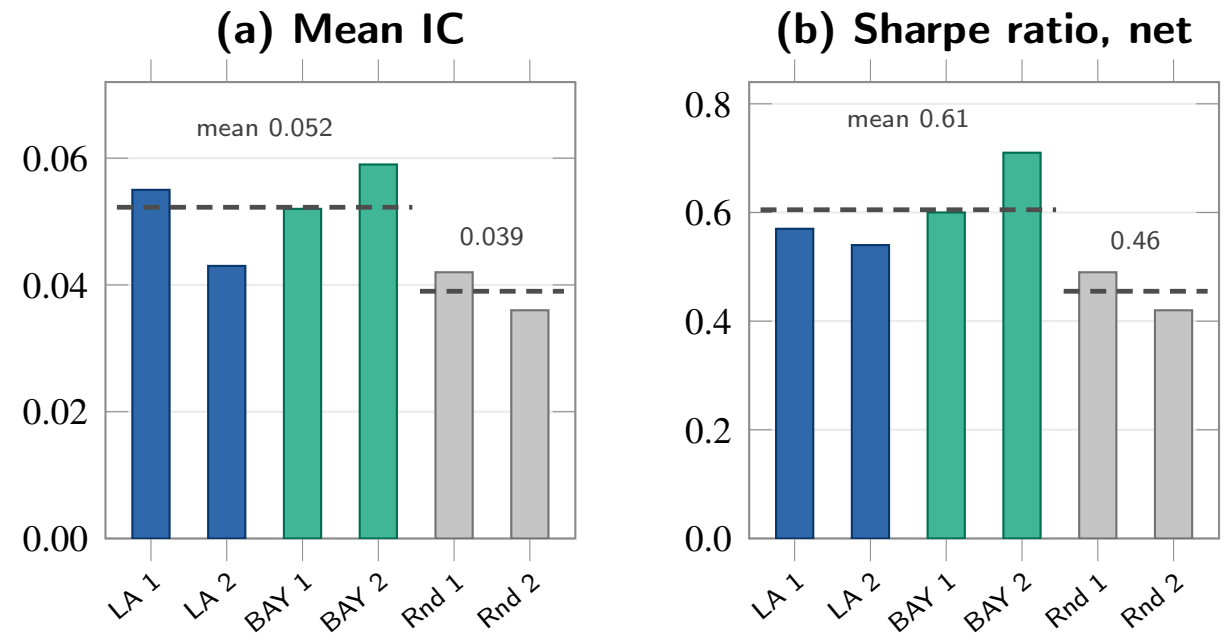

Table~\ref{tab:finance} reports, for two encoders per traffic dataset, the mean monthly cross-sectional information coefficient (IC), its rank version, the information ratio of the IC series, and the annualised Sharpe ratio of a long-short decile portfolio net of $1$ basis point of cost. A control repeats the fine-tuning from a random initialisation of the same architecture, with two runs. Figure~\ref{fig:finance} plots IC and Sharpe ratio for all six runs.
 
\begin{table}[!t]
\centering
\caption{Transfer to 43 Fama--French Industry Portfolios (172 Scored Months, January 2011 to April 2025). Runs \#1 and \#2 Use METR-LA Seeds 202 and 303 and PEMS-BAY Seeds 101 and 303. R-IC Is Rank IC, IR the IC Information Ratio, SR the Annualised Sharpe Ratio Net of 1~bp and $p$ the One-Sided $t$-Test of Mean IC.}
\label{tab:finance}
\scriptsize
\setlength{\tabcolsep}{2pt}
\begin{tabular}{@{}lccccc@{}}
\toprule
Encoder & IC & R-IC & IR & SR & $p$ \\
\midrule
\textbf{ST-GFM (METR-LA $\to$ FF49)}\,\#1  & 0.055 & 0.049 & 0.21 & 0.57 & 0.004 \\
\textbf{ST-GFM (METR-LA $\to$ FF49)}\,\#2  & 0.043 & 0.041 & 0.16 & 0.54 & 0.019 \\
\textbf{ST-GFM (PEMS-BAY $\to$ FF49)}\,\#1 & 0.052 & 0.051 & 0.20 & 0.60 & 0.006 \\
\textbf{ST-GFM (PEMS-BAY $\to$ FF49)}\,\#2 & 0.059 & 0.057 & 0.22 & 0.71 & 0.003 \\
\midrule
Random frozen backbone, \#1 & 0.042 & 0.040 & 0.16 & 0.49 & 0.020 \\
Random frozen backbone, \#2 & 0.036 & 0.034 & 0.14 & 0.42 & 0.038 \\
\bottomrule
\end{tabular}
\end{table}
 
All four transferred runs give a mean IC significantly above zero, with IC between $0.043$ and $0.059$ and Sharpe ratios between $0.54$ and $0.71$. The result survives a Holm correction across the four runs (largest adjusted $p=0.019$). The signal is noisy: an information ratio near $0.2$ means the monthly IC has a standard deviation about five times its mean.
 
The random control sets the scale of the transfer effect, and it is modest. The random-initialisation control reaches an IC of $0.042$ and $0.036$, both significant at the $5\%$ level, and Sharpe ratios of $0.49$ and $0.42$. Averaged over runs, traffic pre-training raises the IC from $0.039$ to $0.052$ and the Sharpe ratio from $0.46$ to $0.61$. The random controls reach about three quarters of the mean IC of the transferred runs. Both arms use the same design for the re-initialised node-dependent modules. Every transferred run is at or above every control run on both measures, but the weakest transferred run (IC $0.043$) and the strongest control (IC $0.042$) are close. The small number of trained runs limits assessment of training variability, and we do not report a direct statistical comparison between the transfer and random-initialisation conditions. Under this adaptation protocol, traffic initialisation gives a consistent numerical advantage over random initialisation.
 
\subsection{Computational Cost}
\label{subsec:cost}
 
ST-GFM has $5.22$ million parameters. A training run takes $3$ to $4$ hours on METR-LA and $8$ to $9$ hours on PEMS-BAY on one RTX 5060 Ti, and inference on a batch of $64$ windows takes about $15$~ms. The model is larger than some baselines, and two observations argue against size as the main source of its accuracy. T-Graphormer reports that its $4.15$-million-parameter variant does worse on METR-LA than its $1.57$-million one because it overfits~\cite{bai2025tgraphormer}, and ST-GFM loses $0.30$ MAE when MSR is removed with the parameter count unchanged.
 
\begin{figure}
\centering
\begin{tikzpicture}
\begin{axis}[
  width=\columnwidth, height=5.6cm,
  xmin=0, xmax=6.0, ymin=3.07, ymax=3.385,
  xtick={0,1,2,3,4,5,6}, ytick={3.10,3.15,3.20,3.25,3.30,3.35},
  yticklabel style={/pgf/number format/fixed,/pgf/number format/fixed zerofill,/pgf/number format/precision=2},
  xlabel={parameters (millions)}, ylabel={METR-LA, 60-minute MAE (mph)},
  label style={font=\footnotesize}, tick label style={font=\scriptsize},
  grid=major, grid style={black!8},
  axis line style={black!45}, tick style={black!45},
  clip=false,
]
\addplot[cSTAE, line width=0.9pt, dash pattern=on 4pt off 2pt, domain=0:6, samples=2] {3.34};
\node[font=\scriptsize, text=cSTAE!85!black, anchor=south east, inner sep=1.5pt] at (axis cs:5.95,3.343) {STAEformer, published (3.34)};
\addplot[cSTAE, line width=0.9pt, densely dotted, domain=0:6, samples=2] {3.21};
\node[font=\scriptsize, text=cSTAE!85!black, anchor=south east, inner sep=1.5pt] at (axis cs:5.95,3.213) {STAEformer, our pipeline (3.21)};
\addplot[cOurs, line width=1.5pt, mark=*, mark size=2.4pt, mark options={fill=white, line width=1.1pt}] coordinates {(0.93,3.15) (1.98,3.12) (3.41,3.10) (5.22,3.11)};
\addplot[only marks, mark=star, mark size=4.0pt, line width=1.2pt, cOurs] coordinates {(5.22,3.098)};
\node[font=\scriptsize, anchor=south, align=center] at (axis cs:0.93,3.153) {$d=128$\\3.15, 1.2 h};
\node[font=\scriptsize, anchor=south west, align=left] at (axis cs:2.02,3.122) {$d=192$\\3.12, 1.7 h};
\node[font=\scriptsize, anchor=south west, align=left] at (axis cs:3.45,3.102) {$d=256$\\3.10, 2.2 h};
\node[font=\scriptsize\bfseries, text=cOurs, anchor=south, align=center] at (axis cs:5.22,3.118) {$d=320$, full\\3.11, 4 h};
\node[font=\scriptsize, text=cOurs, anchor=north] at (axis cs:5.22,3.092) {five-seed mean};
\end{axis}
\end{tikzpicture}
\caption{Accuracy against model size on METR-LA at 60 minutes. Each label gives the width, the 60-minute MAE and the training time on one RTX 5060 Ti. Circles are single runs with seed 42, and Supplementary Table~S4 lists their values. The star is the five-seed mean of the full model. The dashed line is the published STAEformer MAE and the dotted line STAEformer retrained under our pipeline.}
\label{fig:width}
\end{figure}
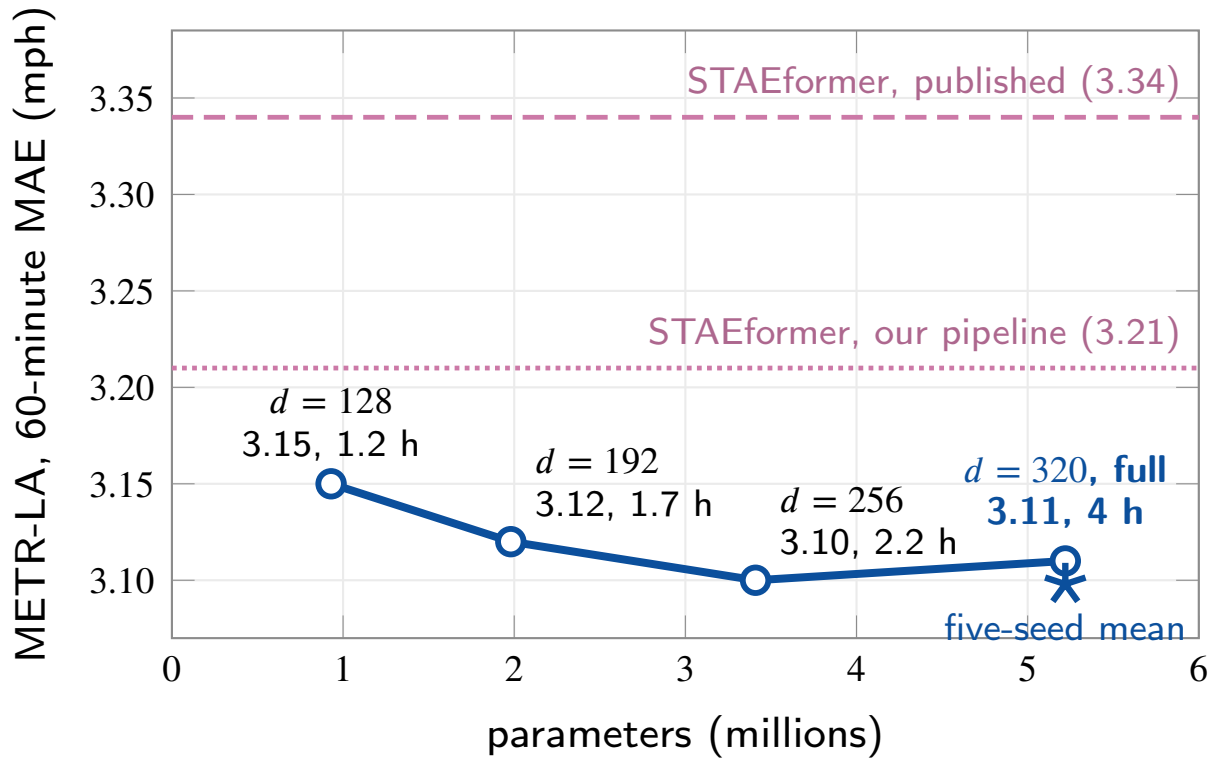

Figure~\ref{fig:width} varies the hidden width with everything else fixed, using seed $42$ at every width, and Supplementary Table~S4 lists the values. At $d=128$ the model has $0.93$ million parameters, trains in $1.2$ hours and reaches a 60-minute MAE of $3.15$ on METR-LA. That is $0.04$ above the full model with the same seed, below every baseline in Table~\ref{tab:metrla} except TITAN, and below STAEformer retrained under our pipeline ($3.21$). Accuracy stops improving by $d=256$. The single runs give $3.10$ at $d=256$ and $3.11$ at $d=320$, a difference smaller than the seed standard deviation of $0.014$, so going from $d=256$ to $d=320$ adds $1.8$ million parameters and $1.8$ hours of training without lowering the error. At $d=192$ the model comes within $0.01$ of the full model in less than half its training time.
 
\section{Discussion}
\label{sec:discussion}
\begin{itemize}
    \item \textbf{Where the accuracy comes from:} The same-pipeline comparison separates two cases. On PEMS-BAY the margin over a retrained STAEformer stays between $12.9\%$ and $15.0\%$ at every horizon, so most of the gain survives a shared pipeline. On METR-LA the 60-minute margin falls from $7.2\%$ against the published value to $3.5\%$ against the retrained baseline. At 15 minutes the models are level at the precision reported for the retrained baseline. The spectral encoder and MSR have the largest ablation effects, and collapsing the time axis also increases error. The width study retains competitive accuracy at much smaller sizes, which suggests that parameter count alone does not explain the performance.
    \item \textbf{What the operator bounds establish:} The numerical content of the analysis is the pair of branch constants computed from the trained model. Proposition~\ref{prop:tight} shows that the Chebyshev bound is first-order sharp when nothing further is assumed about $\Theta_k$ and $Z$. The low observed-to-bound ratios show that the bound is conservative for the sampled edge-dropout perturbations. They do not determine the smallest uniform constant for the trained weights. The approximately linear observed response is consistent with the dependence of the bound on perturbation size. The diffusion bound is about ten times looser for row-filling edits, mostly through the $\sqrt{s_\Delta}$ factor.
    \item \textbf{What the MSR evidence shows:} Every run reaches its best validation MAE near the end of the self-supervised phase, and removing MSR while the other two losses keep the same schedule still costs $0.30$ MAE. Both observations are consistent with MSR acting as a regulariser during the early training phases, although they do not establish its mechanism. Its link to the frequency channel rests on the form of the loss (Section~\ref{subsec:msr_is}), which keeps the fast, low-power bins of weekday peak windows, the high-shift regime of RQ4, in the objective. Section~\ref{sec:limitations} lists what this evidence does not cover.
\end{itemize}

\section{Scope and Limitations}
\label{sec:limitations}
 
All four datasets come from Californian freeways with at most 325 sensors. The dense learned adjacency costs $O(N^2)$ per graph multiplication, so networks of the size of LargeST~\cite{liu2023largest} would need a sparse learned operator. Most baseline numbers are published values, and checkpoint averaging, which ST-GFM uses, is not part of the published baseline training procedures. The two retrained baselines score better under our pipeline than in their papers, and we have not isolated the cause of that gain. They are single runs that follow their own training recipes, so the same-pipeline comparison shares data handling and evaluation and not the full training protocol. The ablated variants, the width study and the flow-dataset results are single runs. The random perturbation sweeps use one draw per run and level, and the structured study covers one perturbation family, edge dropout of the static graph, with three draws per level. Edge dropout removes edges and keeps every input reading, so it does not model a failed sensor, which loses both. $\delta^{\mathrm{top}}$ cannot be measured on real data because the true coupling of a window is not observed, so the operator channel is studied through controlled perturbations. The evidence on the frequency channel is weaker. RQ4 locates frequency shift but scores ST-GFM only on the full test set and on METR-LA weekends, not on weekday peak windows. It does not compare variants with and without MSR across shift levels, and MSR has not been compared with a time-domain reconstruction loss on the same target. MSR also works on seven frequency bins of one readout of the encoder output, which limits the spectral detail it can enforce on a 12-step window. The exploratory transfer study uses 43 portfolios, four transferred runs and two random controls. In both arms the re-initialised node-dependent modules stay frozen at random values, except the decoder node embedding, which is trained.
 
\section{Conclusion}
\label{sec:conclusion}
 
ST-GFM splits spectral shift into an operator channel and a frequency channel. For the operator channel, both graph branches admit Frobenius-norm bounds. The Chebyshev constant computed from the trained model is $17.9$ on METR-LA and $21.5$ on PEMS-BAY, and under the tested edge-dropout perturbations the observed Chebyshev-branch response is at most $3.1\%$ of the corresponding computed bound. The frequency channel motivates MSR, which aligns the log spectra of the encoder output and its input and has one of the two largest ablation effects. With $5.22$ million parameters on one consumer GPU, ST-GFM has a lower MAE and MAPE than all twelve published baselines at every evaluated horizon on PEMS-BAY, and a lower MAE than a retrained STAEformer at 30 and 60 minutes on both speed datasets.
 
The next step is the sparse learned operator. Replacing the dense softmax term with a top-$k$ neighbourhood per node would bring the cost of a graph multiplication from $O(N^2d)$ to $O(Nkd)$ and would let the bounds, which depend on the number of nonzero entries per row of the perturbation, be stated for networks with thousands of sensors.
 





\section*{Supplementary Material}
Supplementary material for this article is provided as a separate file. It contains the full training configuration (Table S1), the training procedure (Algorithm S1), the results of every individual run (Table S2), the Chebyshev constant against perturbation size (Figure S1), the ablation chart (Figure S2), the edge-dropout values (Table S3) and the width-study values (Table S4).
 


\bibliographystyle{cas-model2-names}
\bibliography{Bibliography}

\appendix 
\section{Mathematical Proofs}
\label{app:proofs}
 
\subsubsection*{Proof of Lemma~\ref{lem:diff}}
Write $\mathcal{D}=\diag(\hat{A}\mathbf{1})$ and $\mathcal{D}'=\diag(\hat{A}'\mathbf{1})$, so that $P_f=\hat{A}\mathcal{D}^{-1}$ and $\|\mathcal{D}^{-1}\|_2,\|\mathcal{D}'^{-1}\|_2\le1/m$. Hence $\|P_f\|_2,\|P_b\|_2\le p:=a/m$, and the same holds under $\Lt'$. Since $\Delta\hat{A}=-\tfrac12\Delta L$,
\[
  P_f'-P_f = \Delta\hat{A}\,\mathcal{D}'^{-1} + \hat{A}(\mathcal{D}'^{-1}-\mathcal{D}^{-1}).
\]
The first term has Frobenius norm at most $\|\Delta L\|_\F/(2m)$. For the second, $|1/d_i'-1/d_i|\le|d_i'-d_i|/m^2$, and the vector of row-sum changes is $\Delta\hat{A}\mathbf{1}=-\tfrac12\Delta L\mathbf{1}$. Row $i$ of $\Delta L$ has at most $s_\Delta$ nonzero entries, so by Cauchy--Schwarz $\|\Delta L\mathbf{1}\|_2\le\sqrt{s_\Delta}\,\|\Delta L\|_\F$. The second term is therefore at most $a\sqrt{s_\Delta}\,\|\Delta L\|_\F/(2m^2)$, and
\[
  \|P_f'-P_f\|_\F\le\delta:=\frac{1}{2m}\Big(1+\frac{a\sqrt{s_\Delta}}{m}\Big)\|\Delta L\|_\F .
\]
The same bound holds for $P_b'-P_b=(P_f'-P_f)^\top$. Let $Q=P_f+P_b$, so that $\|Q\|_2\le2p$ and $\|Q'-Q\|_\F\le2\delta$. Step $s'$ applies $P_fQ^{s'-1}$ or $P_bQ^{s'-1}$ to $Z$. Changing one factor of the product at a time,
\begin{align*}
  &\|P_f'Q'^{s'-1}-P_fQ^{s'-1}\|_\F\\
  &\quad\le\delta(2p)^{s'-1}+(s'-1)\,p\,(2p)^{s'-2}\,2\delta=s'(2p)^{s'-1}\delta,
\end{align*}
and the same holds for $P_b$. Multiplying by $\|Z\|_2\|W\|_2$ and summing over steps and both directions gives~\eqref{eq:diff_bound}. \hfill$\square$
 
\subsubsection*{Proof of Proposition~\ref{prop:stack}}
Fix the input and let $e_1,\dots,e_4$ be the Frobenius distances between outputs under $L'$ and $L$ after $\mathrm{DCSE}_1$, $\mathrm{TempAttn}_1$, $\mathrm{DCSE}_2$ and $\mathrm{TempAttn}_2$. Both runs feed $\mathrm{DCSE}_1$ the same input, so $e_1\le \kappa_L\|\Delta L\|_\F$. The temporal block does not depend on the operator, so $e_2\le \kappa_T\,e_1$. Then $e_3\le \kappa_Z\,e_2+\kappa_L\|\Delta L\|_\F$ and $e_4\le \kappa_T\,e_3$. Substituting gives $e_4\le \kappa_T\,\kappa_L\,(1+\kappa_Z\,\kappa_T)\|\Delta L\|_\F$. \hfill$\square$
 
\subsubsection*{Proof of Proposition~\ref{prop:is}}
With $P_X/P_Z=e^{-u}$, each term of $D_{\mathrm{IS}}$ equals $e^{-u}+u-1=\tfrac12u^2-\tfrac16u^3+O(u^4)$. Summing over bins, $\sum u^2=2D_{\mathrm{IS}}+\tfrac13\sum u^3+O(\sum u^4)$. On unclamped bins $\mathcal{L}_{\mathrm{freq}}=\sum u^2/(NW's_X^2)$, which gives the result. \hfill$\square$
 
\section{Implementation Details}
\label{app:impl}
 
\subsubsection*{Preprocessing}
Readings are standardised per node with training-set statistics. The static adjacency is $A_{ij}=\exp(-\mathrm{dist}_{ij}^2/2\sigma^2)$ when this value is at least $0.1$ and zero otherwise, with $\sigma$ the standard deviation of the finite pairwise road distances. Self-loops are removed and the matrix is symmetrised. Time of day uses 288 five-minute bins and day of week 7 bins, each mapped to a $24$-dimensional learned embedding. Supplementary Table~S1 lists the full configuration and Supplementary Algorithm~S1 the training procedure.
 
\subsubsection*{Fama--French Transfer}
Monthly value-weighted returns of the 43 industry portfolios with complete histories are each divided by a trailing standard deviation over the previous $60$ months, with at least $20$ months and a one-month lag, so month $t$ is scaled by months $t-60$ to $t-1$. The scaled returns are winsorised at $\pm5$ and standardised cross-sectionally each month. The first $20$ months have no volatility estimate and are dropped. Windows hold $12$ months of input and $12$ months of targets. The graph links portfolios in the same one of nine broad sector groups (218 edges) and is fixed during fine-tuning. Section~\ref{subsec:transfer} lists the trained and frozen modules. Data before 2010 are used for fine-tuning (717 training and 162 validation windows), and months from 2010 onward for evaluation. The long-short portfolio buys the top decile of predicted returns and sells the bottom decile each month. The random control uses the same architecture, trainable modules, loss, learning rates and early stopping, with every module initialised at random instead of from traffic.


\end{document}